\documentclass[letterpaper]{article}
\usepackage[preprint]{aaai2027}
\usepackage[hyphens]{url}
\usepackage{graphicx}
\usepackage{natbib}
\usepackage{caption}
\usepackage{algorithm}
\usepackage{algorithmic}
\usepackage{newfloat}
\usepackage{listings}
\usepackage{booktabs}
\usepackage{placeins}
\DeclareCaptionStyle{ruled}{labelfont=normalfont,labelsep=colon,strut=off}
\floatstyle{ruled}
\newfloat{listing}{tb}{lst}{}
\floatname{listing}{Listing}

\usepackage{booktabs}
\usepackage{amsmath}
\usepackage{amssymb}
\usepackage{amsfonts}
\usepackage{nicefrac}
\usepackage{microtype}
\usepackage{xcolor}
\usepackage{multirow}
\usepackage{array}
\usepackage{makecell}
\usepackage{pifont}
\usepackage{tcolorbox}
\tcbuselibrary{skins,breakable}
\usepackage{enumitem}

\definecolor{promptshade}{RGB}{247,248,250}
\definecolor{prompttitle}{RGB}{48,55,64}
\definecolor{bestblue}{RGB}{31,78,121}

\newcolumntype{L}[1]{>{\raggedright\arraybackslash}p{#1}}
\tcbset{
  promptbox/.style={
    breakable, enhanced, frame hidden,
    colback=promptshade, boxrule=0pt,
    left=6pt, right=6pt, top=5pt, bottom=5pt,
    before skip=6pt, after skip=4pt,
    fonttitle=\bfseries\small\sffamily,
    coltitle=prompttitle,
    colbacktitle=promptshade,
    boxed title style={frame hidden, colback=promptshade},
    attach boxed title to top left={xshift=0pt,yshift=-1mm},
    borderline west={1.2pt}{0pt}{black!35}
  },
  guibox/.style={promptbox},
  codebox/.style={promptbox},
  originbox/.style={promptbox},
  claudebox/.style={promptbox}
}

\newcommand{\cmark}{\ding{51}}

\title{MatToolBench: Benchmarking Multimodal Agents in Real-World Materials Science Workflows}
\author{
  Mei Wu\textsuperscript{\rm 1,2}\equalcontrib,
  Rui Xie\textsuperscript{\rm 1,4}\equalcontrib,
  Runyu Zhang\textsuperscript{\rm 1},
  Yuqiang Li\textsuperscript{\rm 3},
  Tianfan Fu\textsuperscript{\rm 3},\\
  Bo Chen\textsuperscript{\rm 2}\corresponding,
  Kai Yu\textsuperscript{\rm 1,2,6},
  Xin Chen\textsuperscript{\rm 2},
  Lu Chen\textsuperscript{\rm 1,2,5,6}\corresponding
}
\affiliations{
  \small
  {\fontsize{11}{13}\selectfont\color{bestblue}\url{https://mattoolbench.github.io/}}\\[0.35em]
  \textsuperscript{\rm 1}X-LANCE Lab, School of Computer Science, Shanghai Jiao Tong University, Shanghai, China\\
  \textsuperscript{\rm 2}Suzhou Laboratory, Suzhou, China\quad
  \textsuperscript{\rm 3}Shanghai Artificial Intelligence Laboratory, Shanghai, China\\
  \textsuperscript{\rm 4}State Key Laboratory for General Artificial Intelligence, BIGAI, Beijing, China\\
  \textsuperscript{\rm 5}Shanghai Innovation Institution, Shanghai, China\quad
  \textsuperscript{\rm 6}Jiangsu Key Lab of Language Computing, Suzhou, China\\[0.25em]
  Correspondence: chenb@szlab.ac.cn; chenlusz@sjtu.edu.cn
}

\begin{document}
\maketitle

\begin{abstract}
Multimodal GUI agents have achieved impressive results on general software benchmarks,
yet their ability to operate professional scientific software remains largely unexplored.
In materials science, sparse domain-specific web data, specialized interfaces, and
tacit workflow conventions create blind spots that general-purpose pretraining cannot
readily bridge.

We present \textbf{MatToolBench}, the first real-environment benchmark for evaluating
multimodal GUI agents on professional materials science software, comprising
\textbf{204 tasks} across \textbf{10 tools} in three modalities: GUI operation,
OriginPro scripting, and code-based database queries, all executed inside a
Windows~11 VM. Each task is decomposed into fine-grained sub-criteria by domain
experts, enabling interpretable partial-credit scoring; the GUI component of our
multi-level evaluation pipeline achieves an average F1 of \textbf{0.98}. For
OriginPro figure-generation tasks, we further conduct a human--LLM agreement study
to validate the use of a multimodal judge for secondary aesthetic assessment.

Our experiments show that strong performance on general benchmarks does not
transfer to professional scientific workflows, and that this gap is not a
visual-grounding problem alone: failures arise from domain-specific operational knowledge, sparse
pretraining coverage of scientific software, weak cross-tool artifact handoff,
and critical states exposed only visually. Even the best model reaches only \textbf{25\%} success
rate on GUI tasks and \textbf{45\%} on code tasks. MatToolBench therefore serves as
a challenging diagnostic benchmark and real-environment testbed for data-scarce,
knowledge-intensive scientific workflows.

\end{abstract}
\section{Introduction}

With the rapid development of multimodal large language models and GUI-specialized visual agents~\cite{gpt4o, cogagent}, GUI agents have achieved remarkable progress in visual perception, decision-making, and grounding capabilities. They have demonstrated strong performance in general-purpose scenarios~\cite{osworld, webarena}, such as web browsing~\cite{webarena}, desktop operation~\cite{osworld}, or synthetic environments~\cite{alfworld}. However, existing research remains predominantly confined to everyday general-purpose software. Due to the scarcity of web data related to professional scientific software, the pre-trained knowledge of large models in this vertical domain is inevitably constrained, leaving it an open question whether they can exhibit similarly excellent performance in complex, real-world scientific workflows.

Materials chemistry tools, characterized by complex graphical interfaces and strong domain dependencies, offer an ideal platform for systematically testing the domain-specific capabilities of multimodal GUI agents.
\begin{figure}[!t]
  \centering
  \includegraphics[width=\linewidth]{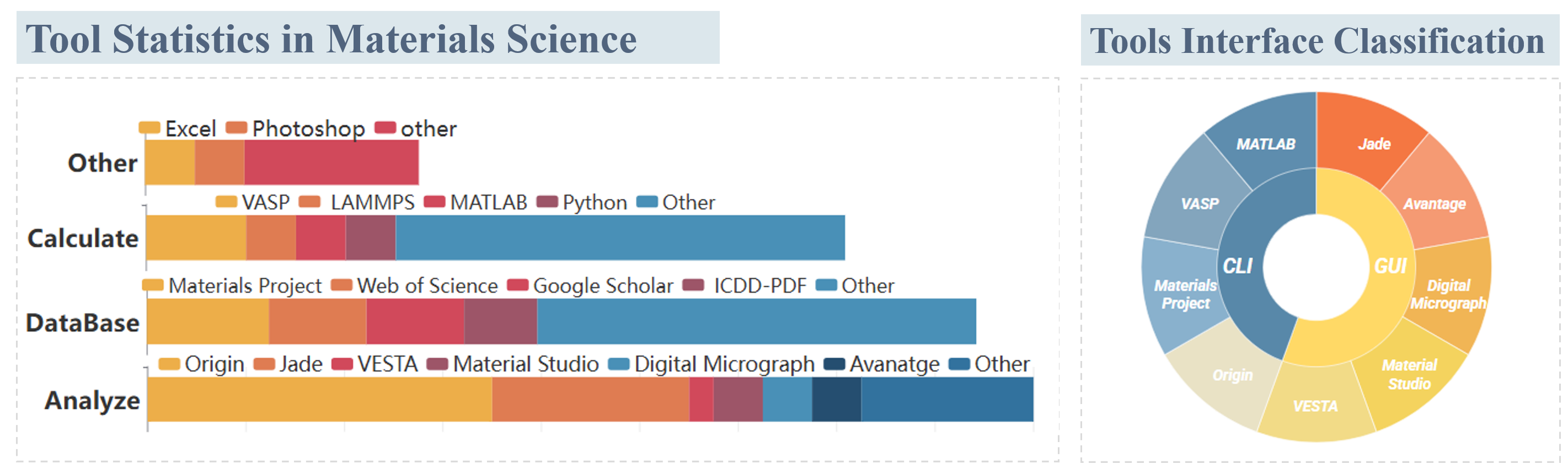}
  \caption{Survey results from 25 materials graduate students:
  tool usage frequency across sub-fields (left) and
  distribution of primary interaction interface types (right).}
  \label{fig1}
\end{figure}
Compared to general-purpose scenarios, the use of materials tools exhibits distinctive complexity: (1) Highly fragmented tool ecosystem. As shown in Figure \ref{fig1}, a survey of materials science graduate students reveals that different research subfields and usage requirements correspond to a large number of heterogeneous tools, which often belong to different operating environments and interaction interfaces. (2) Cross-tool artifact handoff. As shown in Figure \ref{fig2}, complete materials analysis may require switching between tools and preserving intermediate files, paths, and scientific meaning; we include these cases as diagnostic mixed workflows rather than as the main task split. (3) Severely limited accessibility interfaces. Such professional GUI software commonly features information-dense interfaces, deeply nested menu hierarchies, and implicit feedback mechanisms for critical operation states, lacking unified and standardized accessibility interfaces.

\begin{figure}[!t]
  \centering
  \includegraphics[width=\linewidth]{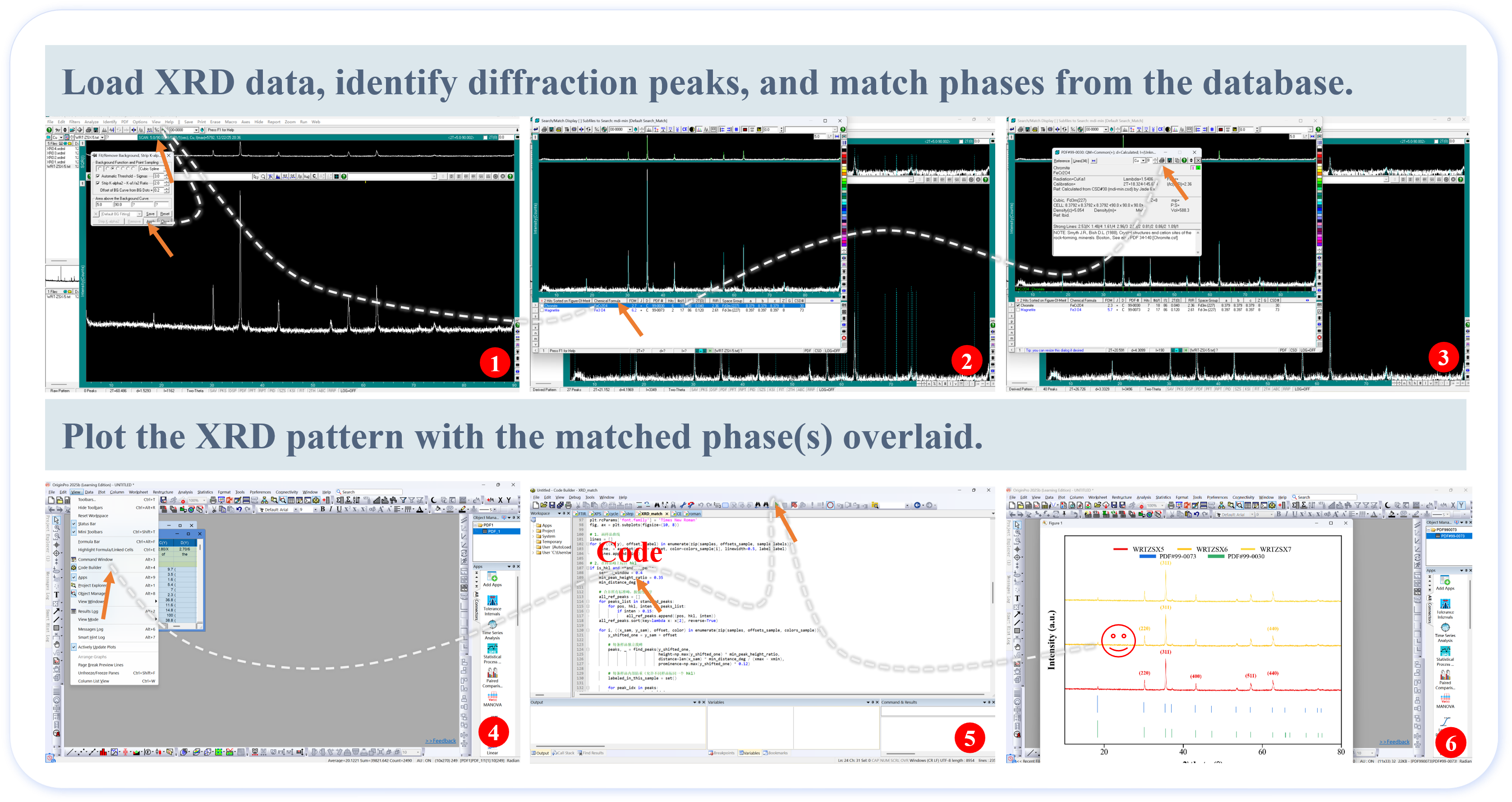}
  \caption{A complete XRD diffraction pattern requires cross-tool collaboration:
  raw XRD data is analyzed in JADE, then plotted and annotated in OriginPro.}
  \label{fig2}
  \vspace{-2pt}
\end{figure}

To fill this gap, we present \textbf{MatToolBench}, the first real-environment benchmark specifically centered on
professional materials-science software workflows, spanning GUI-based
desktop applications, scripted automation, and API-based database queries, and evaluates
multimodal GUI agents across all three modalities within a unified framework.
MatToolBench comprises \textbf{204 tasks} across \textbf{10 tools} covering
GUI operation, OriginPro scripting (a GUI+code hybrid modality), and code-based
database queries, all running inside a Windows~11 VM pre-installed with the complete
target software stack.
Tasks are stratified into three difficulty levels, and each is decomposed by
materials science experts into multiple independent scoring sub-criteria,
enabling fine-grained partial-credit scoring that reveals \emph{where} agents fail
in complex multi-step workflows rather than collapsing performance to a binary outcome.

To ensure reliable scoring, we design modality-specific assessment methods.
GUI results are extracted via OCR and pixel analysis and validated against a labeled
test set, where the GUI evaluator achieves an average F1 of \textbf{0.98}.
OriginPro tasks use a two-stage pipeline: deterministic figure-generation/export checks
for export SR, followed by LLM-assisted aesthetic assessment validated against human ratings through
correlation, confidence-interval, and significance analyses.
Code tasks are evaluated by matching reference outputs when deterministic answers exist.

We evaluate seven frontier multimodal models and find that professional materials science
software poses a substantially harder challenge than general benchmarks suggest.
Even the best-performing model reaches only 25.0\% GUI SR and 45.0\% code SR.
The difficulty is uneven across domains: DigitalMicrograph, with its dense microscopy
interface and implicit operation feedback, does not exceed 20.0\% SR for any model, while
OPTIMADE, whose query syntax is well-documented, reaches up to 85.0\% SR.
Origin tasks, which demand both GUI navigation and scripted automation,
show the widest model performance spread (0.0\%--68.8\% SR),
suggesting that hybrid modalities impose compounding capability requirements.

Ablation studies further show that domain-specific workflow guidance is decisive:
for Doubao-seed-1-8, removing hints collapses JADE SR from 25.0\% to 5.0\%;
for GPT-5.4, MP performance drops from 15.0\% to 0.0\%;
and for Claude-sonnet-4.6 on Origin tasks, removing full workflow support reduces the total score from 17/32 to 5.2/32.
\section{Related Work}
\subsection{GUI-based Agents}
GUI-agent benchmarks have expanded from early web-control environments
(World of Bits~\cite{worldofbits}) to web and mobile interaction
(WebArena~\cite{webarena}, VisualWebArena~\cite{visualwebarena},
WebVoyager~\cite{webvoyager}, Mind2Web~\cite{mind2web},
AITW~\cite{aitw}, Mobile-Env~\cite{mobileenv})
to desktop environments such as OSWorld/OSWorld-Verified~\cite{osworld,osworldverified},
WindowsAgentArena~\cite{windowsagentarena}, and recent 50-step Windows or
cross-platform GUI evaluations~\cite{ossymphony,mmbenchgui}. ProSoftArena~\cite{prosoftarena}
covers broad professional software suites, but these benchmarks do not target
materials-science workflows, specialized scientific GUIs, or domain-specific
evaluation criteria.
Other work benchmarks knowledge-work applications~\cite{workarena}, general
visual agent capabilities~\cite{visualagentbench}, and UI-oriented visual
understanding~\cite{screenai}, but not materials-science tool chains.
The need for software-specific knowledge is also studied in
GUIDE~\cite{guide}, which retrieves task-relevant tutorial videos to guide
planning and grounding.

\subsection{Science Benchmarks}
Scientific-agent benchmarks have progressed from textual reasoning
(SciBench~\cite{scibench}, MatSciBench~\cite{matscibench},
ChemLLMBench~\cite{chemllmbench}) to executable
code and tool use (ScienceAgentBench~\cite{scienceagentbench},
SciCode~\cite{scicode}, MatTools~\cite{mattools},
ChemCrow~\cite{chemcrow}, Coscientist~\cite{coscientist}), and more recently to
multi-step scientific workflows (Spider2-V~\cite{spider2v},
ScienceBoard~\cite{scienceboard}, LabBench~\cite{labbench}).
ChemCrow and Coscientist emphasize tool-augmented chemical reasoning rather than
real desktop manipulation, while ScienceBoard uses realistic scientific environments
but is not centered on the materials tool chain. Most prior benchmarks evaluate
structured states, scripts, or logs; \textsc{MatToolBench} targets software whose
state is often exposed only visually.

\section{MatToolBench Environment}
\begin{figure*}[!t]
    \centering
    \includegraphics[width=0.78\textwidth]{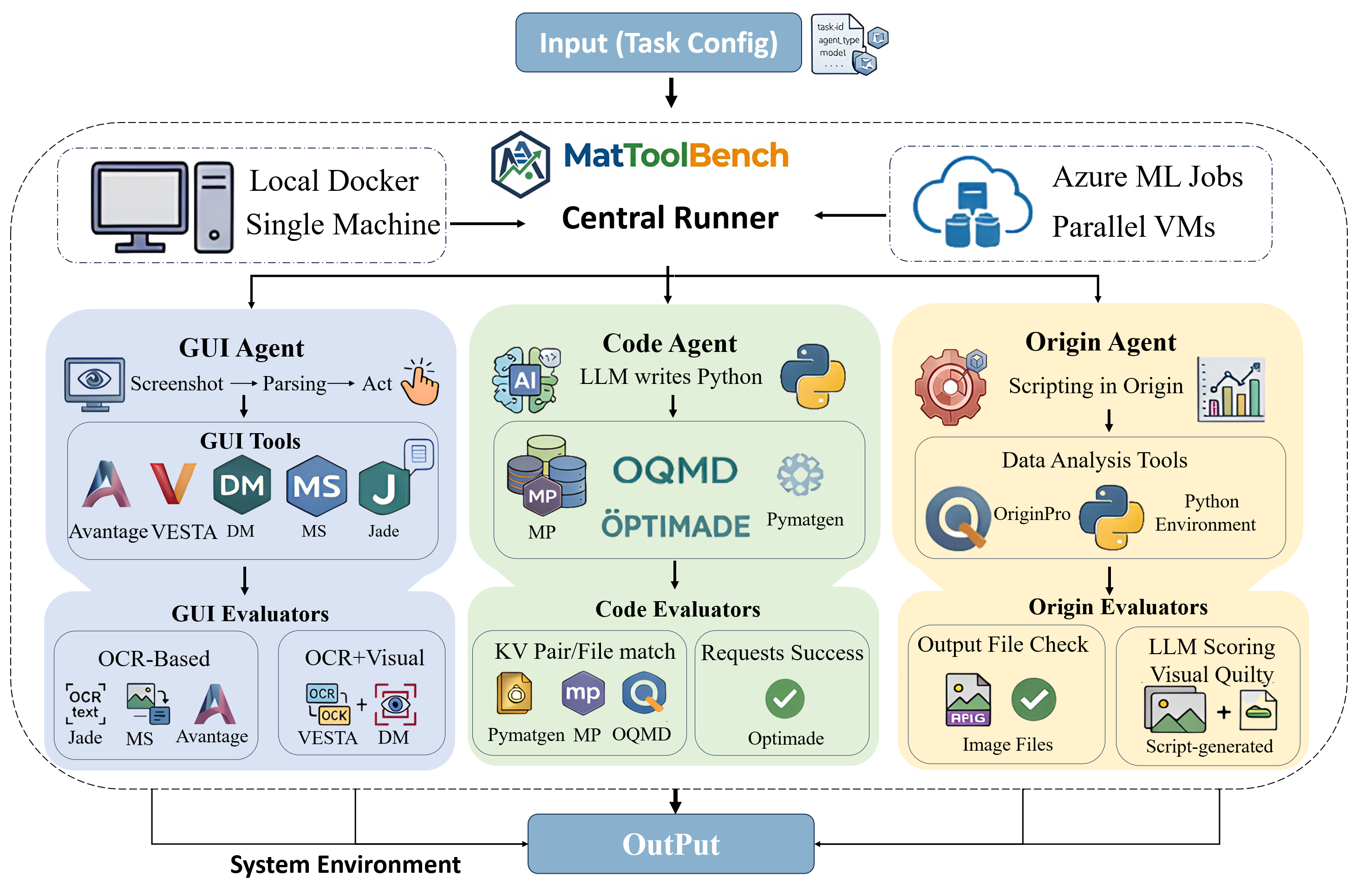}
    \caption{Overview of the \textsc{MatToolBench} framework.
    Tasks are dispatched by a central runner to one of three specialized agents:
    the \emph{GUI Agent} operates materials science desktop applications
    (e.g., JADE, VESTA, Avantage) via screenshot perception and coordinate-based interaction;
    the \emph{Code Agent} generates Python scripts to query materials database
    APIs (e.g., MP, OQMD, OPTIMADE);
    and the \emph{Origin Agent} combines GUI interaction with OriginPro script
    generation for experimental data plotting.
    All agents execute inside a Windows~11 VM hosted within a Docker container,
    communicating through a Flask-based control server.
    Task-specific evaluators assess the final environment state after each episode.}
    \label{fig3}
\end{figure*}

As illustrated in Figure~\ref{fig3}, the \textsc{MatToolBench} environment is built on a
Windows~11 virtual machine and exposes three interaction modalities (GUI
operation, Python code generation, and OriginPro script authoring) through
a unified evaluation interface.

\subsection{Task Definition}
Task execution is formalized as a POMDP
$\langle \mathcal{S}, \mathcal{O}, \mathcal{A}, \mathcal{T}, \mathcal{R} \rangle$.
At each step $t$ the agent receives observation $o_t$ (screenshot/execution feedback),
takes action $a_t$, and transitions to $s_{t+1}$.
Policy $\pi$ conditions on the full history
$m_t = \{g, o_0, a_0, \ldots, o_t\}$ where $g$ is the fixed natural-language instruction.
Episodes terminate on: \texttt{DONE} (agent declares success),
\texttt{FAIL} (agent declares infeasible), or step-budget exhaustion (default 50 steps). A \texttt{WAIT} action is available for slow GUI responses.

\noindent\textbf{Environment Initialization.}
Each task ships with an initialization sequence (\texttt{config})
that is executed automatically by the environment before the episode begins,
bringing the VM to a deterministic starting state.
Initialization steps cover application launch (\texttt{launch}), file transfer
(\texttt{vm\_file}), in-software pre-configuration (\texttt{execute}), and
GUI readiness wait (\texttt{sleep}), ensuring every episode starts from a
reproducible initial state.

\noindent\textbf{Partial-Credit Score Design.}
\textsc{MatToolBench} decomposes each task into multiple verifiable
\textbf{scoring sub-criteria} (pts).
At episode end, the evaluator independently scores each sub-criterion
and produces a fine-grained score vector.
Compared to a binary success label, this decomposed score exposes which
intermediate requirements are satisfied and supports more interpretable
failure analysis across multi-step workflows.
Appendix~A summarizes the JSON task schema.

\subsection{Observation Space}
\noindent\textbf{GUI-Based Tasks.}
Each observation $o_t$ for desktop-application tasks comprises the task instruction
$g$, a full-desktop screenshot $v_t$ (1920$\times$1080), foreground-window metadata,
and clipboard content. We use screenshot-based observations in all reported
experiments, since this is the most portable interface across legacy scientific
GUI software whose accessibility metadata is often incomplete or unavailable.

\noindent\textbf{Code Tasks.}
For database tasks $o_t$ reduces to text-only execution
feedback (\texttt{stdout}, \texttt{stderr}, return code); no visual signal is provided.

\noindent\textbf{Origin Tasks.}
Origin tasks combine GUI interaction with script authoring, so the observation mirrors
that of GUI-based tasks: the full-desktop screenshot $v_t$ is used both for GUI
operations (e.g., clicking menus, creating new files) and for inspecting the current
OriginPro workspace to verify that the generated plot meets requirements.

\subsection{Action Space}
\noindent\textbf{GUI Actions.}
GUI tasks use a hierarchical \texttt{Computer} interface with five sub-modules:
\texttt{Mouse} for coordinate-based clicking, scrolling, and dragging, \texttt{Keyboard} for input
and hotkeys, \texttt{Clipboard} for copy-paste, \texttt{OS} for program launch,
and \texttt{WindowManager} for fuzzy-match application switching.
Actions are transmitted as Python code strings executed via \texttt{exec()} on the VM.
The full API is in Appendix~C, Table~\ref{tab:computer_api}.

\noindent\textbf{Code and Origin Actions.}
Code tasks reduce to a generate-and-correct loop: the agent writes a complete Python
script per step, executed in an isolated virtual environment per domain;
\texttt{stdout}/\texttt{stderr}/return-code are fed back for self-correction.
Origin tasks inject scripts into OriginPro's Code Builder via clipboard paste and \texttt{F5},
combining code generation with GUI navigation.

\subsection{Execution Environment}
The environment has two layers: an \textbf{outer} Linux Docker container hosting agent
logic, screenshot capture, and evaluation services; and an \textbf{inner} 1920$\times$1080 Windows~11
VM pre-installed with all target applications and per-domain isolated Python venvs.
The code agent selects the appropriate venv by task category and activates it via
\texttt{activate.bat} before running the generated script.
A Flask server in the VM serves as the sole control interface, with external API calls
routed through tinyproxy. The framework exposes a unified evaluation interface and
supports concurrent multi-worker evaluation.
Large-scale experiments are run on Azure ML using \texttt{Standard\_D8\_v3} compute instances
(see Appendix~D for infrastructure details).

\section{The \textsc{MatToolBench} Benchmark}
\subsection{Task Statistics}

\textsc{MatToolBench} comprises \textbf{204 tasks} spanning \textbf{10 domain-specific
tools}, organized into three agent types: GUI-based (JADE, Avantage, VESTA~\cite{vesta}, DM, MS),
Origin-based (Origin), and code-based (MP~\cite{materialsproject,materialsprojectapi}, OQMD~\cite{oqmd}, OPTIMADE~\cite{optimade}, Pymatgen~\cite{pymatgen}). Brief descriptions of the ten software tools are provided in Appendix~O.
Table~\ref{tab:task-stats} summarizes the task distribution, difficulty split,
and scoring sub-criteria across domains.
Appendix~B, Figure~\ref{fig:task_distribution} illustrates the sub-category distribution within each domain.
The complete task list with instructions and difficulty levels is provided in
Appendix~E.
Mixed cross-tool diagnostic tasks are summarized in Appendix~F.
To support realistic evaluation, the VM is pre-loaded with authentic experimental
data files for each GUI and Origin domain (e.g., real XPS spectra, TEM images, XRD
patterns, and crystal structure files); the full inventory is listed in
Appendix~G.
Before deployment in the VM, task workflows were cross-checked against vendor
tutorials, public operation videos, and domain documentation. Similar low-level
actions may recur across tasks, but the target artifacts, software states, and
scoring criteria differ.

\noindent\textbf{Scoring Scheme.}
Each task is decomposed into one or more \textbf{scoring sub-criteria} (pts),
each corresponding to a distinct intermediate or final state the agent must
correctly achieve. The per-task score equals the fraction of sub-criteria satisfied,
enabling partial credit on multi-step tasks. A task is counted as a full success
(SR\,=\,1) only when all sub-criteria are met.

\noindent\textbf{Difficulty Levels.}
Tasks are stratified into three levels according to the expected number of
operations and scoring sub-criteria.
For GUI tasks, the number of scoring sub-criteria is the main proxy:
\textbf{Easy} (1--3\,pts), \textbf{Medium} (4\,pts), and \textbf{Hard} (5--7\,pts).
For Origin and code tasks, where scoring dimensions are more standardized,
E/M/H labels reflect workflow complexity and are listed in Appendix~E.
Mixed tasks use the same point-count proxy because each stage is evaluated through
explicit artifact-level sub-criteria.

\begin{table}[!t]
  \centering
  \footnotesize
  \renewcommand{\arraystretch}{1.02}
  \setlength{\tabcolsep}{2.7pt}
  \captionsetup{justification=raggedright,singlelinecheck=false}
  \caption[Benchmark task statistics.]{Benchmark task statistics.\\[-1pt]
  E/M/H: easy\,/\,medium\,/\,hard task counts. Mixed tasks are reported separately
  as cross-tool diagnostics; total sub-criteria are 454 single-tool criteria plus
  30 mixed diagnostic criteria.}
  \label{tab:task-stats}
  \resizebox{\linewidth}{!}{%
  \begin{tabular}{llcccr}
  \toprule
  \textbf{Cat.} & \textbf{Domain} & \textbf{\#} & \textbf{E/M/H}
    & \textbf{Avg} & \textbf{Tot.} \\
  \midrule
  \multirow{5}{*}{GUI}
    & Avantage & 20 & 14/2/4            & 3.25 & 65 \\
  & DM       & 20 & 13/5/2            & 3.35 & 67 \\
  & JADE     & 20 & 15/5/0            & 2.95 & 59 \\
  & MS       & 20 & \phantom{0}6/11/3 & 4.10 & 82 \\
  & VESTA    & 20 & 13/2/5            & 3.45 & 69 \\
  \midrule
  Origin & Origin & 16 & \phantom{0}0/7/9 & 2.00 & 32 \\
  \midrule
  Code   & \makecell[l]{MP/OQMD/\\OPTIMADE/Pymatgen}
	                   & 80 & 37/41/2       & 1.00 & 80 \\
  \midrule
  Mixed  & Mixed   &  8 & \phantom{0}3/4/1 & 3.75 & 30 \\
  \midrule
  \textbf{Total} & & \textbf{204} & & & \textbf{484} \\
  \bottomrule
  \end{tabular}%
  }
\end{table}

\subsection{Evaluation Protocol}

Evaluation follows a two-stage pipeline after each episode: a domain-specific
\textbf{getter} extracts the task-relevant result, and a \textbf{metric function}
scores it against a reference. GUI results are extracted via OCR or pixel analysis;
code results are written to files and compared against a gold-standard directory;
Origin results use deterministic export checks plus an optional Gemini-3.1-pro visual judge.
Metric details appear in Appendices~H and M, especially
Tables~\ref{tab:eval-gui-getters} and \ref{tab:eval-metrics}.
A step-by-step evaluation walkthrough is in Appendix~L;
pixel-detection method examples are in Appendix~M.

\noindent\textbf{Metrics.}
We report \textbf{Score} (normalized fraction of sub-criteria satisfied, enabling
partial credit) and \textbf{SR} (success rate: all sub-criteria met).
Efficiency is captured by \textbf{Steps} (mean steps across all episodes),
\textbf{Steps*} (mean steps on successful episodes only),
and \textbf{FTR} (false termination rate: fraction of agent-terminated episodes where
SR\,=\,0). All step counts are out of a 50-step maximum; fewer steps indicate higher efficiency.

\begin{table*}[!t]
  \centering
  \small
  \renewcommand{\arraystretch}{1.12}
  \setlength{\tabcolsep}{3.5pt}
  \caption{Accuracy results on \textsc{MatToolBench}.
  Sc.~(Score): normalized task score (\%) averaged over sub-criteria;
  SR: success rate (\%, all sub-criteria satisfied).
  \textbf{Bold}: best average metric within each category;
  \textbf{\textit{bold italic}}: best overall average metric.}
  \label{tab:results-accuracy}
  \resizebox{0.97\textwidth}{!}{%
  \begin{tabular}{ll cc cc cc cc cc cc cc}
  \toprule
  & & \multicolumn{2}{c}{\makecell{\textbf{Doubao}\\[-1pt]\textbf{seed-1-8}}}
      & \multicolumn{2}{c}{\makecell{\textbf{Kimi}\\[-1pt]\textbf{k2.5}}}
      & \multicolumn{2}{c}{\makecell{\textbf{Claude}\\[-1pt]\textbf{sonnet-4.6}}}
      & \multicolumn{2}{c}{\makecell{\textbf{\textit{GPT}}\\[-1pt]\textbf{\textit{5.4}}}}
      & \multicolumn{2}{c}{\makecell{\textbf{Qwen3-VL}\\[-1pt]\textbf{235B}}}
      & \multicolumn{2}{c}{\makecell{\textbf{Qwen3-VL}\\[-1pt]\textbf{32B}}}
      & \multicolumn{2}{c}{\makecell{\textbf{Qwen3-VL}\\[-1pt]\textbf{8B}}} \\
  \cmidrule(lr){3-4}   \cmidrule(lr){5-6}   \cmidrule(lr){7-8}
  \cmidrule(lr){9-10} \cmidrule(lr){11-12} \cmidrule(lr){13-14} \cmidrule(lr){15-16}
  \textbf{Cat.} & \textbf{Domain}
    & \textbf{Sc.} & \textbf{SR}
    & \textbf{Sc.} & \textbf{SR}
    & \textbf{Sc.} & \textbf{SR}
    & \textbf{Sc.} & \textbf{SR}
    & \textbf{Sc.} & \textbf{SR}
    & \textbf{Sc.} & \textbf{SR}
    & \textbf{Sc.} & \textbf{SR} \\
  \midrule
  \multirow{6}{*}{GUI}
    & Avantage & 44.6 & 35.0 & 32.3 & 20.0 & 41.3 & 35.0 & 32.3 & 15.0 & 30.8 & 10.0 & 26.2 & 15.0 & 18.5 & 15.0 \\
  & JADE     & 54.2 & 25.0 & 50.8 & 25.0 & 57.6 & 25.0 & 50.8 & 20.0 & 45.8 & 10.0 & 37.3 & 10.0 & 37.3 & 5.0 \\
  & DM       & 56.7 & 20.0 & 52.2 & 20.0 & 40.3 & 20.0 & 50.7 & 20.0 & 52.2 & 10.0 & 6.0 & 0.0 & 38.8 & 10.0 \\
  & MS       & 47.6 & 15.0 & 35.4 & 10.0 & 52.4 & 20.0 & 56.1 & 30.0 & 43.9 & 0.0 & 26.8 & 0.0 & 20.7 & 0.0 \\
  & VESTA    & 52.2 & 20.0 & 59.4 & 25.0 & 58.0 & 25.0 & 65.2 & 20.0 & 52.2 & 10.0 & 29.0 & 10.0 & 30.4 & 10.0 \\
  \cmidrule(lr){2-16}
  & \textit{Avg.} & \textbf{52.1} & 24.0 & 46.0 & 20.0 & 49.9 & \textbf{25.0} & 51.0 & 21.0 & 45.0 & 8.0 & 25.1 & 7.0 & 29.1 & 8.0 \\
  \midrule
  Origin & OriginPro & 11.7 & 12.5 & 22.5 & 25.0 & 53.1 & 56.3 & \textbf{63.8} & \textbf{68.8} & 6.3 & 6.3 & 5.3 & 6.3 & 0.0 & 0.0 \\
  \midrule
  \multirow{5}{*}{Code}
    & Pymatgen & 30.0 & 30.0 & 15.0 & 15.0 & 25.0 & 25.0 & 35.0 & 35.0 & 35.0 & 35.0 & 29.4 & 29.4 & 15.0 & 15.0 \\
  & MP       & 25.0 & 25.0 & 22.5 & 20.0 & 21.9 & 15.0 & 17.5 & 15.0 & 25.0 & 25.0 & 10.0 & 10.0 & 10.0 & 10.0 \\
  & OQMD     & 33.2 & 10.0 & 50.3 & 30.0 & 60.8 & 50.0 & 58.2 & 45.0 & 26.2 & 10.0 & 11.5 & 0.0 & 18.4 & 0.0 \\
  & OPTIMADE & 70.0 & 70.0 & 80.0 & 80.0 & 75.0 & 75.0 & 85.0 & 85.0 & 20.0 & 20.0 & 30.0 & 30.0 & 10.0 & 10.0 \\
  \cmidrule(lr){2-16}
  & \textit{Avg.} & 39.6 & 33.8 & 42.0 & 36.2 & 45.7 & 41.3 & \textbf{48.9} & \textbf{45.0} & 26.6 & 22.5 & 20.2 & 17.4 & 13.4 & 8.8 \\
  \midrule
  Mixed & Mixed & 50.0 & 12.5 & \textbf{60.0} & \textbf{37.5} & 56.7 & 25.0 & 50.0 & 25.0 & 50.0 & 25.0 & 33.3 & 25.0 & 53.3 & \textbf{37.5} \\
  \midrule
    \multicolumn{2}{l}{\textbf{Overall Avg.}}
    & 43.4 & 26.0 & 43.1 & 27.5 & 48.8 & 33.8 & \textbf{\textit{51.2}} & \textbf{\textit{34.3}} & 34.9 & 14.2 & 21.9 & 11.7 & 21.6 & 8.8 \\
  \bottomrule
  \end{tabular}%
  }
  \end{table*}

\section{Experiments}
\label{sec:experiments}
\subsection{Experimental Setup}
\label{sec:setup}
\vspace{-2pt}
\paragraph{Models.}
We evaluate seven state-of-the-art multimodal models spanning five commercial providers:
\textbf{Doubao-seed-1-8} (ByteDance), \textbf{Kimi-k2.5} (Moonshot), \textbf{Claude-sonnet-4.6} (Anthropic),
\textbf{GPT-5.4} (OpenAI), \textbf{Qwen3-VL-235B} and \textbf{Qwen3-VL-32B} (Alibaba),
and \textbf{Qwen3-VL-8B} (Alibaba).
All models are accessed via API.
For GUI and Origin tasks, each model is wrapped in the \textsc{NaviAgent} framework
with raw screenshots as the sole visual input and a system prompt tailored to the agent type.
For code tasks, no visual input is provided.
The maximum step budget is 50 for all task types.

\vspace{-2pt}
\paragraph{Reproducibility Details.}
All reported evaluations use the same VM image, Docker environment, task
configuration files, input artifacts, gold/reference files, Origin reference
scripts, decoding settings, and post-action wait time. Full runtime and
reproducibility details are provided in Appendix~D.

\vspace{-2pt}
\paragraph{Execution Protocol.}
To reduce avoidable run-to-run variation in GUI evaluation, every episode starts
from a reset VM state with fixed screen resolution, identical input files, the same
task initialization script, and the same 50-step budget. Worker instances use
copy-on-write VM storage, so parallel runs do not share application state. All
success checks are performed after the episode terminates by deterministic getters
or by the validated Origin quality judge described below.

\subsection{Main Results}
\label{sec:main-results}

Table~\ref{tab:results-accuracy} reports Score and SR for all seven models across GUI, Origin,
Code, and the diagnostic Mixed row.
The overall performance level is consistently low, with the best-performing models
achieving only 24--25\% SR on GUI tasks and 41--45\% SR on code tasks.
GPT-5.4 obtains the best overall average, followed by Claude-sonnet-4.6.
Compared with reported success rates above 70\% on general desktop benchmarks
such as OSWorld, the best \textsc{MatToolBench} GUI SR reaches only 25.0\%,
providing diagnostic evidence of a domain-transfer gap (Appendix~Q). The gap is compound: models
often complete isolated sub-steps, but fail to maintain domain-correct state
across visual interfaces, scripts, and files.
Because failure modes differ across GUI grounding, API usage, and Origin
scripting, we report both category averages and domain-level breakdowns.

\vspace{-2pt}
\paragraph{GUI Tasks.}
On GUI tasks, Doubao-seed-1-8 achieves the highest average Score of 52.1\%
and a near-best average SR of 24.0\%, while Claude-sonnet-4.6 achieves the
highest average SR of 25.0\%.
GPT-5.4 remains competitive, with the second-highest average Score of 51.0\%
and an average SR of 21.0\%.
Models without adequate GUI grounding (notably Qwen3-VL-32B at 7.0\% avg.\ SR)
struggle on tasks requiring precise multi-step interaction sequences.

Among individual domains, \textbf{MS} (Materials Studio) has the lowest average SR across models,
while \textbf{DM} remains uniformly difficult, with no model exceeding 20\% SR.
These results reflect highly nested menu hierarchies and implicit feedback after operations.
In contrast, \textbf{VESTA} and \textbf{JADE} show relatively higher scores,
likely because these tools provide more explicit visual confirmation of state changes.

A notable gap exists between Score and SR across all GUI domains:
models routinely complete a fraction of scoring sub-criteria (Score\,$\approx$50\%)
while rarely satisfying all of them simultaneously (SR\,$\approx$20\%).
This pattern confirms the utility of the partial-credit scoring mechanism,
revealing that models \emph{do} make meaningful progress on complex tasks
even when they fall short of full completion.

\begin{table*}[!t]
\centering
\caption{Ablation conditions for each task type. \textbf{Env Setup}
indicates whether the task environment automatically opens Code Builder (Alt+4)
before the episode starts. In \texttt{no\_hint} mode this step is suppressed,
requiring the agent to discover the workflow tool independently.}
\label{tab:ablation}

\resizebox{\textwidth}{!}{%
\begin{tabular}{llllcccc}
\toprule
\textbf{Type} & \textbf{Parameter} & \textbf{Value} & \textbf{Condition}
& \textbf{Script} & \textbf{Hint} & \textbf{Env Setup} & \textbf{Injected Content} \\
\midrule

& & \texttt{hint}      & full     & — & \checkmark & — & GUI workflow instructions \\
\multirow{-2}{*}{GUI}
& \multirow{-2}{*}{\texttt{gui\_hint\_mode}}
& \texttt{no\_hint} & baseline & — & $\times$ & — & generic guidelines only \\

\midrule

& & \texttt{hint}      & full     & — & \checkmark & — & API examples \& docs \\
\multirow{-2}{*}{Code}
& \multirow{-2}{*}{\texttt{code\_hint\_mode}}
& \texttt{no\_hint} & baseline & — & $\times$ & — & generic guidelines only \\

\midrule

& & \texttt{script}+\texttt{hint}
& full
& \checkmark
& \checkmark
& \checkmark
& template + workflow \\

& &
\texttt{no\_script}+\texttt{hint}
& partial
& $\times$
& \checkmark
& \checkmark
& workflow only \\

\multirow{-3}{*}{Origin}
&
\multirow{-3}{*}{\makecell[l]{\texttt{origin\_mode} $\times$\\
\texttt{origin\_hint\_mode}}}
&
\texttt{no\_script}+\texttt{no\_hint}
&
baseline
&
$\times$
&
$\times$
&
$\times$
&
generic guidelines only \\

\bottomrule
\end{tabular}
}
\end{table*}

\vspace{-2pt}
\paragraph{Code Tasks.}
Code tasks show higher performance overall, consistent with LLMs' strength in structured
text generation.
GPT-5.4 achieves the best average SR of 45.0\%, followed by Claude-sonnet-4.6 (41.3\%).
Performance varies markedly across databases.
\textbf{OPTIMADE} yields the highest SRs (20--85\%), partly because its
instructions can return multiple valid provider-dependent results. We therefore
evaluate whether the script issues a valid query and saves valid output values,
rather than exact-matching a single gold file.
\textbf{MP} is more challenging (10--25\% SR) as it requires accurate use of
the \texttt{mp-api} Python client with evolving API conventions;
results are validated by key-value pair matching within a 0.1\% tolerance.
\textbf{OQMD} tasks occupy a middle ground (0--50\% SR), with large performance gaps
between frontier models (Claude 50\%, GPT 45\%) and smaller ones (Qwen3-VL-8B 0\%);
outputs are compared against reference key-value pairs.
\textbf{Pymatgen} tasks use line-by-line file comparison against a gold-standard reference.

The smaller Qwen3-VL models (32B, 8B) underperform substantially on code tasks
despite reasonable GUI scores, suggesting that reliable scientific API usage
requires deeper domain-specific knowledge that scales differently from general
GUI interaction.

\vspace{-2pt}
\paragraph{Origin Tasks.}
Origin tasks show the sharpest performance bifurcation. GPT-5.4 leads with 68.8\% SR, followed by Claude-sonnet-4.6 at 56.3\%, while all other models remain below 25\%; the Qwen series reaches at most 6.3\%, and Qwen3-VL-8B scores 0\%. For Origin tasks, SR is determined by deterministic file-existence checks: \texttt{exact\_match} verifies that the required \texttt{.png} is exported to the specified path. Correctness, completeness, and aesthetics are assessed separately by Gemini-3.1-pro as secondary quality scores rather than SR criteria. We exclude the LLM-based quality judge from SR to preserve a deterministic and reproducible definition of task success. Models must successfully use OriginPro Code Builder, generate a valid Python script, and save the file before visual judging; otherwise both scores are zero. Mid-tier models such as Kimi-k2.5 (25.0\%) and Doubao-seed-1-8 (12.5\%) show mostly binary outcomes, confirming that failures are concentrated in script execution rather than visual assessment.

\vspace{-2pt}
\paragraph{Mixed Cross-Tool Tasks.}
The eight Mixed tasks are reported only as a diagnostic row for artifact handoff
across code$\to$code, code$\to$GUI, GUI$\to$Origin, and GUI$\to$GUI stages.
We do not use this small split as a headline ranking.
Their scores are relatively high because code-first workflows can earn partial
credit by producing verifiable intermediate files, while full SR remains lower
when GUI-export artifacts cannot be located, parsed, or reused by the downstream
tool.

\noindent\textbf{Efficiency.}
Appendix~J, Table~\ref{tab:results-efficiency} reports Steps, Steps* and FTR.
GUI tasks consume approximately 35--42 steps on average, reflecting frequent
exploratory actions, while code tasks use only approximately 1--4 steps since each step is a
complete script execution.
High FTR values (40--70\%) on GUI tasks reveal a persistent premature-termination
failure mode across all models.
Qualitative trajectory inspection identifies four recurring failure patterns:
GUI grounding errors, premature termination, deadlock loops, and domain knowledge gaps,
each illustrated with real trajectory screenshots in Appendix~P.

\subsection{Ablation Study}
\label{sec:ablation}

Table~\ref{tab:ablation} summarizes the ablation conditions. We ablate the
effect of domain-specific prompt hints (\texttt{hint\_mode}) by comparing the
full agent prompt against a \texttt{no\_hint} baseline that omits all
domain-specific workflow instructions, evaluated on Doubao-seed-1-8 and GPT-5.4.
The agent prompt structure for each modality is detailed in Appendix~N.

\begin{figure}[!t]
\centering
\includegraphics[width=\linewidth]{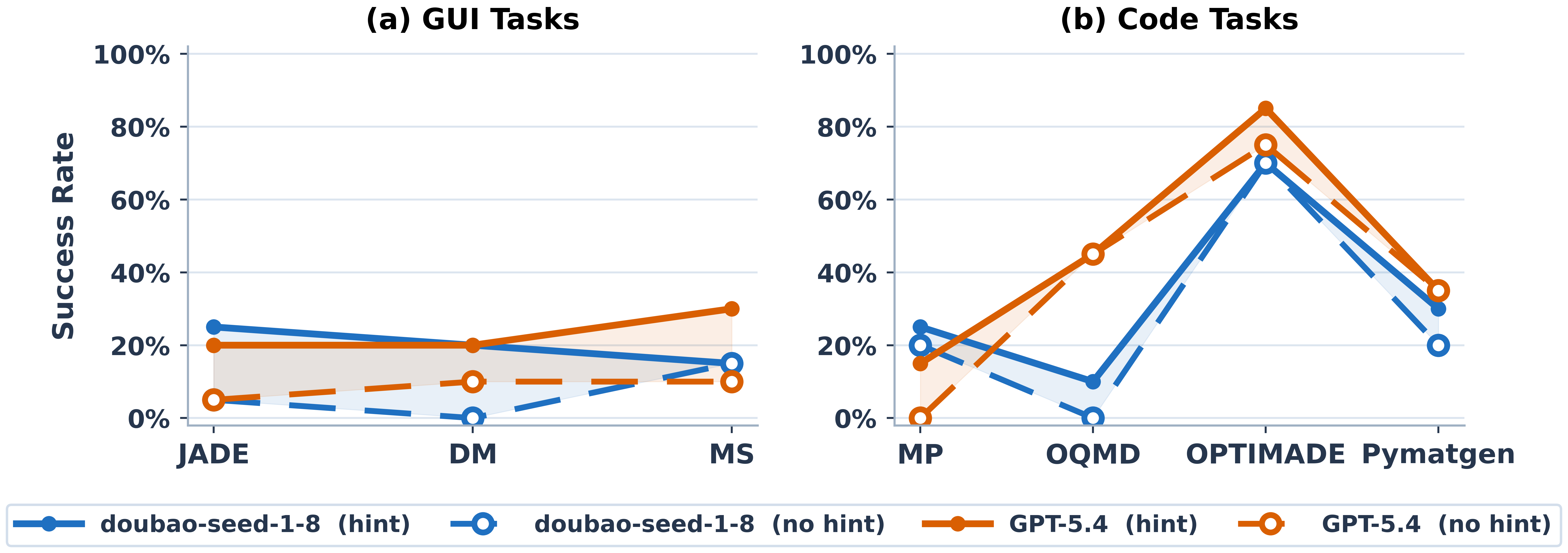}
\caption{Ablation study: SR comparison between \textbf{hint} (solid) and
\textbf{no\_hint} (dashed) prompt modes for Doubao-seed-1-8 (blue) and GPT-5.4 (orange).
Left: GUI tasks. Right: Code tasks.
Shaded bands indicate the gap between hint and no\_hint conditions.}
\label{fig:ablation}
\end{figure}

\vspace{-2pt}
\paragraph{GUI Tasks.}
Figure~\ref{fig:ablation} (left) shows that removing hints consistently degrades GUI performance.
Doubao-seed-1-8 drops from 25.0\% to 5.0\% SR on JADE and from 20.0\% to 0.0\% on DM;
GPT-5.4 shows similar trends.
Avantage and VESTA are not included in the ablation because their workflows are
sufficiently self-evident from the interface and do not require additional guidance.
The main failures are tool-specific startup traps: JADE opens a reference-pattern
dialog that blocks task-file loading, while DM starts with a floating panel that can
occlude the imported filename and disrupt file-import verification.

\vspace{-2pt}
\paragraph{Code Tasks.}
Figure~\ref{fig:ablation} (right) shows a more nuanced picture.
For OPTIMADE, both models retain nearly the same SR with and without hints
(Doubao-seed-1-8: 70\% $\to$ 70\%; GPT-5.4: 85\% $\to$ 75\%),
suggesting that its standardized query syntax is sufficiently covered in pretraining data.
However, on MP and OQMD, removing hints causes larger drops
(GPT-5.4 MP: 15\% $\to$ 0\%), because the code agent prompt includes one-shot API usage
examples; without these examples, models must reconstruct library-specific conventions
from memory, which is less reliable for evolving or niche APIs.
This asymmetry suggests that hints mainly help when the tool convention is niche or
unstable rather than broadly standardized.

\vspace{-2pt}
\paragraph{Origin Tasks.}
Origin tasks present a unique challenge: the standard workflow requires the agent to
open OriginPro's built-in Code Builder editor via \texttt{Alt+4} and execute a
Python script to generate figures programmatically.
We ablate three levels of workflow support using the conditions in
Table~\ref{tab:ablation}. Removing only the template script mostly affects visual
quality, because the agent still knows to use Code Builder; removing the hint as
well often pushes the agent toward GUI-only exploration, reducing both file
generation and aesthetic quality.

\begin{figure}[!t]
  \centering
  \includegraphics[width=0.95\linewidth]{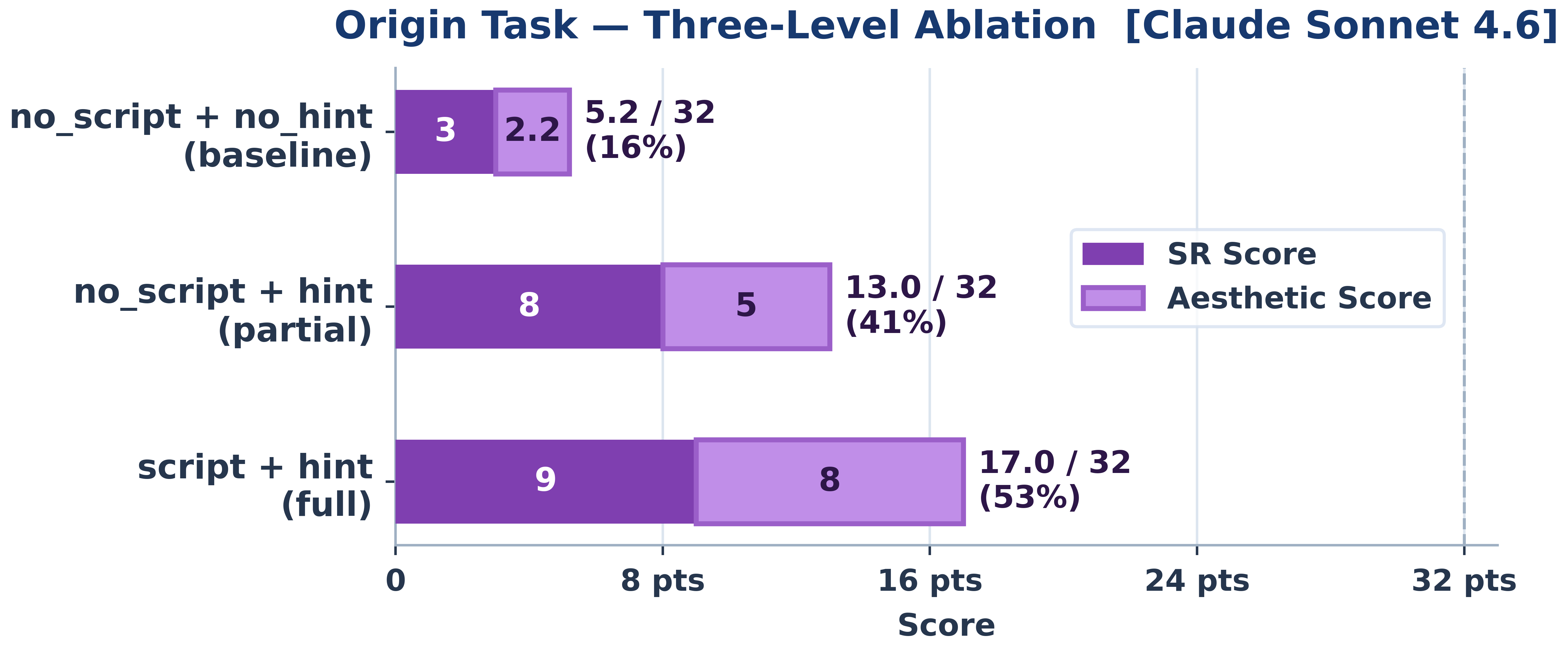}
  \caption{Origin three-level ablation for Claude Sonnet 4.6.
  Bars show deterministic success points (dark purple, max 16~pts) and
  Aesthetic Score (light purple, max 16~pts) for each condition.
  Total scores and percentages are annotated at bar ends.}
  \label{fig:ablation-origin}
\end{figure}

Figure~\ref{fig:ablation-origin} confirms this: deterministic success drops
from 9 to 8 and then to 3 at baseline — a \textbf{67\% reduction}.
Aesthetic Score follows the same trend (8 $\to$ 5 $\to$ 2.2), and the combined
total score falls from 17/32 (53\%) to 5.2/32 (16\%) at baseline.
Representative output figures under both conditions are shown in
Appendix~K (Figure~\ref{fig:origin-examples}).

\section{Evaluator Reliability}
\label{sec:reliability}
\subsection{GUI Evaluator}
For each GUI domain, we apply task-specific evaluators to the collected
trajectory final states, forming an $n\times n$ cross-evaluation matrix
($n=20$). Each final state is evaluated against every task configuration in the
same domain, measuring both matched detection and cross-task false positives.
We report \textbf{Score} agreement over sub-criterion labels and \textbf{SR}
agreement over task-level success labels; Table~\ref{tab:overall} summarizes
the results, with per-task details in Appendix~I.

\begin{table}[t]
\centering
\caption{GUI evaluator reliability under an $n\times n$ cross-evaluation protocol ($n=20$). DM uses the filtered audit described in Appendix~I.}
\label{tab:overall}
\scriptsize
\setlength{\tabcolsep}{2pt}
\renewcommand{\arraystretch}{1.05}
\resizebox{\linewidth}{!}{%
\begin{tabular}{lcccccccc}
\toprule
& \multicolumn{4}{c}{\textbf{Score}} & \multicolumn{4}{c}{\textbf{SR}} \\
\cmidrule(lr){2-5} \cmidrule(lr){6-9}
\textbf{Domain} & Acc. & Prec. & Rec. & F1 & Acc. & Prec. & Rec. & F1 \\
\midrule
JADE      & 1.00 & 1.00 & 0.99 & 0.99 & 1.00 & 1.00 & 1.00 & 1.00 \\
MS        & 1.00 & 0.99 & 1.00 & 0.99 & 1.00 & 1.00 & 1.00 & 1.00 \\
Avantage  & 0.99 & 0.98 & 0.98 & 0.98 & 1.00 & 0.98 & 1.00 & 0.99 \\
VESTA     & 0.98 & 0.94 & 0.99 & 0.97 & 0.99 & 0.94 & 1.00 & 0.97 \\
DM        & 0.99 & 0.99 & 0.94 & 0.96 & 1.00 & 0.98 & 1.00 & 0.99\\
\bottomrule
\end{tabular}%
}
\end{table}

\subsection{Code Evaluator}
Code tasks use deterministic checks whenever possible: Pymatgen uses reference-file comparison, while MP/OQMD use tolerance-based value matching. The evaluator also checks that generated outputs are parseable, valid, and task-specific, so malformed files or unrelated values do not receive credit. For provider-dependent OPTIMADE, we require constrained queries with saved valid parsed output values rather than a single gold file.

\subsection{Origin Aesthetic Evaluator}
Origin SR is determined by deterministic file-existence \texttt{exact\_match}
checks on the expected exported figure path; the LLM judge provides only a
secondary aesthetic score.
We validate four judges on 20 human-rated conditions and 43 generated Origin figures.
Detailed statistics are reported in Appendix~I, Table~\ref{tab:origin-judge-validation}.
Qwen3-VL-235B has a slightly higher Pearson correlation than Gemini-3.1-pro,
but the Williams dependent-correlation test shows no significant difference
($\Delta r=0.043$, $t=0.582$, $p=0.569$). Since Qwen3-VL-235B also gives more
score-saturated ratings, we use Gemini-3.1-pro as the more conservative
secondary judge to reduce score inflation. This choice does not affect any
reported SR.

\begin{figure}[!t]
  \centering
  \includegraphics[width=0.98\linewidth]{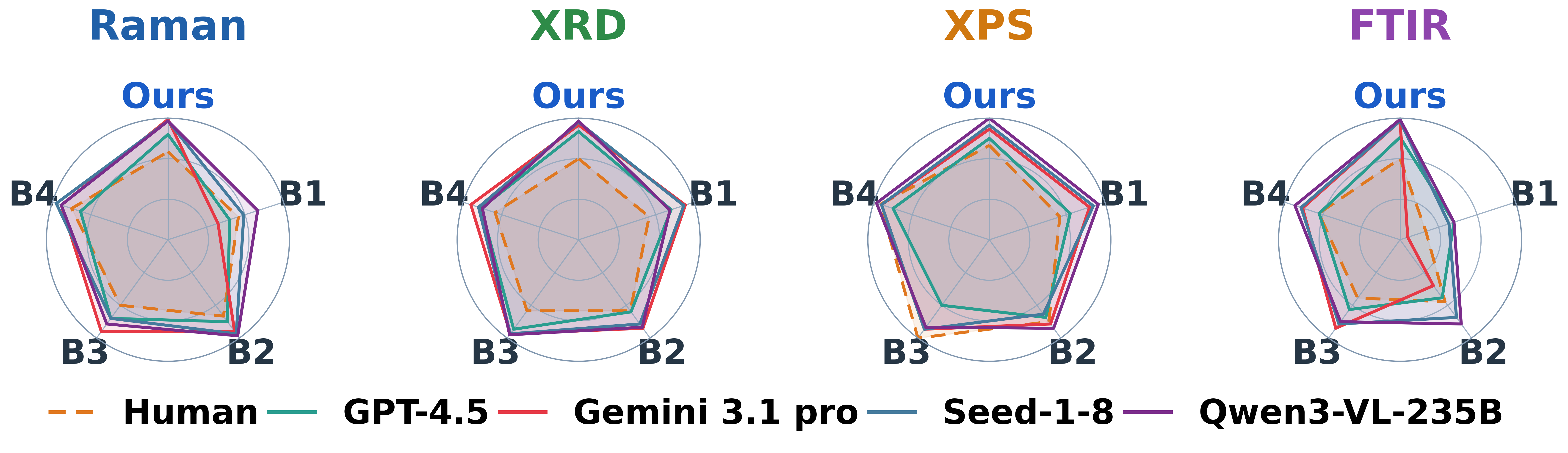}
  \caption{Radar chart comparing Origin figure quality across four task types
    (Raman, XRD, XPS, FTIR). Subplots are task types; axes are visual
    correctness, aesthetic quality, and task completeness (1--5). Dashed lines
    are means from 9 humans; solid lines are four LLM judges.}
  \label{fig:radar-origin-quality}
\end{figure}

\FloatBarrier
\section{Conclusion}
We introduced \textsc{MatToolBench}, a real-environment benchmark with 204 tasks across 10 materials-science tools, spanning GUI, OriginPro, and code workflows. Results provide diagnostic evidence of a domain-transfer gap from general benchmarks to scientific software: the best models achieve only 25\% SR on GUI tasks and 45\% on code tasks. The gap between partial scores and full success reveals end-to-end failures. Ablations show that workflow support is crucial, especially when interfaces and APIs are sparsely represented in pretraining. Deterministic success criteria and validated quality judges enable reliable, interpretable evaluation across workflows. \textsc{MatToolBench} offers a practical testbed for improving scientific agents. Limitations and broader impacts are discussed in Appendix~R.

\noindent\textbf{Open-source repository:}\\
\url{https://github.com/meiwu5/MatToolBench}.

\bibliography{ref}
\appendix
\setcounter{section}{0}
\renewcommand{\thesection}{\Alph{section}}
\clearpage
\section{Appendix A. Task Configuration Schema}
\label{sec:task-config-schema}

\noindent Each task is stored as a JSON file in a per-domain directory. The schema
separates the agent-facing instruction, VM initialization, runtime metadata, and
sub-criterion evaluators, making task execution reproducible and assignment
explicit. Table~\ref{tab:task-config-schema} summarizes the core fields in each
task configuration.

\begin{center}
\centering
\captionof{table}{Core fields in a task configuration.}
\label{tab:task-config-schema}
\scriptsize
\renewcommand{\arraystretch}{1.08}
\begin{tabular}{>{\raggedright\arraybackslash}p{0.35\columnwidth}
                >{\raggedright\arraybackslash}p{0.55\columnwidth}}
\toprule
\textbf{Field} & \textbf{Content} \\
\midrule
\texttt{id}, \texttt{snapshot} & Task identifier and VM/software snapshot. \\
\texttt{instruction} & Natural-language goal shown to the agent. \\
\texttt{config} & Ordered setup actions: \texttt{launch}, \texttt{vm\_file}, \texttt{execute}, \texttt{sleep}. \\
\texttt{trajectory}, \texttt{related\_apps} & Output path and involved applications. \\
\texttt{difficulty} & Expert-assigned easy/medium/hard label. \\
\texttt{evaluator} & Aligned \texttt{func}, \texttt{result}, target, and conjunction logic. \\
\texttt{workflow\_stages} & Mixed-task stages with domain, agent type, goal, and artifacts. \\
\bottomrule
\end{tabular}
\end{center}

\noindent The evaluator field defines the partial-credit score structure. GUI tasks
pair domain-specific getters, such as file-open, visual-state, pixel, or OCR
checks, with metric functions. Origin tasks add a secondary figure-quality score
after deterministic file checks. Code tasks use key-value or file-output matching
when deterministic references exist, while OPTIMADE uses valid-output checks
because provider results may vary. Mixed tasks add \texttt{workflow\_domains} and
\texttt{workflow\_stages} to evaluate artifact transfer between code and GUI tools.

\noindent Listing~\ref{lst:task-json-example} shows a representative task
configuration, shortened from an OriginPro example. The same schema is used across
GUI, code, Origin, and mixed tasks, with domain-specific initialization and evaluator
items.

\begin{lstlisting}[language={},basicstyle=\ttfamily\tiny,breaklines=true,
numbers=none,xleftmargin=0pt,frame=single,framerule=0.6pt,
caption={Abbreviated task JSON example.},label={lst:task-json-example}]
{
  "id": "5ab55722-9775-4074-ae3c-ca56028acc5f-wos",
  "snapshot": "originlab",
  "instruction": "Open ... CE2.ogwu, plot a voltage-time curve, annotate '5 mA cm^-2, 5 mAh cm^-2', and save CE2.png.",
  "config": [
    {"type": "launch",
     "parameters": {"command": ["C:\\Program Files\\OriginLab\\Origin2026\\Origin64.exe"]}},
    {"type": "sleep", "parameters": {"seconds": 30}},
    {"type": "execute",
     "parameters": {"command": ["python", "-c", "pyautogui.hotkey('ctrl','o')"]}}
  ],
  "related_apps": ["Origin"],
  "evaluator": {
    "func": ["exact_match", "pass_through"],
    "result": [
      {"type": "file_exists",
       "file_path": "C:\\Users\\Docker\\Desktop\\setup\\output_result\\origin\\CE2.png"},
      {"type": "origin_aesthetic_score",
       "file_path": "C:\\Users\\Docker\\Desktop\\setup\\output_result\\origin\\CE2.png",
       "description": "The image is a galvanostatic long-cycle voltage-time curve."}
    ],
    "expected": [{"type": "rule", "rules": {"expected": true}}, null],
    "func_conj": "and"
  },
  "difficulty": "medium"
}
\end{lstlisting}

\section{Appendix B. Additional Task Distribution}
\label{sec:additional-task-distribution}

\begin{figure}[H]
  \centering
  \includegraphics[width=0.92\linewidth]{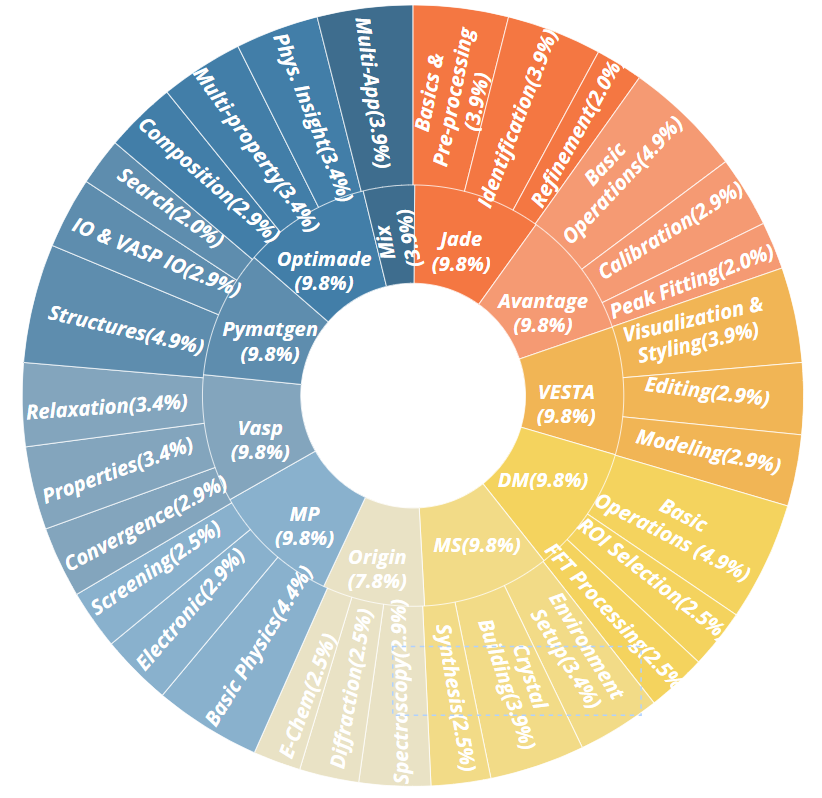}
  \caption{Task distribution across 10 domains and their sub-categories.}
  \label{fig:task_distribution}
\end{figure}

\noindent
Figure~\ref{fig:task_distribution} complements
Table~\ref{tab:task-stats} by showing the distribution of tasks across
10 tool domains and their operational sub-categories. The outer ring reports
domain-level proportions, while the inner segments summarize the composition of
task families within each domain.

\subsection{Distribution and Taxonomy}

The distribution is intentionally non-uniform. GUI-intensive tools are divided
into finer categories because their difficulty depends strongly on visual
states, modal dialogs, and tool-specific interaction conventions. In contrast,
code and database tasks are grouped mainly by query or structure-analysis
intent, where variation arises from API constraints, returned fields, and
output validation rather than interface manipulation.

The taxonomy covers three major patterns of scientific work. First,
\textbf{spectroscopy tasks}  include file import, peak/background processing,
axis or view adjustment, and report or export operations. Second,
\textbf{crystal-structure tasks} involve display modification, atom or boundary
editing, orientation adjustment, and structural inspection. Third,
\textbf{database tasks} focus on filtering, property extraction, structure
retrieval, and validation of saved outputs. This taxonomy enables the
partial-credit criteria to assess both intermediate actions and final
scientific outputs, providing a more fine-grained view of agent performance
than binary success alone.

\section{Appendix C. Complete Action Interface}
\label{sec:action-api}

\noindent
The \texttt{Computer} object provides a unified Python API for agent--VM
interaction. Commands are sent via HTTP POST and executed server-side through
a compact action space covering mouse, keyboard, clipboard, program control,
and window management. Table~\ref{tab:computer_api} summarizes the complete
action interface and its execution constraints.

\begin{table*}[p]
\centering
\caption{Complete \texttt{Computer} action interface and execution constraints
on the Windows VM. Calls are sent as Python strings via HTTP POST and executed
server-side with \texttt{exec(code, \{"computer": computer\})}.}
\label{tab:computer_api}
\scriptsize
\renewcommand{\arraystretch}{1.04}
\setlength{\tabcolsep}{3.2pt}
\begin{tabular}{p{2.15cm}p{3.65cm}p{3.10cm}p{7.05cm}}
\toprule
\textbf{Module} & \textbf{Method} & \textbf{Parameters} & \textbf{Description} \\
\midrule
\multirow{7}{*}{\texttt{mouse}}
  & \texttt{move\_id(id)}      & Element ID (int)              & Optional; available only when SoM or accessibility-tree observations are enabled. Not used in the reported screenshot-only experiments \\
  & \texttt{move\_abs(x, y)}   & Normalized coords $\in[0,1]$  & Move cursor to a relative screen position; $(0,0)$=top-left, $(1,1)$=bottom-right \\
  & \texttt{single\_click()}   & —                             & Left single-click at the current cursor position \\
  & \texttt{double\_click()}   & —                             & Left double-click at the current cursor position \\
  & \texttt{right\_click()}    & —                             & Right-click at the current cursor position \\
  & \texttt{scroll(dir)}       & \texttt{"up"} / \texttt{"down"} & Scroll the active window by 400 units in the given direction \\
  & \texttt{drag(x, y)}$^\dagger$ & Screen pixel coords (int)  & Left-button drag from current position to the target screen coordinates \\
\midrule
\multirow{2}{*}{\texttt{keyboard}}
  & \texttt{write(text)}$^\ddagger$ & ASCII string            & Type text character-by-character via \texttt{pyautogui.write()}; suitable only for short labels and filenames \\
  & \texttt{press(key)}        & Key name or combo           & Press a single key (e.g.\ \texttt{"f5"}, \texttt{"enter"}) or hotkey combo (e.g.\ \texttt{"ctrl+a"}, \texttt{"alt+4"}) \\
\midrule
\multirow{5}{*}{\texttt{clipboard}}
  & \texttt{copy\_text(text)}  & Unicode string (optional)   & Copy \textit{text} to the system clipboard via \texttt{pyperclip.copy()}; supports full Unicode. If called with no argument, sends \texttt{Ctrl+C} to copy the current selection \\
  & \texttt{paste()}           & —                           & Paste clipboard content at the current focus via \texttt{Ctrl+V} \\
  & \texttt{view\_clipboard()} & —                           & Return the current clipboard text as a Python string (via \texttt{pyperclip.paste()}) \\
  & \texttt{copy\_image(id)}   & Element ID (int)            & Crop the SoM element's bounding box from the screenshot and write it to the clipboard as a DIB bitmap via \texttt{win32clipboard} \\
  & image-copy note            & Element ID + description    & Used only when the observation includes SoM element IDs; screenshot-only agents cannot call ID-based image copy reliably \\
\midrule
\multirow{4}{*}{\texttt{os}}
  & \texttt{open\_program(name)}    & Executable name (str) & Launch the named program via \texttt{start}, then maximize and bring to foreground; no-op if not installed \\
  & \texttt{maximize\_window()}     & Window handle (opt.)  & Maximize the specified window, or the current foreground window if omitted, via \texttt{win32gui} \\
  & \texttt{is\_installed(name)}    & Executable name (str) & Return \texttt{True} if the program is found in the Windows registry or PATH \\
  & program policy                  & Application alias     & Prefer explicit program launch or window switching over desktop-icon clicking to reduce dependence on icon layout \\
\midrule
\multirow{3}{*}{\makecell[l]{\texttt{window}\\\texttt{\_manager}}}
  & \makecell[l]{\texttt{switch\_to}\\\texttt{\_application(name)}} & App name (str) & Restore and maximize the closest-matching visible window (fuzzy match via \texttt{difflib}), then bring it to the foreground \\
  & \makecell[l]{\texttt{find\_open}\\\texttt{\_applications()}}    & —              & Return a list of title strings for all currently visible, non-empty windows \\
  & focus recovery                         & App name or title fragment & Used after setup, file dialogs, or script execution to recover the intended foreground application before the next action \\
\midrule
\multirow{6}{*}{runtime}
  & transport                              & HTTP POST request     & The client sends one Python code block per step to the VM-side executor; the executor returns status, stdout, stderr, and any exception message \\
  & namespace                              & \texttt{computer} object & The executed code receives only the exposed \texttt{Computer} object and standard Python runtime needed for action composition \\
  & coordinate frame                       & Normalized or pixel coords & \texttt{move\_abs} uses normalized screen coordinates, while \texttt{drag} uses raw screen pixels; mixing the two is a common failure mode \\
  & Unicode policy                         & Clipboard preferred   & Long scripts, non-ASCII text, and multi-line commands should be copied through the clipboard rather than typed through \texttt{keyboard.write} \\
  & error handling                         & JSON status payload   & Invalid method calls, bad parameters, or backend exceptions return an error status; the episode continues unless the agent terminates \\
  & termination                            & Agent-level signal    & The agent marks completion with the framework-level \texttt{DONE} decision; evaluators then capture a fresh VM state and score sub-criteria \\
\bottomrule
\end{tabular}
\end{table*}

\section{Appendix D. Execution Environment and Reproducibility}
\label{sec:azure-env}
\noindent This appendix records the runtime settings needed to reproduce the
main evaluation. All large-scale experiments reported in this paper were executed on
\textbf{Microsoft Azure Machine Learning} using the \texttt{run\_azure.py} launcher.
Each evaluation job runs inside a Docker container on an Azure ML
\textbf{Compute Instance} of type \texttt{Standard\_D8\_v3}
(8\,vCPUs, 28\,GiB RAM), which provides sufficient resources to host the
outer Linux container together with the inner QEMU-based Windows~11 VM.

\paragraph{Episode initialisation.}
Each task is launched from its JSON configuration, which specifies the user
instruction, setup actions, target files, and scoring criteria. Before the agent
acts, the VM is reset to the task-specific initial state using the same input
files and screen resolution. The agent then receives observations until it
issues \texttt{DONE} or exhausts the 50-step budget. Post-hoc evaluation is
run only after the episode terminates, so intermediate rewards do not alter the
agent policy during the benchmark run.

\paragraph{Parallelism and isolation.}
The launcher spawns one Compute Instance per worker and distributes tasks across
workers via the \texttt{num\_workers} parameter in \texttt{experiments.json}.
Each instance is created on-demand at job start and deleted after the job completes,
keeping costs proportional to actual evaluation time.
For local multi-worker runs, each worker uses a copy-on-write overlay derived
from the same base VM image. Thus, parallel workers share neither application
state nor generated output files.

\paragraph{Storage.}
Task configuration files, VM disk images, and result artifacts are stored in
\textbf{Azure Blob Storage} and mounted to each Compute Instance at runtime.
This decouples storage from compute and allows results to be incrementally
synchronised to a local machine via \texttt{sync\_results.py} while jobs are
still running.

\paragraph{Startup sequence.}
Each Compute Instance runs \path{compute-instance-startup.sh}, which (1)~pulls
the \texttt{mattoolbench/mattoolbench:latest} Docker image, (2)~mounts the Azure
Blob datastore, and (3)~launches the evaluation container with the appropriate task
configuration. External API calls are routed through a \texttt{tinyproxy} instance
injected into the container environment on boot.

\paragraph{API and decoding settings.}
Unless otherwise stated for ablations, all model calls use the same decoding and
execution settings in Table~\ref{tab:api-settings}. All agent types use the same
maximum output-token budget.

\begin{center}
\captionof{table}{Shared API and execution settings for main experiments.}
\label{tab:api-settings}
\scriptsize
\renewcommand{\arraystretch}{1.05}
\setlength{\tabcolsep}{3pt}
\resizebox{\linewidth}{!}{%
\begin{tabular}{ll}
\toprule
\textbf{Parameter} & \textbf{Value} \\
\midrule
Observation input & Raw screenshot for GUI/Origin; text feedback for code \\
Accessibility/SoM & Not used in main experiments; screenshot-only observations \\
Action interface & Python \texttt{Computer}/PyAutoGUI actions or generated Python scripts \\
Temperature / top-$p$ & 1.0 / 0.9 \\
Max output tokens & 8000 for GUI/code; 8000 for Origin \\
Step budget & 50 agent steps per task \\
Post-action wait & 3 seconds before the next observation \\
Origin judge & Gemini-3.1-pro-preview for secondary aesthetic scoring \\
Released artifacts & Task JSON, input files, gold/reference outputs, Origin scripts \\
Parallel isolation & One copy-on-write VM storage per worker \\
\bottomrule
\end{tabular}
}
\end{center}

\paragraph{Agent wrappers.}
GUI and Origin tasks use the same \textsc{NaviAgent} loop with screenshots as
visual input. The GUI agent emits calls to the \texttt{Computer} action API,
whereas the Origin agent interacts with OriginPro and may generate longer script
fragments. Code tasks use a text-only code agent: the prompt contains the task
instruction and available files, and the generated Python program is executed in
the domain-specific environment configured for that task.

\paragraph{Evaluation and aggregation.}
For GUI tasks, final states are extracted by deterministic OCR, pixel, SSIM, or
file-state getters, depending on the task configuration. For Origin tasks,
success is determined by deterministic output-file checks; the LLM aesthetic
judge only contributes a secondary quality score. For code tasks, outputs are
checked against reference files, numeric/key-value tolerances, or validated saved
values when the underlying database provider may return non-unique records. A
task is counted as successful only when all required sub-criteria are satisfied;
the reported Score is the sum of partial-credit sub-criteria normalized by the
available points.

\paragraph{Reproducibility controls.}
To limit avoidable infrastructure variation, all evaluations use the same VM
image, Docker image, task setup scripts, input files, decoding settings, step
budget, and post-action wait time. The released benchmark package includes task
JSON files, input artifacts, gold/reference outputs, evaluator definitions, and
Origin reference scripts, allowing the same environment and scoring pipeline to
be redeployed.

\paragraph{Comparison with local execution.}
Table~\ref{tab:local-vs-azure} summarises the key differences between local and
Azure execution modes.

\begin{center}
\captionof{table}{Local vs.\ Azure execution at a glance.}
\label{tab:local-vs-azure}
\scriptsize
\renewcommand{\arraystretch}{1.05}
\setlength{\tabcolsep}{2.5pt}
\resizebox{\linewidth}{!}{%
\begin{tabular}{lll}
\toprule
\textbf{Aspect} & \textbf{Local} & \textbf{Azure} \\
\midrule
Entry point    & \texttt{run-local.sh}       & \texttt{run\_azure.py} \\
Storage        & Local disk (\texttt{vm/storage/}) & Azure Blob Storage \\
Concurrency    & Serial, single machine      & Parallel ML Jobs (\texttt{num\_workers}) \\
Docker image   & Built locally               & Pulled from registry \\
VM image       & Persisted locally           & Uploaded to datastore \\
Compute type   & Any local machine           & \texttt{Standard\_D8\_v3} \\
\bottomrule
\end{tabular}
}
\end{center}

\section{Appendix E. Complete Task List}
\label{sec:task-list}
\noindent This appendix lists the 196 individually specified single-tool tasks in
\textsc{MatToolBench}; the 8 mixed cross-tool diagnostic tasks are reported
separately in the main text. Each entry is formatted as
\textbf{Domain-Difficulty} followed by the shortened task instruction, where
\textbf{E}, \textbf{M}, and \textbf{H} denote easy, medium, and hard.

\scriptsize
\begin{description}[leftmargin=1.78cm,labelwidth=1.62cm,labelsep=0.10cm,
  itemsep=0.15pt,parsep=0pt,topsep=2pt,style=standard,align=left]
\item[\textbf{Avantage-E}] Import O1s Scan.VGD, select Display Modes to show data in stacked graph mode.
\item[\textbf{Avantage-H}] Import C1s Scan.VGD + C1s Scan2.VGD; copy spectrum to window B; apply Add Constant 3333; arrange vertically; save.
\item[\textbf{Avantage-E}] Import Zn2p Scan.VGD, adjust X-Axis font size.
\item[\textbf{Avantage-E}] Import Zn2p Scan.VGD, reverse the X-axis direction.
\item[\textbf{Avantage-E}] Import Zn2p Scan.VGD, set X-Axis title colour.
\item[\textbf{Avantage-E}] Import O1s Scan.VGD, perform charge correction.
\item[\textbf{Avantage-E}] Import XPS Survey.VGD, copy the spectrum to window B.
\item[\textbf{Avantage-E}] Import C1s Scan.VGD, adjust grid properties.
\item[\textbf{Avantage-E}] Import Zn2p Scan.VGD, perform automatic peak fitting.
\item[\textbf{Avantage-E}] Import O1s Scan.VGD, intelligently add O1s background.
\item[\textbf{Avantage-H}] Import C1s Scan.VGD, copy spectrum to window B; apply Add Constant; arrange vertically; save as stacked\_3333.vgp.
\item[\textbf{Avantage-E}] Import C1s Scan.VGD + C1s Scan2.VGD, display both spectra.
\item[\textbf{Avantage-E}] Import XPS Survey.VGD, smooth the spectrum.
\item[\textbf{Avantage-H}] Import C1s Scan.VGD, add and lock a Shirley background.
\item[\textbf{Avantage-E}] Import C1s Scan.VGD, double-click to zoom.
\item[\textbf{Avantage-M}] Import O1s Scan.VGD, intelligently add background.
\item[\textbf{Avantage-E}] Import C1s Scan.VGD (file open only).
\item[\textbf{Avantage-M}] Import C1s Scan.VGD + C1s Scan2.VGD, configure comparison view.
\item[\textbf{Avantage-E}] Import Zn2p Scan.VGD, add background using manual method.
\item[\textbf{Avantage-H}] Import O1s Scan.VGD, add a Peak (range start/end), set peak parameters, save.
\item[\textbf{DM-E}] Import dm4.dm3, use Box tool to draw a frame on the image.
\item[\textbf{DM-E}] Import dm2.dm3, use Rectangle\allowbreak ROI at centre of image.
\item[\textbf{DM-E}] Import dm5.dm3, use Oval tool to draw an ellipse.
\item[\textbf{DM-H}] Import dm7.dm3, perform FFT and inverse FFT.
\item[\textbf{DM-H}] Import dm9.dm3, perform FFT analysis, apply mask, inverse FFT.
\item[\textbf{DM-E}] Import dm8.dm3, select rotate, rotate image.
\item[\textbf{DM-M}] Import dm6.dm3, perform FFT analysis and apply mask.
\item[\textbf{DM-M}] Import dm1.dm3, use Box inside lattice image for FFT.
\item[\textbf{DM-E}] Import dm4.dm3, perform FFT on lattice image.
\item[\textbf{DM-E}] Import dm3.dm3, drag the dm3 file to the workspace.
\item[\textbf{DM-M}] Import dm1.dm3, use RectangleROI to draw a region.
\item[\textbf{DM-E}] Import dm1.dm3, change the scale bar to nm.
\item[\textbf{DM-E}] Import dm2.dm3, change the scale bar font size.
\item[\textbf{DM-M}] Import dm5.dm3, use Box to draw a frame and measure.
\item[\textbf{DM-E}] Import dm10.dm3, select mask tools in FFT.
\item[\textbf{DM-E}] Import dm4.dm3, use Profile tool to draw a line profile.
\item[\textbf{DM-E}] Import dm6.dm3, select operator filter.
\item[\textbf{DM-E}] Import dm9.dm3, measure average and standard deviation.
\item[\textbf{DM-M}] Import dm2.dm3, on optimised lattice image draw ROI.
\item[\textbf{DM-E}] Import dm7.dm3, select scale and set scale bar.
\item[\textbf{JADE-E}] Open WRT-ZSX-5.txt, perform Whole Pattern Fitting.
\item[\textbf{JADE-E}] Open XRD1.xrdml, perform background subtraction with default parameters.
\item[\textbf{JADE-E}] Open XRD1.xrdml, switch x-axis from 2$\theta$ to d-spacing, apply smoothing.
\item[\textbf{JADE-E}] Open XRD4.xrdml, perform peak search with default parameters.
\item[\textbf{JADE-M}] Open WRT-ZSX-5.txt, remove background, perform peak search.
\item[\textbf{JADE-E}] Open XRD4.xrdml, apply smoothing once.
\item[\textbf{JADE-E}] Open WRT-ZSX-5.txt, perform profile fitting.
\item[\textbf{JADE-E}] Open XRD1.xrdml, perform peak search with default parameters.
\item[\textbf{JADE-E}] Import WRT-ZSX-5.txt file.
\item[\textbf{JADE-E}] Open XRD1.xrdml, switch x-axis from 2$\theta$ to d-spacing directly.
\item[\textbf{JADE-M}] Open WRT-ZSX-5.txt (Fe and Ni), perform Rietveld refinement.
\item[\textbf{JADE-E}] Open WRT-ZSX-5.txt, perform peak search then profile fitting.
\item[\textbf{JADE-E}] Open WRT-ZSX-5.txt, select Zoom from menu bar.
\item[\textbf{JADE-E}] Open XRD4.xrdml, apply smoothing once and perform background removal.
\item[\textbf{JADE-M}] Open WRT-ZSX-5.txt, smooth once, remove background, perform peak search.
\item[\textbf{JADE-M}] Open XRD1.xrdml, perform full-pattern three-phase Rietveld refinement.
\item[\textbf{JADE-E}] Open WRT-ZSX-5.txt, add two peaks.
\item[\textbf{JADE-M}] Open XRD4.xrdml, apply smoothing, perform background removal and peak search.
\item[\textbf{JADE-E}] Open XRD2.xrdml, perform profile fitting.
\item[\textbf{JADE-E}] Open WRT-ZSX-5.txt, calculate interplanar spacing.
\item[\textbf{MS-M}] Open untitled.stp, create a new 3D Atomistic document.
\item[\textbf{MS-E}] Open untitled.stp, create a new 3D Atomistic document.
\item[\textbf{MS-H}] Build homo\allowbreak polymer (propylene, chain 40), select all atoms, open Forcite.
\item[\textbf{MS-E}] Create a new project, save to task\_1.stp.
\item[\textbf{MS-M}] Open untitled.stp, import file, configure supercell.
\item[\textbf{MS-M}] Open untitled.stp, import file, set force-field parameters.
\item[\textbf{MS-M}] Open untitled.stp, create 3D Atomistic document, add atom.
\item[\textbf{MS-M}] Open untitled.stp, import file, run geometry optimisation.
\item[\textbf{MS-H}] Open untitled.stp, build homopolymer (propylene, chain 40), run Forcite.
\item[\textbf{MS-E}] Open untitled.stp, import file.
\item[\textbf{MS-M}] Open untitled.stp, import file, set space group.
\item[\textbf{MS-H}] Build polypropyl acrylate chain (40 units, isotactic), run Forcite.
\item[\textbf{MS-M}] Open untitled.stp, import file, configure lattice parameters.
\item[\textbf{MS-M}] Open untitled.stp, import file, set atom properties.
\item[\textbf{MS-M}] Open untitled.stp, create 3D Atomistic document, configure display.
\item[\textbf{MS-E}] Create 3D Atomistic document (fractional coordinates), add default atom.
\item[\textbf{MS-E}] Open untitled.stp, import file.
\item[\textbf{MS-M}] Open untitled.stp, import file, set symmetry.
\item[\textbf{MS-E}] Open untitled.stp, import file.
\item[\textbf{MS-M}] Open untitled.stp, import file, configure bonds.
\item[\textbf{VESTA-H}] Import GaAs.cif, adjust boundary ranges, change style to Polyhedral.
\item[\textbf{VESTA-H}] Import GaN.cif, set y(max)=2, change style to Polyhedral, set coordination.
\item[\textbf{VESTA-E}] Import MgO.cif, create lattice plane with hkl indices.
\item[\textbf{VESTA-E}] Import Si.cif, remove several Si atoms.
\item[\textbf{VESTA-H}] Import GaN.cif, set Polyhedral style, toggle coordination polyhedra.
\item[\textbf{VESTA-E}] Import C.cif, check coordinate axes display.
\item[\textbf{VESTA-E}] Import GaAs.cif, delete atoms, save modified file.
\item[\textbf{VESTA-E}] Import GaN.cif, view N atom details, delete atoms.
\item[\textbf{VESTA-E}] Import Au.cif, change display style to Wireframe.
\item[\textbf{VESTA-E}] Import GaN.cif, zoom in by 100\%.
\item[\textbf{VESTA-E}] Import GaN.cif, view coordinates and atom details.
\item[\textbf{VESTA-E}] Import C.cif, translate structure 400 units right.
\item[\textbf{VESTA-E}] Import Fe.cif, select two Fe atoms, view bonding structure.
\item[\textbf{VESTA-H}] Import Si.cif, set fractional coordinate ranges, change style.
\item[\textbf{VESTA-E}] Import Fe.cif, rotate crystal to standard crystallographic orientation.
\item[\textbf{VESTA-M}] Import Al2O3.cif, adjust fractional coordinate ranges.
\item[\textbf{VESTA-E}] Import MgO.cif, set up vector to [3,3,3].
\item[\textbf{VESTA-E}] Import Cu.cif, rotate crystal 90\textdegree\ left around y-axis.
\item[\textbf{VESTA-H}] Import NaCl.cif, remove bond, modify structure, save.
\item[\textbf{VESTA-M}] Import MgO.cif, change atomic radius to 1.5, save.
\item[\textbf{Origin-H}] Open CPO1.opju, plot FTIR spectrum in transmittance mode, add legend.
\item[\textbf{Origin-M}] Open Cycle2.ogwu, plot battery cycling performance figure with labels.
\item[\textbf{Origin-M}] Open CPO2.ogwu, plot FTIR spectrum in transmittance mode, set sample name.
\item[\textbf{Origin-H}] Open book2.ogwu, plot Raman stacked/offset plot, annotate peaks.
\item[\textbf{Origin-M}] Open CE2.ogwu, plot galvanostatic long-cycle voltage-time curve, annotate.
\item[\textbf{Origin-H}] Open book9.ogwu, plot Raman stacked/offset plot, annotate peaks.
\item[\textbf{Origin-M}] Open XPS1.ogwu, plot XPS spectrum with labels [Zn 2p1/2, Zn 2p3/2].
\item[\textbf{Origin-H}] Open XRD2.txt + card file, plot XRD pattern with phase annotations.
\item[\textbf{Origin-M}] Open XPS2.ogwu, plot XPS spectrum with labels [O-C=O, C-O, C-C].
\item[\textbf{Origin-H}] Open CE1.ogwu, plot galvanostatic long-cycle voltage-time curve, annotate.
\item[\textbf{Origin-H}] Open XRD1.txt + card file, plot XRD pattern with phase annotations.
\item[\textbf{Origin-M}] Open CPO1.opju, plot FTIR spectrum in transmittance mode (no inset).
\item[\textbf{Origin-M}] Open Cycle1.ogwu, plot battery cycling performance figure with labels.
\item[\textbf{Origin-H}] Open step.opju, plot free energy step diagram with labels.
\item[\textbf{Origin-H}] Open WRT-ZSX-5.txt + card files, plot XRD pattern with phase annotations.
\item[\textbf{Origin-H}] Open book3.ogwu, plot Raman stacked/offset plot, annotate peaks.
\item[\textbf{MP-E}] Get relaxed structure of mp-153 (hcp Mg), extract lattice vector lengths a and b.
\item[\textbf{MP-E}] Search all TiO2 materials, count total material IDs.
\item[\textbf{MP-E}] Query mp-1201: band gap, magnetic ordering, is\_magnetic.
\item[\textbf{MP-M}] Query mp-19005 formation energy per atom and decomposition products.
\item[\textbf{MP-M}] Query mp-149 structure, convert to VASP POSCAR format.
\item[\textbf{MP-M}] Search Fe-Ni materials with energy\_above\_hull=0, extract stable phases.
\item[\textbf{MP-E}] Get space group symbol and number of mp-1143.
\item[\textbf{MP-E}] Query elasticity of mp-134: K\_VRH and G\_VRH.
\item[\textbf{MP-E}] Get phonon born effective charges of mp-2490.
\item[\textbf{MP-M}] Get band structure of mp-1434, extract high-symmetry point labels.
\item[\textbf{MP-M}] Search materials with band gap 2.0--3.0\,eV, exclude precious metals.
\item[\textbf{MP-M}] Get spin-polarised DOS of mp-13 (bcc Fe), record Spin.Up and Spin.Down maxima.
\item[\textbf{MP-M}] Query band gap of mp-19005, check if Hubbard U was enabled.
\item[\textbf{MP-E}] Search Al-O materials with energy\_above\_hull=0, list material IDs.
\item[\textbf{MP-M}] Search Fe-O materials, filter by ferromagnetic ordering, record first 5.
\item[\textbf{MP-M}] Get mp-126 structure, use SlabGenerator with Miller index (1,1,1).
\item[\textbf{MP-M}] Search insertion electrodes with Li, Fe, P, O (num\_elements=4).
\item[\textbf{MP-M}] Query piezoelectric total tensor of mp-2104, find maximum component.
\item[\textbf{MP-M}] Query phonon band structure of mp-149 via new API.
\item[\textbf{MP-E}] Get total dielectric tensor of mp-2534.
\item[\textbf{OQMD-E}] Query elemental Cu (ntypes=1), extract band\_gap/volume/delta\_e/stability.
\item[\textbf{OQMD-M}] Query Zn-O binary polymorphs (ntypes=2, limit=10), sort by delta\_e.
\item[\textbf{OQMD-M}] Query Li-Fe-O ternary compounds (ntypes=3, limit=10), calculate average delta\_e.
\item[\textbf{OQMD-M}] Query Ni-Al binary compounds (ntypes=2, limit=10), sort by delta\_e.
\item[\textbf{OQMD-E}] Query Mg-O compounds (limit=1), save first result.
\item[\textbf{OQMD-M}] Query Ga-N compounds with band gap 1.0--2.0\,eV.
\item[\textbf{OQMD-M}] Query Li ternary compounds twice (ntypes=3, limit=1), read total count.
\item[\textbf{OQMD-E}] Query Ba-Ti-O compounds (limit=10), save total count.
\item[\textbf{OQMD-E}] Query Fe2O3 entries (ntypes=2, natoms=5), extract first result.
\item[\textbf{OQMD-M}] Query Ce-O binary compounds with band\_gap$>$0, sort by band\_gap.
\item[\textbf{OQMD-M}] Query Cu-Au binary compounds (ntypes=2, limit=10), sort by delta\_e.
\item[\textbf{OQMD-E}] Count total entries containing Fe and O.
\item[\textbf{OQMD-M}] Query Fe binary compounds (ntypes=2, limit=10), sort by delta\_e.
\item[\textbf{OQMD-M}] Query La-Mn-O ternary compounds (ntypes=3, limit=10), sort by delta\_e.
\item[\textbf{OQMD-M}] Query Al-O compounds with volume 8--12\,\AA$^3$/atom.
\item[\textbf{OQMD-E}] Query first SiO2 entry, extract name/composition/delta\_e/stability.
\item[\textbf{OQMD-E}] Query first NaCl entry, extract name/composition/spacegroup/volume.
\item[\textbf{OQMD-E}] Count all binary compounds containing Al (ntypes=2).
\item[\textbf{OQMD-M}] Query Ti-O compounds (limit=10), sort by delta\_e ascending.
\item[\textbf{OQMD-E}] Query elemental Fe (ntypes=1), extract name/delta\_e/stability.
\end{description}
\begin{figure*}[t]
\centering

\begin{minipage}[t]{0.38\textwidth}
\centering
\includegraphics[width=\linewidth]{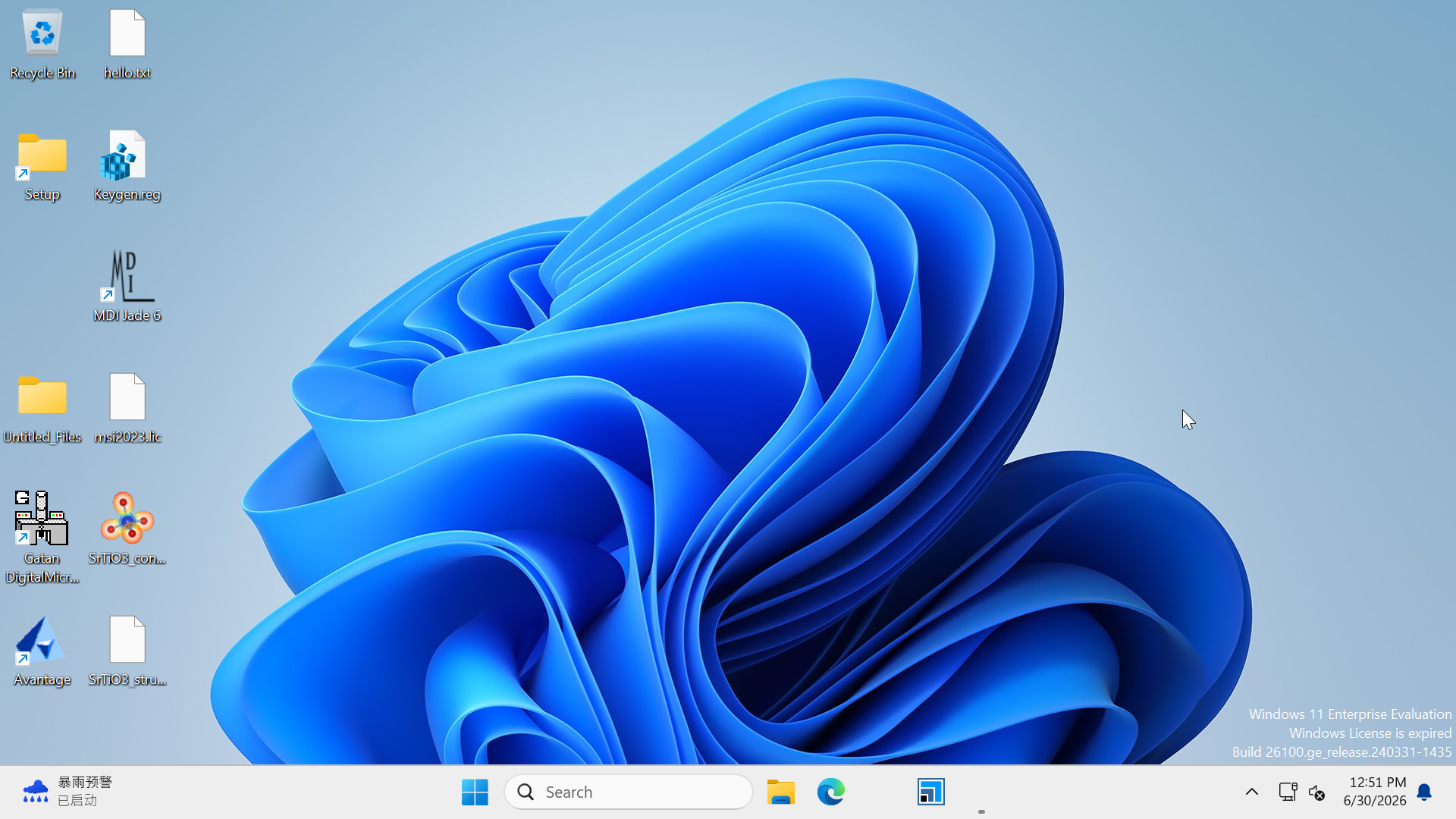}
\vspace{-3pt}

{\small\textbf{(a) Code stage: artifact creation.}}\\[-1pt]
{\scriptsize A Pymatgen/API stage writes a structure artifact.}

\end{minipage}
\hspace{0.02\textwidth}
\begin{minipage}[t]{0.38\textwidth}
\centering
\includegraphics[width=\linewidth]{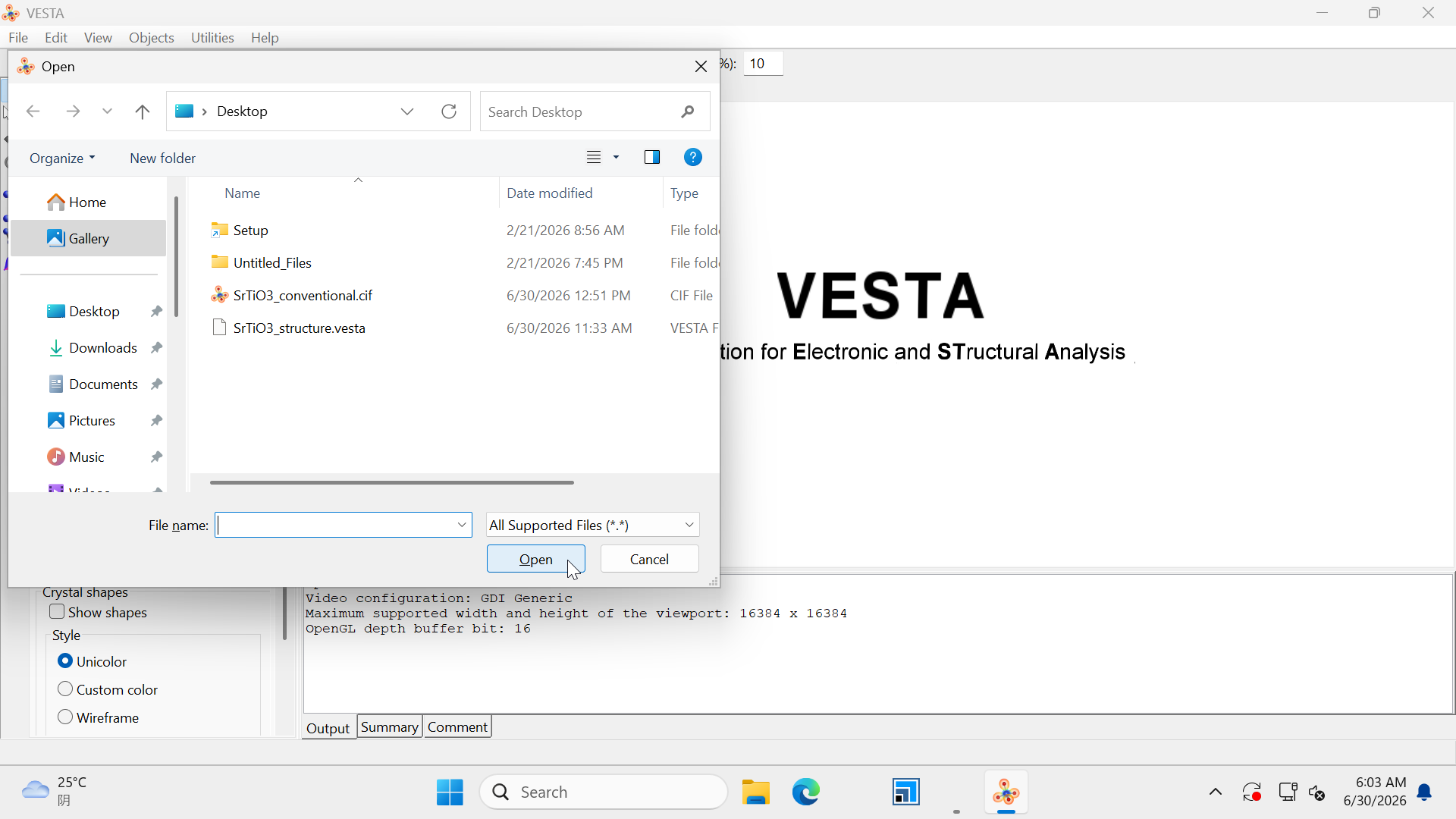}
\vspace{-3pt}
{\small\textbf{(b) GUI stage: artifact consumption.}}\\[-1pt]
{\scriptsize VESTA opens the generated structure file.}
\end{minipage}
\vspace{5pt}
\begin{minipage}[t]{0.38\textwidth}
\centering
\includegraphics[width=\linewidth]{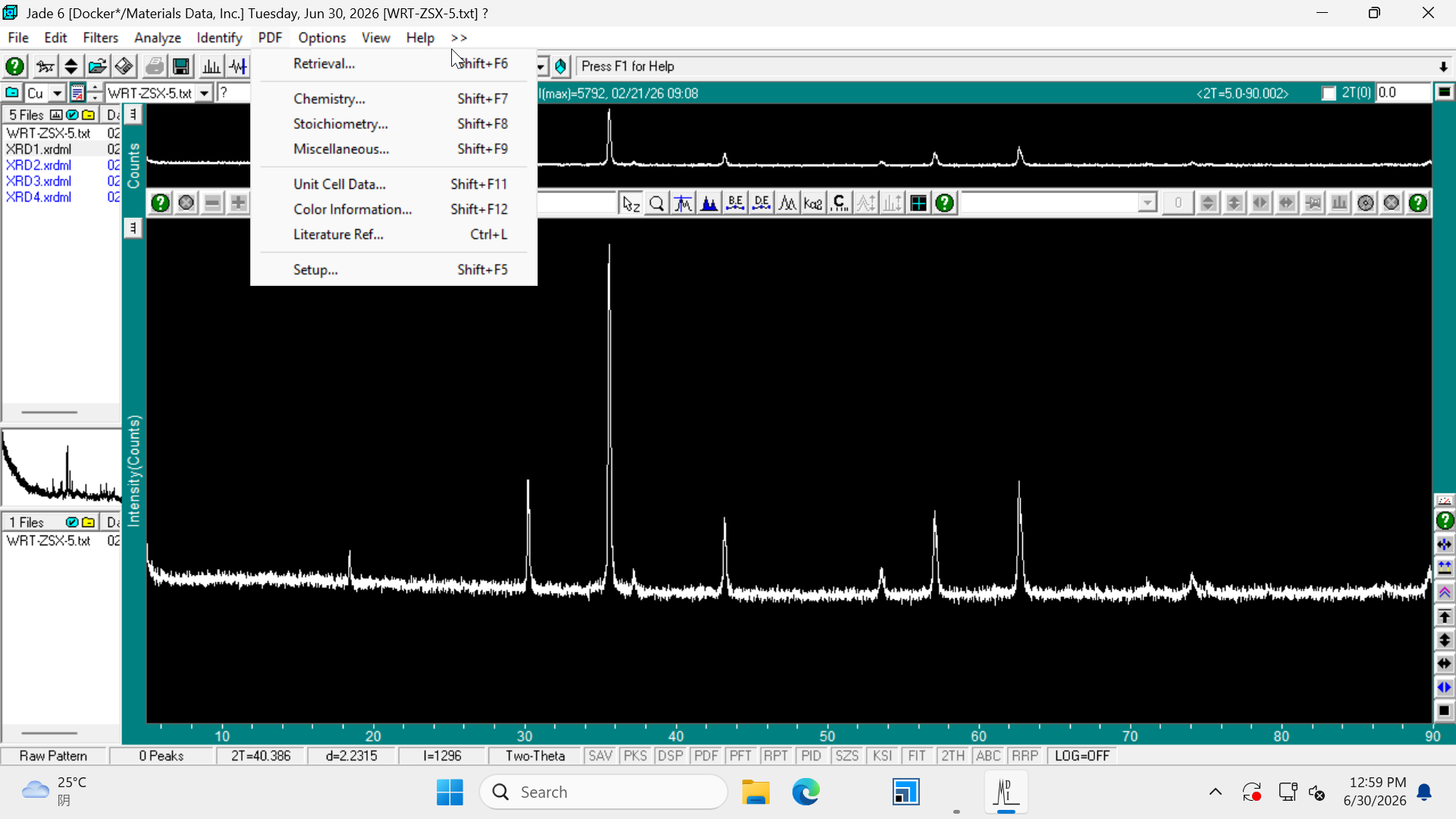}
\vspace{-3pt}
{\small\textbf{(c) GUI analysis stage.}}\\[-1pt]
{\scriptsize JADE performs XRD analysis in a GUI-only interface.}
\end{minipage}
\hspace{0.02\textwidth}
\begin{minipage}[t]{0.38\textwidth}
\centering
\includegraphics[width=\linewidth]{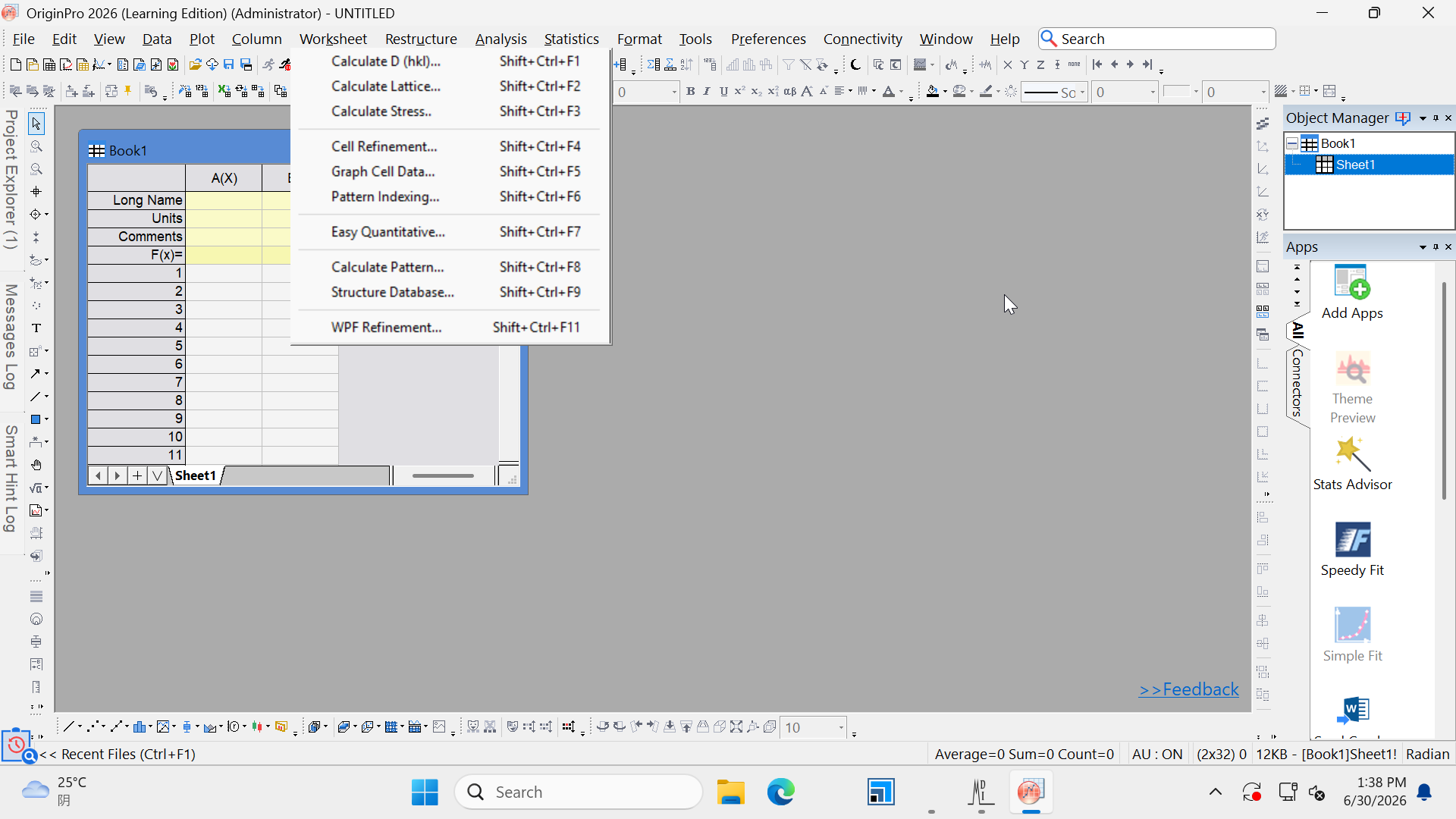}
\vspace{-3pt}
{\small\textbf{(d) Origin handoff stage.}}\\[-1pt]
{\scriptsize Origin receives the downstream plotting/analysis stage.}
\end{minipage}
\caption{Representative mixed-task trajectories. Mixed tasks require
stage-wise handoffs across software boundaries.}
\label{fig:mixed-examples}
\end{figure*}

\begin{description}
\item[\textbf{OPTIMADE-M}] Query structures with exactly 4 elements (nelements=4), randomly select 5.
\item[\textbf{OPTIMADE-E}] Query all structures containing Li, save total count.
\item[\textbf{OPTIMADE-M}] Query structures with formula\_\allowbreak prototype=``AB2'', query at least 2 databases.
\item[\textbf{OPTIMADE-E}] Query structures composed exclusively of C and H, save total count.
\item[\textbf{OPTIMADE-E}] Query structures with at most 2 sites (nsites$\leq$2), save total count.
\item[\textbf{OPTIMADE-M}] Query binary Ti-O structures (nelements=2), randomly select 5 results.
\item[\textbf{OPTIMADE-M}] Query structures with volume 50--200\,\AA$^3$ and exactly 3 elements.
\item[\textbf{OPTIMADE-M}] Query non-magnetic insulators with band gap$>$1\,eV via Materials Project OPTIMADE.
\item[\textbf{OPTIMADE-E}] Query structures with a single element type (nelements=1), save count.
\item[\textbf{OPTIMADE-M}] Query structures with chemical\_formula\_reduced=``SiO2'', count per database.
\item[\textbf{OPTIMADE-M}] Query structures with Ca, Ti, O as only three elements.
\item[\textbf{OPTIMADE-H}] Query structures with formula\_prototype=``ABC3'' (perovskite), query $\geq$2 databases.
\item[\textbf{OPTIMADE-M}] Query structures containing Fe, Co, or Ni with computed band gap.
\item[\textbf{OPTIMADE-H}] Query structures with Fe and S, at most 3 element types.
\item[\textbf{OPTIMADE-M}] Query records with formula\_prototype=``AB'' (NaCl-type), extract properties.
\item[\textbf{OPTIMADE-E}] Query structures with only Group-IV elements (C, Si, Ge, Sn, Pb).
\item[\textbf{OPTIMADE-E}] Query binary structures (nelements=2) with both elements in Group IV.
\item[\textbf{OPTIMADE-M}] Query Al-O compounds (nelements=2) that are insulators.
\item[\textbf{OPTIMADE-E}] Query structures containing both Fe and O, save count.
\item[\textbf{OPTIMADE-E}] Query structures with chemical\_formula\_reduced=``H2O'', save count.
\item[\textbf{Pymatgen-E}] Create Si diamond structure (a=5.43), save frac\_coords.
\item[\textbf{Pymatgen-E}] Calculate volume of hexagonal Ti structure (a=4.0, c=6.0).
\item[\textbf{Pymatgen-M}] Define Si structure, convert to Poscar, save VASP format text.
\item[\textbf{Pymatgen-E}] Create Incar object, set NSW=100 and EDIFF=1e-6.
\item[\textbf{Pymatgen-E}] Generate 8$\times$8$\times$8 Monkhorst-Pack k-point grid.
\item[\textbf{Pymatgen-E}] Parse vasprun.xml, extract Fermi level efermi.
\item[\textbf{Pymatgen-E}] Parse vasprun.xml, extract final energy from last ionic step.
\item[\textbf{Pymatgen-E}] Create hexagonal lattice (a=4.0, c=6.0), save lattice matrix.
\item[\textbf{Pymatgen-M}] Parse vasprun.xml, extract stress tensor from last ionic step.
\item[\textbf{Pymatgen-E}] Create hexagonal Ti structure at [0,0,0], save structure description.
\item[\textbf{Pymatgen-M}] Read POSCAR, extract num\_sites and formula.
\item[\textbf{Pymatgen-E}] Create C-C molecular fragment, save molecular coordinates.
\item[\textbf{Pymatgen-M}] Read POSCAR, extract num\_sites and atom types.
\item[\textbf{Pymatgen-E}] Measure C-C interatomic distance using get\_distance.
\item[\textbf{Pymatgen-M}] Read EIGENVAL, compute band gap via eigenvalue\_band\_gap.
\item[\textbf{Pymatgen-M}] Create CsCl primitive structure (space group 225, a=4.2).
\item[\textbf{Pymatgen-E}] Create CsCl structure, extract num\_sites (should be 2).
\item[\textbf{Pymatgen-M}] Create CsCl structure, build [2,2,1] supercell.
\item[\textbf{Pymatgen-E}] Create LiO molecule, call as\_dict(), save dictionary string.
\item[\textbf{Pymatgen-M}] Reconstruct LiO molecule from dictionary, verify formula is LiO.
\item[\textbf{Mixed-E}] Download mp-149 Si as CIF via Materials Project, open it in VESTA, remove one Si site, and save the modified structure.
\item[\textbf{Mixed-E}] Query elemental Si via OPTIMADE, export POSCAR, then use Pymatgen to identify space group and crystal system.
\item[\textbf{Mixed-E}] Download SrTiO3 (mp-5229), extract the conventional structure with Pymatgen, open in VESTA, switch to Ball-and-Stick, and save.
\item[\textbf{Mixed-M}] Build bilayer MoS2 with Pymatgen, export CIF, then open in Materials Studio, build a surface, and save the model.
\item[\textbf{Mixed-M}] Create a polyethylene crystal in Materials Studio, export CIF, then open in VESTA, set standard orientation and fractional ranges.
\item[\textbf{Mixed-M}] Process C1s XPS data in Avantage, export the report, then plot the exported spectrum in Origin.
\item[\textbf{Mixed-M}] Analyze WRT-ZSX-5 XRD data in JADE, export the matching PDF card, then plot the annotated XRD match in Origin.
\item[\textbf{Mixed-H}] Retrieve LiCoO2, generate three Li-content structures, and create separate VASP input directories for each composition.
\end{description}
\normalsize

\begin{table*}[t]
\centering
\caption{Bundled sample data files pre-installed in the evaluation VM, grouped by domain.}
\label{tab:dependencies}
\small
\renewcommand{\arraystretch}{1.15}
\resizebox{0.95\textwidth}{!}{%
\begin{tabular}{llp{8cm}}
\toprule
\textbf{Domain} & \textbf{File(s)} & \textbf{Description} \\
\midrule
Avantage & \texttt{C1s Scan.VGD}   & XPS narrow-scan spectrum for C\,1s core level \\
& \texttt{O1s Scan.VGD}   & XPS narrow-scan spectrum for O\,1s core level \\
& \texttt{XPS Survey.VGD} & Full XPS survey spectrum \\
& \texttt{Zn.vgp}         & Avantage project file for Zn sample \\
& \texttt{Zn2p Scan.VGD}  & XPS narrow-scan spectrum for Zn\,2p core level \\
\midrule
DM & \texttt{dm1.dm3}--\texttt{dm10.dm3} & TEM/STEM images in DigitalMicrograph format \\
& \texttt{5.dm3}          & Additional TEM image \\
\midrule
JADE & \texttt{WRT-ZSX-5.txt}  & Raw XRD data in plain-text format \\
& \texttt{WRT-ZSX-5.jip}  & JADE project file for WRT-ZSX-5 sample \\
& \texttt{XRD1--XRD4.xrdml} & XRD patterns in PANalytical XRDML format \\
& \texttt{XRD1--XRD4.jip} & Corresponding JADE project files \\
\midrule
MS & \texttt{AIGH-mol.xsd}   & Molecular structure of AIGH compound \\
& \texttt{Al2O3.xsd}      & Crystal structure of aluminium oxide \\
& \texttt{Fe.xsd}         & Crystal structure of iron \\
& \texttt{LiF.xsd}        & Crystal structure of lithium fluoride \\
& \texttt{Novolac4.xsd}   & Polymer structure of Novolac resin \\
& \texttt{SuperSi.xsd}    & Crystal structure of silicon supercell \\
& \texttt{TMOS.xsd}       & Molecular structure of TMOS \\
& \texttt{urea.xsd}       & Molecular structure of urea \\
\midrule
Origin & \texttt{Book2/3/9.ogwu} & Origin workbook files with tabular data \\
& \texttt{CE1/2.ogwu}, \texttt{CPO1.opju}, \texttt{CPO2.ogwu} & Cyclic voltammetry and electrochemical data \\
& \texttt{Cycle1/2.ogwu}  & Cycling performance data \\
& \texttt{step.opju}      & Origin project with step-function data \\
& \texttt{XPS\_1.ogwu}, \texttt{XPS2.ogwu} & XPS spectral data imported into Origin \\
& \texttt{XRD1/2.txt}, \texttt{XRD2.xrdml} & XRD data files for Origin import tasks \\
\midrule
Pymatgen & \texttt{POSCAR\_*} (20 files) & POSCAR input files for various crystal structures \\
& \texttt{vasprun.xml}    & Sample VASP-format output for parsing tasks \\
\midrule
VESTA
  & \texttt{mp-*\_*.cif} (10 files) & CIF crystal structure files from the Materials Project
    (Al$_2$O$_3$, MgO, Fe, Si, NaCl, GaAs, Cu, C, GaN, Au) \\
\bottomrule
\end{tabular}%
}
\end{table*}
\section{Appendix F. Mixed Cross-Tool Diagnostic Tasks}
\label{sec:mixed-diagnostics}
\noindent The eight mixed tasks are designed as cross-tool diagnostic examples,
not as another per-domain result category. Each task is split into ordered
stages in which an upstream environment creates an intermediate artifact that
is then consumed by a different tool. Figure~\ref{fig:mixed-examples}
illustrates this interaction pattern through representative trajectories rather
than aggregate model scores.


\noindent The first pattern is \textbf{code--GUI handoff}: a Python stage uses
Pymatgen or a database API to generate a structure file, and the next stage must
open that file in a GUI visualisation or modelling program.  The second pattern
is \textbf{GUI--Origin handoff}: a domain GUI such as JADE or Avantage must
produce a processed spectrum or table that is then replotted or further analyzed
inside Origin.  These examples expose an artifact-boundary bottleneck: the
agent must not only finish the first tool's operation, but also save the result
with the right path, extension, and semantic content for the next software.

\section{Appendix G. Bundled Dependency Files}
\label{sec:dependencies}

\noindent Table~\ref{tab:dependencies} documents the concrete files bundled
inside the VM. These files are part of the benchmark definition rather than
incidental examples: each task starts from a fixed file path and software state,
so the agent must operate on the same scientific input regardless of model,
worker, or execution date. The table also makes explicit which native formats
are exercised by each tool, such as \texttt{.VGD} for XPS analysis,
\texttt{.dm3} for microscopy, \texttt{.xrdml} for diffraction, \texttt{.xsd}
for Materials Studio, \texttt{.cif} for VESTA, and \texttt{POSCAR}/
\texttt{vasprun.xml} for Pymatgen.

\begin{table*}[t]
\centering
\caption{Getter types used in GUI-based domains, all paired with \texttt{exact\_match}.}
\label{tab:eval-gui-getters}
\small
\resizebox{.95\textwidth}{!}{%
\begin{tabular}{llp{9cm}}
\toprule
\textbf{Domain} & \textbf{Getter Type} & \textbf{Description} \\
\midrule
\multirow{6}{*}{Avantage}
  & \texttt{check\_multiple\_files\_imported} & Confirms that multiple XPS scan channels (\textit{e.g.}, C\,1s, O\,1s, Zn\,2p) are loaded. \\
  & \texttt{check\_dialog\_exists}            & Detects the presence of a specific dialog window via OCR. \\
  & \texttt{check\_dialog\_parameter}         & Extracts and validates a parameter value inside a dialog. \\
  & \texttt{check\_stacked\_graph}            & Verifies that spectra are displayed in stacked mode. \\
  & \texttt{check\_background\_added}         & Confirms background curve addition to a spectrum. \\
  & \texttt{check\_energy\_axis\_reversed}    & Validates that the binding-energy axis is inverted. \\
\midrule
\multirow{6}{*}{DM}
  & \texttt{check\_file\_import}             & Verifies that a \texttt{.dm3}/\texttt{.dm4} file is imported. \\
  & \texttt{check\_ocr\_text}               & Extracts arbitrary text from the UI via OCR and matches it against an expected string. \\
  & \texttt{check\_drawing\_tool}           & Confirms the active drawing tool (box, circle, line, \textit{etc.}). \\
  & \texttt{check\_border\_color}           & Detects the border color of a drawn shape via pixel analysis. \\
  & \texttt{check\_roi\_copy\_and\_enlarge} & Verifies ROI duplication and enlargement. \\
  & \texttt{check\_checkbox\_selected}      & Confirms a checkbox state from the accessibility tree. \\
\midrule
\multirow{8}{*}{JADE}
  & \texttt{check\_file\_opened\_jade}   & Confirms that the target XRD data file is open. \\
  & \texttt{check\_peak\_finding}        & Counts the number of detected peaks in the XRD pattern. \\
  & \texttt{check\_background\_removal}  & Validates background subtraction by detecting a $\geq$20\% drop in $I_{\max}$. \\
  & \texttt{check\_smoothing}            & Confirms the ``Smooth whole Pattern'' operation. \\
  & \texttt{check\_axis\_switch}         & Verifies switching between two-theta and $d$-scale axes. \\
  & \texttt{check\_wpf\_refinement}      & Extracts the $R$-factor from Whole Pattern Fitting. \\
  & \texttt{check\_refinement\_result}   & Counts Rietveld refinement profiles. \\
  & \texttt{check\_dialog\_title\_jade}  & Matches expected dialog titles via OCR. \\
\midrule
\multirow{5}{*}{MS}
  & \texttt{check\_file\_in\_title}    & Confirms the project filename in the window title bar. \\
  & \texttt{check\_text\_keyword}      & Searches for specific keywords in UI panels (\textit{e.g.}, ``3D Atomistic''). \\
  & \texttt{check\_dialog\_title\_ms}  & Matches dialog titles with optional forbidden-keyword filters. \\
  & \texttt{check\_field\_value}       & Extracts and validates form-field values (\textit{e.g.}, space group ``P1''). \\
  & \texttt{check\_visual\_similarity} & Computes SSIM between the current screenshot and a reference image. \\
\midrule
\multirow{8}{*}{VESTA}
  & \texttt{check\_file\_opened\_vesta}   & Verifies that a CIF structure file is loaded. \\
  & \texttt{check\_dialog\_opened\_vesta} & Confirms a dialog window is visible. \\
  & \texttt{check\_atom\_deletion}        & Validates that specified atoms have been removed. \\
  & \texttt{check\_boundary\_settings}    & Checks boundary cell display parameters. \\
  & \texttt{check\_style\_change}         & Verifies rendering style changes (sphere, ball-and-stick). \\
  & \texttt{check\_lattice\_plane}        & Confirms that a lattice plane is displayed. \\
  & \texttt{check\_rotation\_90\_up}      & Validates a 90\textdegree\ upward rotation of the structure. \\
  & \texttt{check\_polyhedral\_style}     & Confirms polyhedral representation is active. \\
\midrule
\multirow{2}{*}{Origin}
  & \texttt{file\_exists}                & Checks whether the output image file was created. \\
  & \texttt{validate\_image\_with\_model}& Encodes the output figure as base64 and queries Gemini-3.1-pro to verify
    it against a natural-language description of the expected plot. \\
\bottomrule
\end{tabular}%
}
\end{table*}

\noindent The dependency set is intentionally heterogeneous, combining raw
instrument outputs that require GUI import and visual confirmation with
structured materials files used by code-based parsers. This allows the same
environment to evaluate file opening, visualization, parameter editing,
scientific analysis, and artifact export across multiple data modalities.

\noindent To support realistic and reproducible evaluation,
\textsc{MatToolBench} ships a curated collection of authentic experimental and
simulation files pre-installed in the Windows~11 VM.
Table~\ref{tab:dependencies} lists these inputs by domain, including XPS
spectra, TEM/STEM images, XRD patterns, crystal structures, Origin workbooks,
and VASP-related outputs. Their native formats are preserved so that tasks test
realistic import, parsing, analysis, and export workflows rather than simplified
synthetic examples.

\noindent The bundled data cover both GUI- and code-oriented scenarios.
Avantage and DigitalMicrograph use spectroscopy and microscopy files, Pymatgen
uses POSCAR and \texttt{vasprun.xml}, and Origin, JADE, Materials Studio, and
VESTA use domain-specific project, structure, and tabular files. Reusing these
controlled inputs reduces setup overhead and evaluation variance, allowing
failures to be attributed to tool operation or reasoning rather than missing
files or inconsistent downloads.

\noindent File dependencies also encode domain conventions, such as core-level
scan names in XPS, distinctions between raw and project files in XRD, and
chemically meaningful atom and lattice information in structure visualization.
Thus, Table~\ref{tab:dependencies} specifies both the benchmark resources and
the scientific objects underlying each workflow.

\vspace{0.1cm}
\begin{table*}[tb]
\centering
\caption{Metric functions used per domain. \cmark: all tasks use this metric;
fractions indicate partial usage.}
\label{tab:eval-metrics}
\small
\begin{tabular}{lccc}
\toprule
\textbf{Domain} & \texttt{exact\_match} & \texttt{detect\_kv\_match} & \texttt{detect\_file\_match} \\
\midrule
Avantage  & \cmark & --     & --     \\
DM        & \cmark & --     & --     \\
JADE      & \cmark & --     & --     \\
MS        & \cmark & --     & --     \\
VESTA     & \cmark & --     & --     \\
Origin    & \cmark & --     & --     \\
MP        & --     & 14/20  & 6/20   \\
OQMD      & --     & \cmark & --     \\
OPTIMADE  & \cmark & --     & --     \\
Pymatgen  & --     & --     & \cmark \\
\bottomrule
\multicolumn{4}{p{0.92\textwidth}}{%
\footnotesize
\textit{exact\_match}: binary $\{0,1\}$ outcome for discrete GUI results and
binary conditions. \quad
\textit{detect\_kv\_match}: parses key-value pairs and matches numeric values
within $r_\text{tol}=0.1\%$, returning partial credit. \quad
\textit{detect\_file\_match}: line-by-line comparison against a gold-standard
reference file, returning a binary outcome.}
\end{tabular}
\end{table*}

\section{Appendix H. Evaluation Protocol Details}
\label{sec:eval-protocol-detail}

\noindent
\textsc{MatToolBench} employs a modular evaluation pipeline in which each task
sub-criterion is assessed by a domain-specific \emph{getter} that extracts the
relevant VM state, followed by a \emph{metric function} that converts the
extracted value into a score. Table~\ref{tab:eval-metrics} summarises the metric
functions, while Table~\ref{tab:eval-gui-getters} lists the GUI getter types.

\noindent A single generic OCR check is insufficient because different software
exposes state through different channels. Avantage and JADE require recognizing
dialogs, spectra, and fitting results; DigitalMicrograph relies on ROI state and
pixel-level properties; Materials Studio and VESTA require checking 3D rendering,
boundary settings, and structure orientation. The getters therefore combine OCR,
pixel analysis, SSIM, dialog recognition, and domain-specific predicates.

\noindent GUI and Origin sub-criteria use binary state checks. MP and OQMD use
key-value matching with numerical tolerance, while Pymatgen uses file-level
comparison. OPTIMADE is evaluated through a valid saved-output predicate rather
than a unique gold-file match, since records may vary across endpoints and time.

This two-stage design, state capture followed by score computation, makes
evaluation independent of the action sequence. Agents may use different menus,
shortcuts, or scripts but receive the same score when their final states satisfy
the required conditions. Getters inspect application state, text, screenshots,
files, structured outputs, and scientific artifacts, while metrics apply exact
matching, tolerance checks, boolean validation, file-existence checks, or image
similarity.

\noindent Formally, a task $x$ is specified by an instruction $g$, an
initialization program $c$, and $n$ scoring sub-criteria
$\mathcal{C}(x)=\{(g_i,m_i,w_i)\}_{i=1}^{n}$, where $g_i$ is a getter, $m_i$
is a metric function, and $w_i$ is the criterion weight. After an agent
trajectory terminates in final VM state $v_T$,
\begin{equation}
r_i = m_i(g_i(v_T), y_i), \qquad r_i \in [0,1],
\end{equation}
where $y_i$ is the reference value or predicate. The task score is
\begin{equation}
\mathrm{Score}(x)=
\frac{\sum_{i=1}^{n} w_i r_i}{\sum_{i=1}^{n} w_i},
\qquad
\mathrm{SR}(x)=\mathbb{I}\!\left[\forall i,\ r_i=1\right].
\end{equation}
This definition is shared across GUI, Origin, and code tasks; only the getter
and metric implementations differ. Partially correct final states can therefore
receive credit for completed scientific milestones.

\noindent The separation also makes failures interpretable by identifying
missing states such as an unopened file, unset parameter, unsaved plot, or
unwritten numeric field. It thereby distinguishes operational failures from
domain-reasoning failures instead of reducing multi-step workflows to a single
pass/fail label.

\newcommand{\lo}[1]{#1}   

\begin{table}[t]
  \centering
  \caption{Per-task evaluator reliability across five GUI domains: Score F1 and SR-F1.
    Values below 1.00 are explicitly reported. Means are macro-averaged over the 20 tasks in each domain:
    JADE\,0.991, MS\,0.994, Avantage\,0.980, VESTA\,0.979, DM\,0.936.}
  \label{tab:per_task_detail}
  \small
  \resizebox{\linewidth}{!}{%
  \setlength{\tabcolsep}{4pt}
  \begin{tabular}{c cc cc cc cc cc}
  \toprule
   & \multicolumn{2}{c}{\textbf{JADE}}
   & \multicolumn{2}{c}{\textbf{MS}}
   & \multicolumn{2}{c}{\textbf{Avantage}}
   & \multicolumn{2}{c}{\textbf{VESTA}}
   & \multicolumn{2}{c}{\textbf{DM}} \\
  \cmidrule(lr){2-3}\cmidrule(lr){4-5}\cmidrule(lr){6-7}\cmidrule(lr){8-9}\cmidrule(lr){10-11}
  \textbf{Task} & F1 & SR-F1 & F1 & SR-F1 & F1 & SR-F1 & F1 & SR-F1 & F1 & SR-F1 \\
  \midrule
  in1  & 1.00 & 1.00 & 1.00 & 1.00 & 1.00 & 1.00 & 1.00 & 1.00 & \lo{0.86} & 1.00 \\
  in2  & 1.00 & 1.00 & 1.00 & 1.00 & 1.00 & 1.00 & 1.00 & 1.00 & \lo{0.80} & \lo{0.86} \\
  in3  & 1.00 & 1.00 & 1.00 & 1.00 & 1.00 & 1.00 & 1.00 & 1.00 & 1.00 & 1.00 \\
  in4  & 1.00 & 1.00 & 1.00 & 1.00 & \lo{0.88} & \lo{0.92} & 1.00 & 1.00 & 1.00 & 1.00 \\
  in5  & 1.00 & 1.00 & 1.00 & 1.00 & 1.00 & 1.00 & 1.00 & 1.00 & 1.00 & 1.00 \\
  in6  & \lo{0.92} & 1.00 & 1.00 & 1.00 & 1.00 & 1.00 & \lo{0.97} & 1.00 & 1.00 & 1.00 \\
  in7  & 1.00 & 1.00 & \lo{0.98} & 1.00 & \lo{0.92} & 1.00 & 1.00 & 1.00 & 1.00 & 1.00 \\
  in8  & 1.00 & 1.00 & 1.00 & 1.00 & \lo{0.92} & \lo{0.89} & 1.00 & 1.00 & \lo{0.67} & 1.00 \\
  in9  & \lo{0.94} & 1.00 & 1.00 & 1.00 & 1.00 & 1.00 & 1.00 & 1.00 & 1.00 & 1.00 \\
  in10 & 1.00 & 1.00 & 1.00 & 1.00 & 1.00 & 1.00 & \lo{0.86} & \lo{0.86} & \lo{0.80} & 1.00 \\
  in11 & \lo{0.96} & 1.00 & \lo{0.99} & 1.00 & 1.00 & 1.00 & 1.00 & 1.00 & \lo{0.93} & 1.00 \\
  in12 & 1.00 & 1.00 & \lo{0.98} & 1.00 & \lo{0.95} & 1.00 & 1.00 & 1.00 & 1.00 & 1.00 \\
  in13 & 1.00 & 1.00 & 1.00 & 1.00 & 1.00 & 1.00 & \lo{0.89} & \lo{0.86} & \lo{0.67} & 1.00 \\
  in14 & 1.00 & 1.00 & 1.00 & 1.00 & 1.00 & 1.00 & 1.00 & 1.00 & 1.00 & 1.00 \\
  in15 & 1.00 & 1.00 & 1.00 & 1.00 & 1.00 & 1.00 & 1.00 & 1.00 & 1.00 & 1.00 \\
  in16 & 1.00 & 1.00 & \lo{0.98} & 1.00 & \lo{0.97} & 1.00 & 1.00 & 1.00 & 1.00 & 1.00 \\
  in17 & 1.00 & 1.00 & \lo{0.98} & 1.00 & 1.00 & 1.00 & \lo{0.98} & \lo{0.95} & 1.00 & 1.00 \\
  in18 & 1.00 & 1.00 & \lo{0.97} & 1.00 & 1.00 & 1.00 & \lo{0.91} & 1.00 & 1.00 & 1.00 \\
  in19 & 1.00 & 1.00 & 1.00 & 1.00 & 1.00 & 1.00 & 1.00 & 1.00 & 1.00 & 1.00 \\
  in20 & 1.00 & 1.00 & 1.00 & 1.00 & \lo{0.96} & 1.00 & \lo{0.97} & 1.00 & 1.00 & 1.00 \\
  \midrule
  \textbf{Mean} & \textbf{0.991} & \textbf{1.000} & \textbf{0.994} & \textbf{1.000} & \textbf{0.980} & \lo{\textbf{0.991}} & \textbf{0.979} & \lo{\textbf{0.984}} & \lo{\textbf{0.936}} & \lo{\textbf{0.993}} \\
  \bottomrule
  \end{tabular}%
  }
  \end{table}

\begin{table*}[t]
\centering
\small
\renewcommand{\arraystretch}{1.08}
\setlength{\tabcolsep}{3pt}
\caption{\textsc{MatToolBench} efficiency metrics.
Steps: mean steps across all episodes (max 50; fewer is better).
Steps*: mean steps on successful (SR\,=\,1) episodes.
FTR (\%): false termination rate.
\textbf{Cat.}: task category.
`NaN': no successful episodes (Steps* undefined).}
\label{tab:results-efficiency}
\resizebox{\textwidth}{!}{%
\begin{tabular}{ll ccc ccc ccc ccc ccc ccc ccc}
\toprule
& & \multicolumn{3}{c}{\makecell{\textbf{Doubao}\\\textbf{seed-1-8}}}
  & \multicolumn{3}{c}{\makecell{\textbf{Kimi}\\\textbf{k2.5}}}
  & \multicolumn{3}{c}{\makecell{\textbf{Claude}\\\textbf{sonnet-4.6}}}
  & \multicolumn{3}{c}{\makecell{\textbf{GPT}\\\textbf{-5.4}}}
  & \multicolumn{3}{c}{\makecell{\textbf{Qwen3-VL}\\\textbf{235B}}}
  & \multicolumn{3}{c}{\makecell{\textbf{Qwen3-VL}\\\textbf{32B}}}
  & \multicolumn{3}{c}{\makecell{\textbf{Qwen3-VL}\\\textbf{8B}}} \\
\cmidrule(lr){3-5}   \cmidrule(lr){6-8}   \cmidrule(lr){9-11}
\cmidrule(lr){12-14} \cmidrule(lr){15-17} \cmidrule(lr){18-20} \cmidrule(lr){21-23}
\textbf{Cat.} & \textbf{Domain}
  & \textbf{Steps} & \textbf{Steps*} & \textbf{FTR}
  & \textbf{Steps} & \textbf{Steps*} & \textbf{FTR}
  & \textbf{Steps} & \textbf{Steps*} & \textbf{FTR}
  & \textbf{Steps} & \textbf{Steps*} & \textbf{FTR}
  & \textbf{Steps} & \textbf{Steps*} & \textbf{FTR}
  & \textbf{Steps} & \textbf{Steps*} & \textbf{FTR}
  & \textbf{Steps} & \textbf{Steps*} & \textbf{FTR} \\
\midrule
\multirow{6}{*}{GUI}
  & Avantage & 36.1 & 19.9 & 25.0 & 42.4 & 36.0 & 60.0 & 41.1 & 26.1 & 25.0 & 39.2 & 7.8  & 50.0 & 39.7 & 8.0  & 71.4 & 46.2 & 31.7 & 60.0 & 44.9 & 42.4 & 66.7 \\
& JADE     & 34.6 & 24.2 & 50.0 & 35.3 & 15.6 & 44.4 & 35.9 & 15.2 & 44.4 & 29.0 & 8.7  & 85.7 & 41.8 & 19.5 & 71.4 & 42.9 & 5.1  & 85.7 & 48.7 & 41.4 & 0.0  \\
& DM       & 38.3 & 27.6 & 42.9 & 42.9 & 30.8 & 40.0 & 45.4 & 28.6 & 0.0  & 39.3 & 27.3 & 57.1 & 38.1 & 18.5 & 77.8 & 50.0 & NaN  & NaN  & 50.0 & 50.0 & NaN  \\
& MS       & 47.6 & 42.0 & 50.0 & 48.8 & 39.0 & 50.0 & 45.1 & 42.1 & 50.0 & 41.6 & 34.9 & 37.5 & 50.0 & NaN  & 100.0& 50.0 & NaN  & NaN  & 50.0 & NaN  & NaN  \\
& VESTA    & 39.1 & 22.3 & 50.0 & 41.1 & 36.6 & 57.1 & 41.1 & 36.6 & 57.1 & 30.3 & 40.8 & 83.3 & 33.5 & 10.5 & 75.0 & 44.1 & 13.3 & 33.3 & 50.0 & 50.0 & NaN  \\
\cmidrule(lr){2-23}
& \textit{Avg.} & 39.1 & 27.2 & 43.6 & 42.1 & 31.6 & 50.3 & 41.7 & 29.7 & 35.3 & 35.9 & 23.9 & 62.7 & 40.6 & 14.1 & 79.1 & 46.7 & 16.7 & 59.7 & 48.7 & ---  & ---  \\
\midrule
Origin & Origin & 47.7 & NaN  & 100.0& 47.2 & 31.7 & 50.0 & 32.7 & 33.7 & 83.3 & 34.3 & 23.5 & 60.0 & 48.0 & 24.5 & 50.0 & 48.6 & 50.0 & 100.0& 50.0 & NaN  & NaN  \\
\midrule
\multirow{5}{*}{Code}
  & Pymatgen & 2.0  & 2.0  & 70.0 & 13.5 & 2.0  & 80.0 & 2.2  & 2.0  & 75.0 & 4.9  & 2.0  & 63.2 & 2.0  & 2.0  & 65.0 & 10.7 & 3.0  & 68.8 & 13.5 & 2.0  & 81.3 \\
& MP       & 2.0  & 2.0  & 75.0 & 2.0  & 2.0  & 80.0 & 2.0  & 2.0  & 85.0 & 2.0  & 2.0  & 100.0& 2.3  & 2.0  & 75.0 & 3.4  & 3.0  & 90.0 & 2.0  & 2.0  & 90.0 \\
& OQMD     & 2.0  & 2.0  & 90.0 & 2.0  & 2.0  & 20.0 & 2.0  & 2.0  & 50.0 & 2.0  & 2.0  & 55.0 & 2.0  & 2.0  & 90.0 & 3.1  & NaN  & 100.0& 2.0  & NaN  & 100.0\\
& OPTIMADE & 2.0  & 2.0  & 30.0 & 2.0  & 2.0  & 20.0 & 2.0  & 2.0  & 25.0 & 2.0  & 2.0  & 15.0 & 2.0  & 2.0  & 80.0 & 3.0  & 3.0  & 70.0 & 2.0  & 2.0  & 90.0 \\
\cmidrule(lr){2-23}
& \textit{Avg.} & 2.0  & 2.0  & 66.3 & 4.9  & 2.0  & 50.0 & 2.1  & 2.0  & 58.8 & 2.7  & 2.0  & 58.3 & 2.1  & 2.0  & 77.5 & 5.1  & 3.0  & 82.2 & 4.9  & 2.0  & 90.3 \\
\midrule
\multicolumn{2}{l}{\textbf{Overall Avg.}}
  & 29.6 & 14.6 & 70.0 & 31.4 & 21.8 & 50.1 & 25.5 & 21.8 & 59.1 & 24.3 & 16.5 & 60.3 & 30.2 & 13.5 & 68.9 & 33.5 & 23.2 & 80.6 & 34.5 & NaN  & NaN  \\
\bottomrule
\end{tabular}%
}
\end{table*}
\section{Appendix I. Evaluator Reliability Details}
\label{sec:reliability-detail}

\paragraph{GUI getter reliability.}
\noindent
We validate the automated GUI evaluators with an $n\times n$ cross-evaluation
protocol over collected trajectory final states ($n=20$ per domain). For each
domain, every final state is evaluated against every task configuration from
the same domain. This creates positive matching pairs and non-matching pairs,
allowing us to measure both successful detection and cross-task false positives.

We report two F1 measures. \textbf{Score F1} is computed over binary
sub-criterion-level partial-credit labels, while \textbf{SR-F1} is computed
over task-level binary success labels for each task--state pair. The
per-domain aggregated results are reported in Table~\ref{tab:overall}, and the
per-task macro breakdown is shown in Table~\ref{tab:per_task_detail}.

For DigitalMicrograph, several final-state cues are intentionally excluded from
the filtered reliability calculation because they are valid benchmark getters
but not discriminative under cross-task evaluation. These include short OCR
tokens that commonly appear across microscopy images (\texttt{nm},
\texttt{scale}, \texttt{rotate}), geometry checks shared by multiple drawing or
ROI tasks, and file-import labels with the same filename stem but different
extensions (\texttt{dm4.dm3}/\texttt{dm4.dm4} and
\texttt{dm5.dm3}/\texttt{dm5.dm5}). This filtering affects only the reliability
audit and does not change the benchmark evaluator or task scores.

Most residual disagreements arise from visual edge cases, including OCR
confusion, final screenshots with transient dialogs or selection highlights,
and rendering-dependent pixel thresholds. The modular getter--metric design
keeps these errors interpretable: each disagreement can be traced to a specific
sub-criterion rather than to an opaque task-level pass/fail judgment.

\begin{table}[t]
\centering
\caption{Origin aesthetic-judge validation. C/A/T denote visual correctness,
aesthetic quality, and task completeness. All judges parse all 43 images.}
\label{tab:origin-judge-validation}
\scriptsize
\setlength{\tabcolsep}{2.3pt}
\renewcommand{\arraystretch}{1.04}
\resizebox{\linewidth}{!}{%
\begin{tabular}{lccccc}
\toprule
\textbf{Judge} & \textbf{Avg C/A/T} & \textbf{Parsed} & \textbf{Pearson} & \textbf{95\% CI} & \textbf{Spearman} \\
\midrule
GPT-5.4 & 4.33/4.15/4.38 & 43/43 & 0.513 & [-0.180, 0.826] & 0.348 \\
Doubao-seed-1-8 & 4.61/4.43/4.80 & 43/43 & 0.677 & [-0.017, 0.896] & 0.402 \\
Gemini-3.1-pro & 4.60/4.25/4.68 & 43/43 & 0.694 & [0.033, 0.891] & 0.346 \\
Qwen3-VL-235B & 4.81/4.46/4.92 & 43/43 & 0.737 & [0.104, 0.928] & 0.518 \\
\bottomrule
\end{tabular}%
}
\end{table}

\paragraph{Origin aesthetic-judge reliability.}
\noindent Origin tasks require a second reliability analysis because figure
quality is not fully captured by file existence. Table~\ref{tab:origin-judge-validation}
therefore reports both parsing stability and agreement with human ratings for
the aesthetic judge. All four candidate judges successfully parse all 43
figures, so the comparison is not affected by formatting failures.

\noindent The table also reveals a calibration difference across judges.
Qwen3-VL-235B obtains the highest Pearson and Spearman correlations, but its
raw C/A/T scores are visibly more saturated and lenient. Gemini-3.1-pro is
slightly more conservative while retaining similar human agreement. A Williams
dependent-correlation test between Qwen3-VL-235B and Gemini-3.1-pro gives
$\Delta r=0.043$, $t=0.582$, and $p=0.569$, indicating that their correlations
are not statistically distinguishable at this sample size. We therefore use
Gemini-3.1-pro as the aesthetic scorer, while emphasizing that this choice only
affects the secondary aesthetic score and never changes Origin SR.

\noindent Together, Tables~\ref{tab:per_task_detail} and
\ref{tab:origin-judge-validation} validate the two evaluator components used by
the benchmark: deterministic state getters for GUI tasks and a calibrated
human-aligned judge for generated scientific figures.

\noindent The residual disagreements are concentrated in visually ambiguous GUI
states: OCR confuses similar characters in small labels, and pixel/SSIM thresholds
occasionally react to anti-aliasing or rendering differences. These cases are rare
and do not change the conclusion that the automated getters closely track expert
annotations across the evaluated GUI domains.

\begin{figure*}[htbp]
  \centering
  \begin{minipage}[t]{0.30\linewidth}
    \centering
    \includegraphics[width=\linewidth]{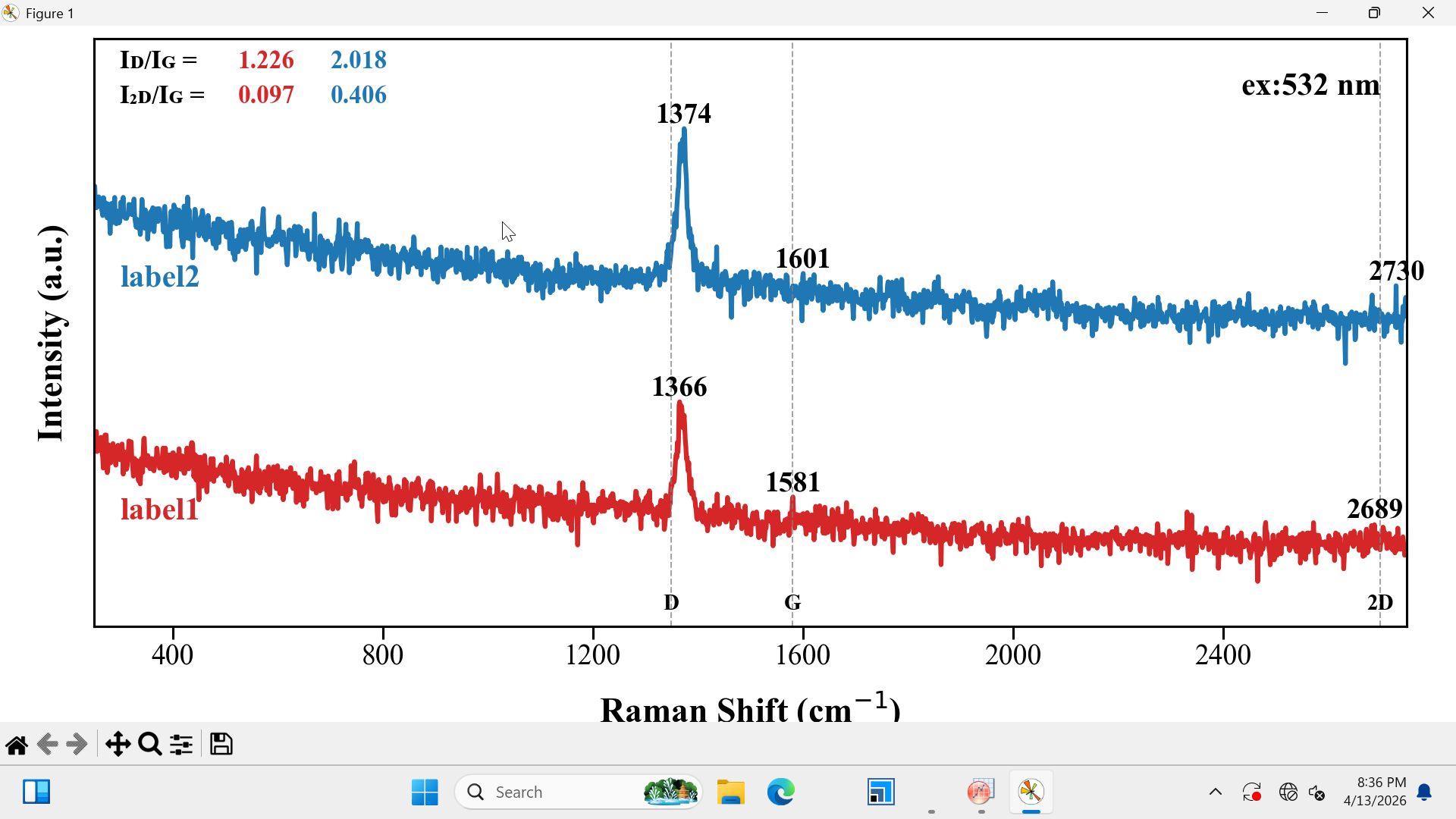}\\[2pt]
    \small $s = 1.00$
  \end{minipage}\hfill
  \begin{minipage}[t]{0.30\linewidth}
    \centering
    \includegraphics[width=\linewidth]{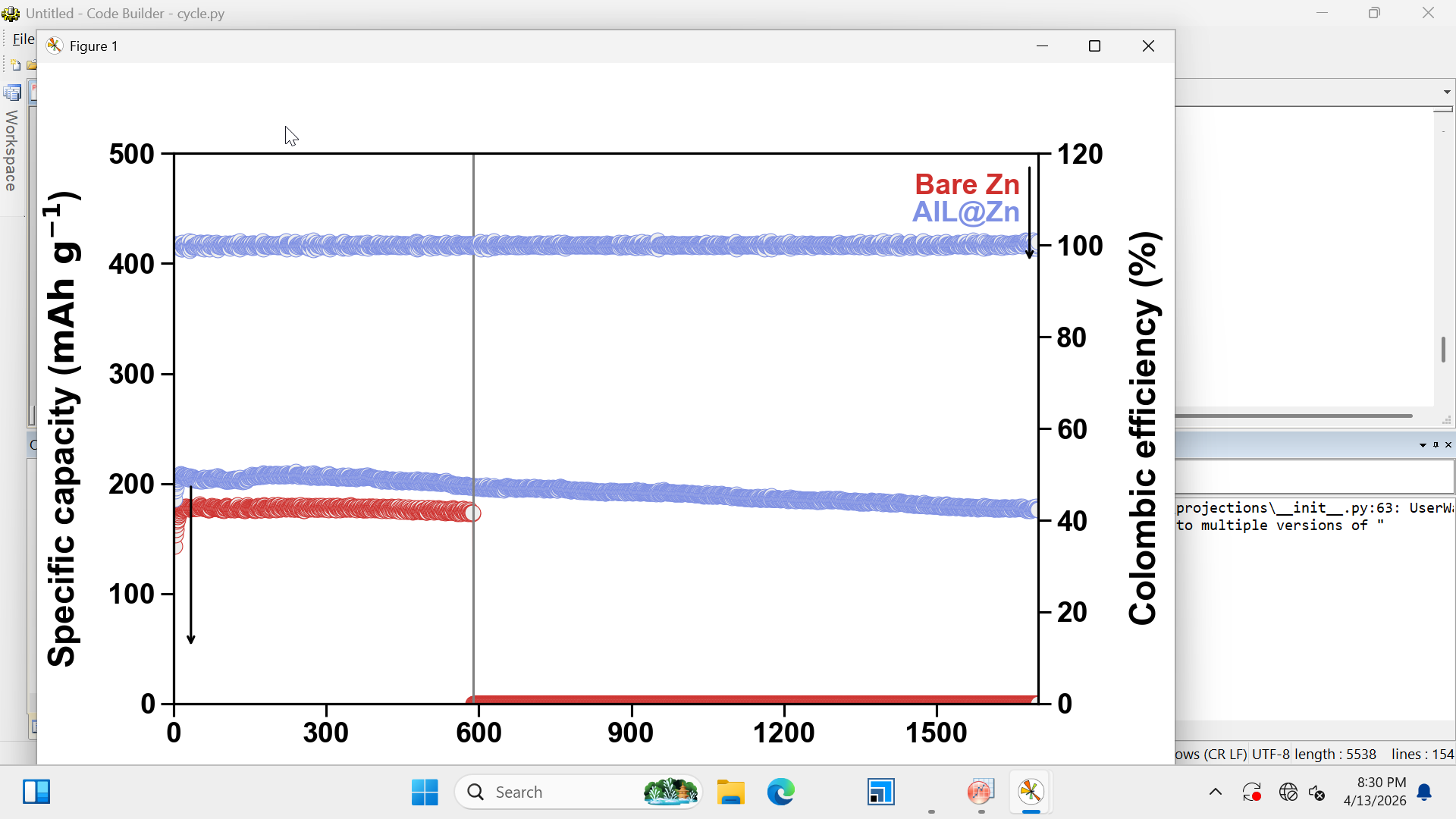}\\[2pt]
    \small $s = 1.00$
  \end{minipage}\hfill
  \begin{minipage}[t]{0.30\linewidth}
    \centering
    \includegraphics[width=\linewidth]{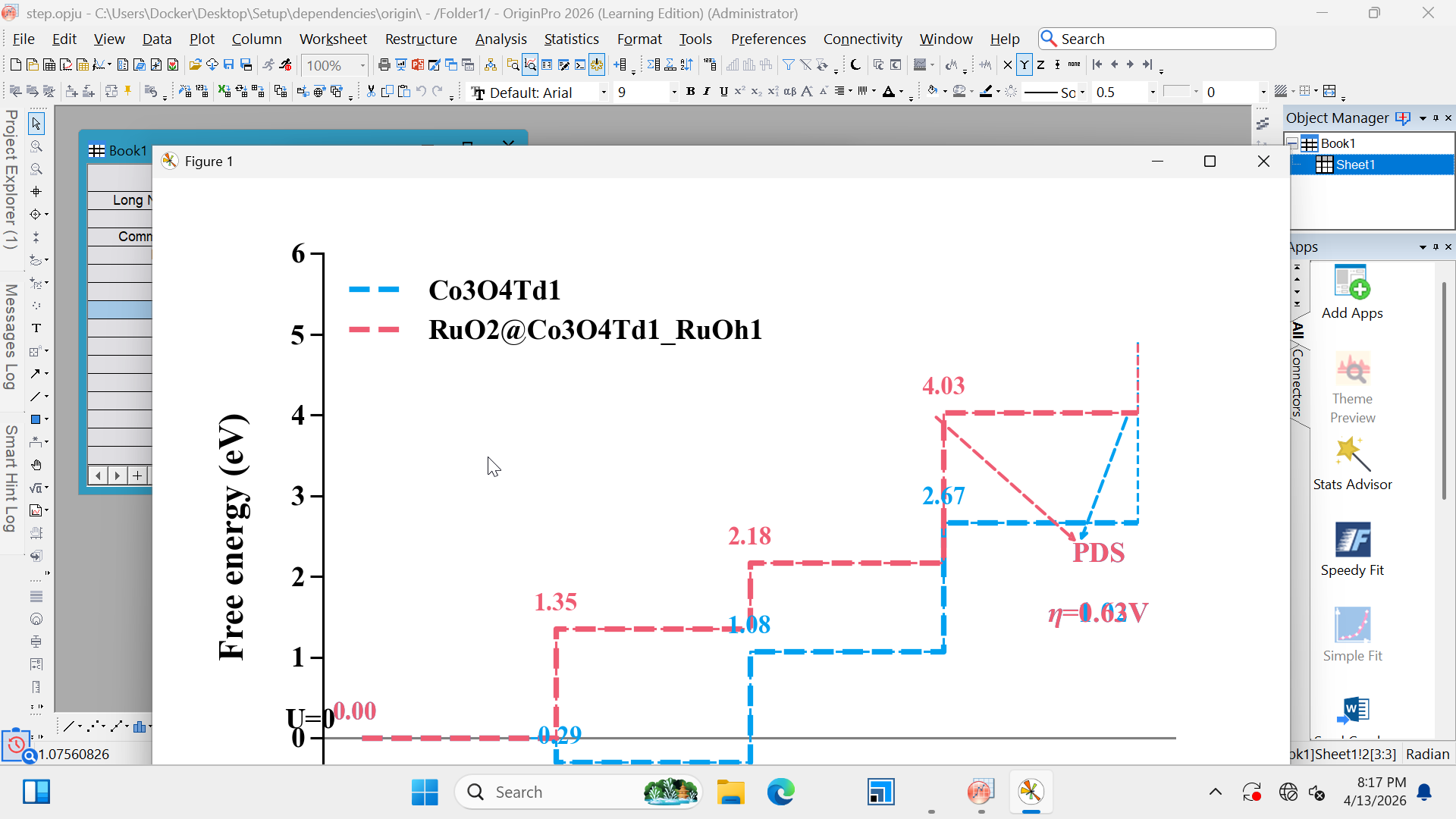}\\[2pt]
    \small $s = 0.98$
  \end{minipage}

  \vspace{8pt}

  \begin{minipage}[t]{0.30\linewidth}
    \centering
    \includegraphics[width=\linewidth]{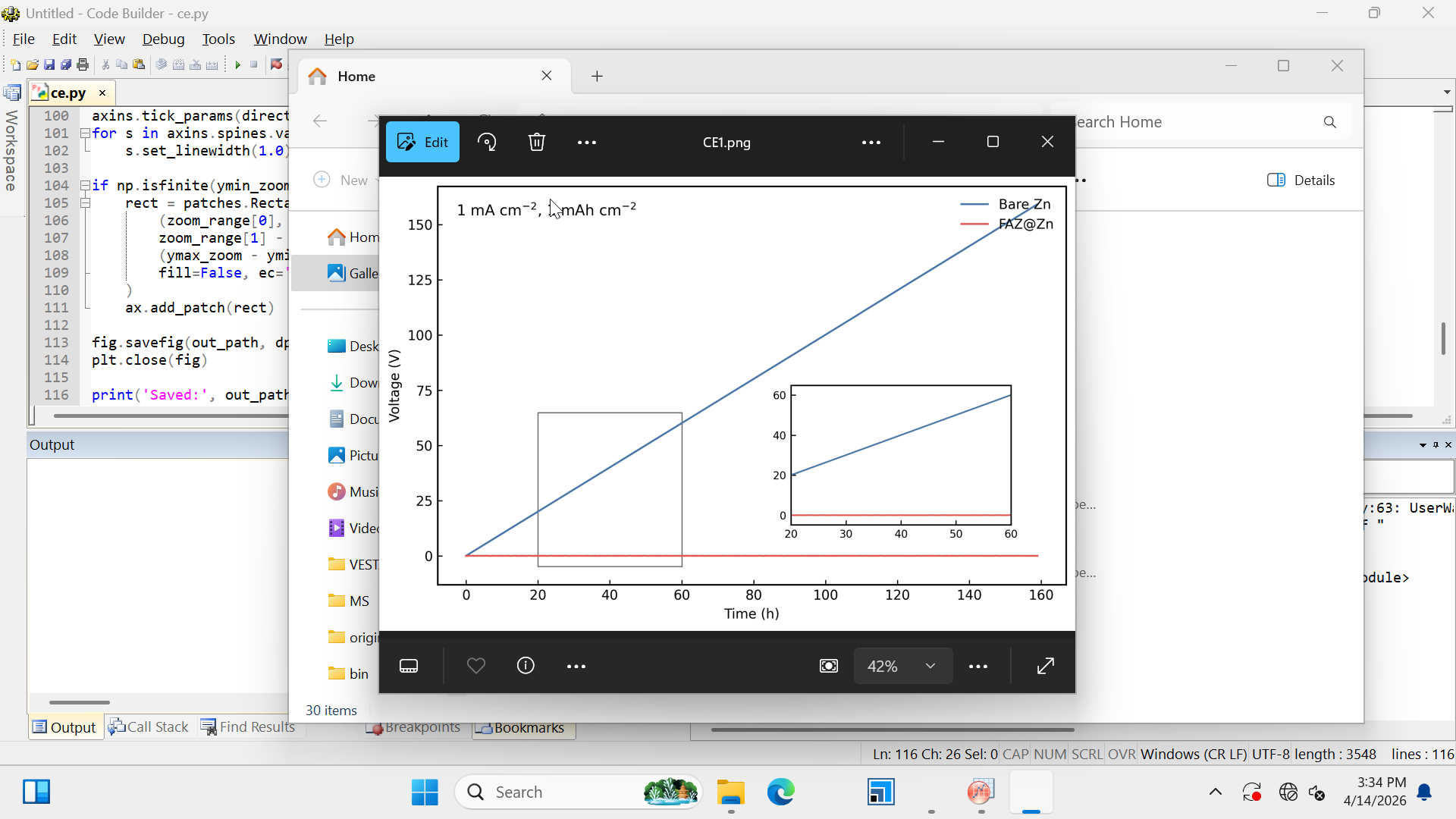}\\[2pt]
    \small $s = 0.70$
  \end{minipage}\hfill
  \begin{minipage}[t]{0.30\linewidth}
    \centering
    \includegraphics[width=\linewidth]{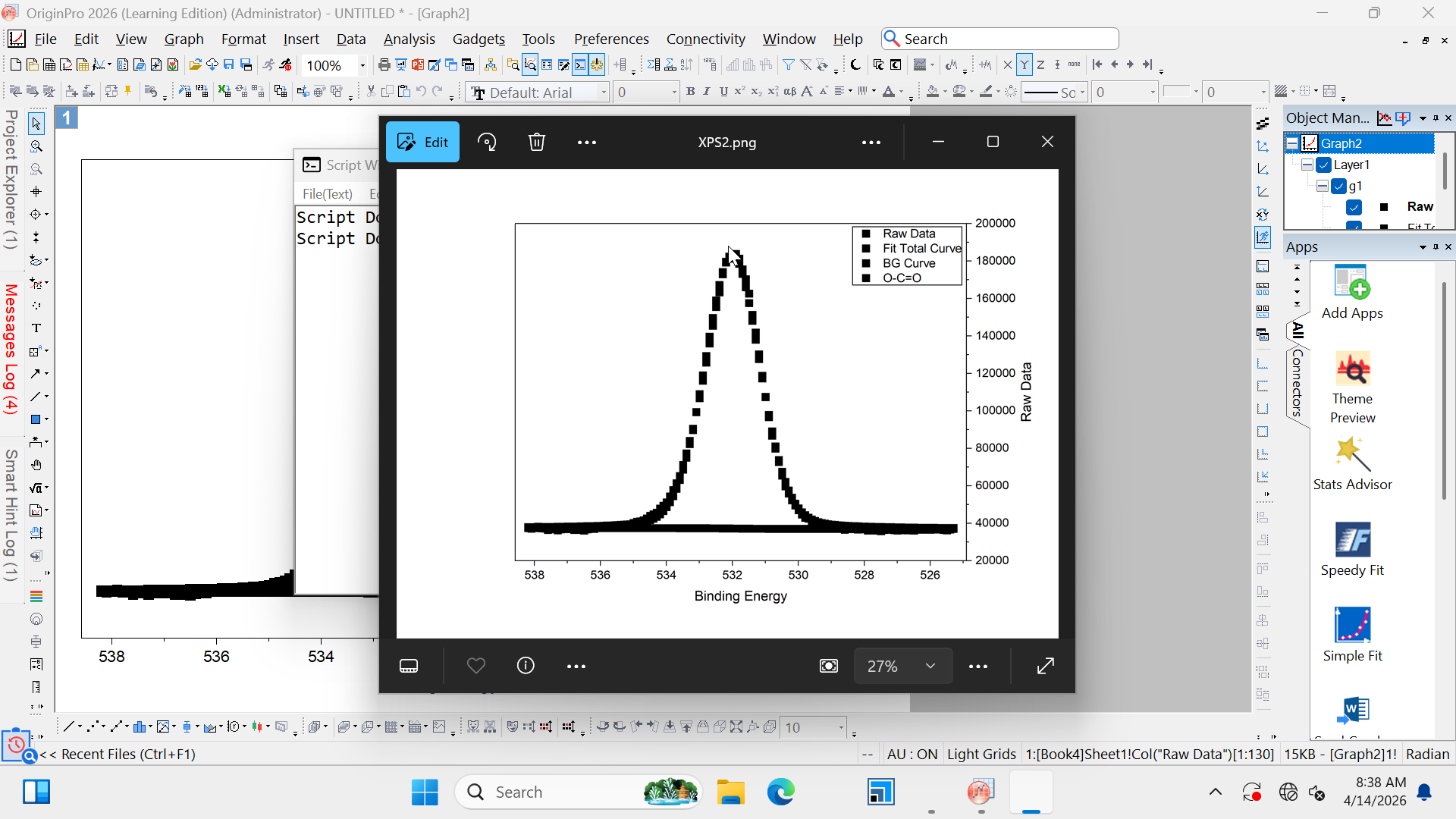}\\[2pt]
    \small $s = 0.40$
  \end{minipage}\hfill
  \begin{minipage}[t]{0.30\linewidth}
    \centering
    \includegraphics[width=\linewidth]{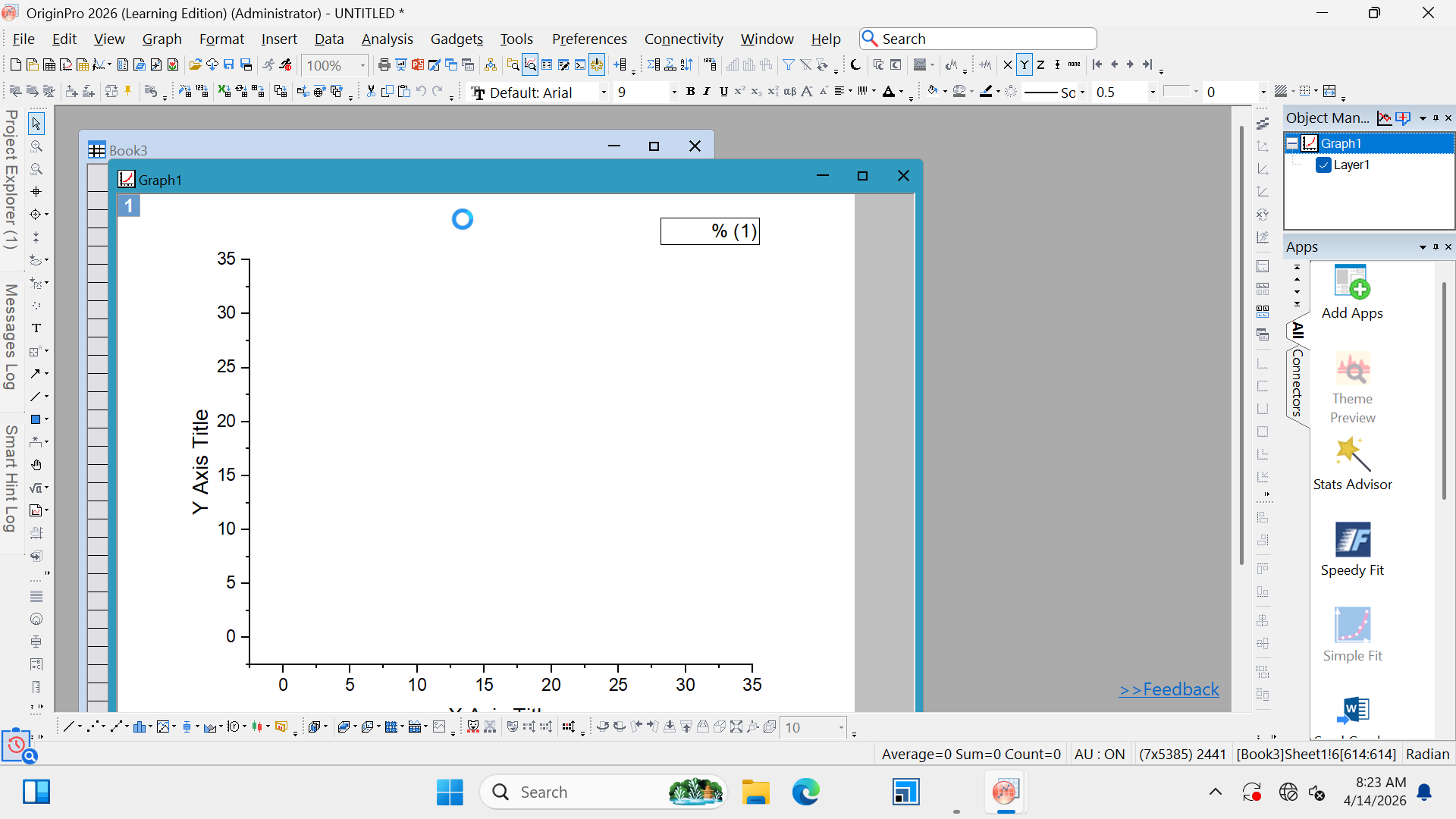}\\[2pt]
    \small $s = 0.00$ (not generated)
  \end{minipage}

  \caption{Representative Origin figures generated by Claude Sonnet 4.6 with
\textbf{script+hint} (top) and \textbf{no\_script+no\_hint} (bottom).
Scores are computed as $s=(C+A+T)/(3\mathbin{\times}5)$.
Template scripts produce high-quality outputs ($s\geq 0.98$), whereas
GUI-only execution reduces quality ($s\leq 0.70$) or fails ($s=0$).}
\label{fig:origin-examples}
\end{figure*}

\section{Appendix J. Efficiency Results}
\label{sec:efficiency}
\noindent Table~\ref{tab:results-efficiency} reports the efficiency results.
\textbf{Efficiency} metrics capture the number of agent steps consumed during
task execution (maximum 50 steps; fewer is better). \textbf{Steps} reports the
mean step count across all episodes; \textbf{Steps*} reports the mean step count
on successful episodes only (SR\,=\,1), removing the confounding effect of
premature termination on overall step counts. \textbf{FTR} quantifies the
fraction of agent-terminated episodes where SR = 0, revealing systematic
failure modes. GUI tasks typically consume 35--42 steps due to exploratory
interaction requirements such as locating UI elements and reading dialog boxes,
whereas code tasks use only 1--4 steps because each step corresponds to a
complete script execution. 

\noindent As shown in Table~\ref{tab:results-efficiency}, high FTR
values (40--70\%) on GUI tasks indicate a persistent premature-termination
failure mode across frontier models, suggesting that agents often struggle to
distinguish true task completion from temporary pauses in the workflow.

\noindent Let $\mathcal{E}$ be the set of evaluated episodes in a domain, $T_e$
the number of executed steps in episode $e$, $s_e$ the final success indicator,
and $d_e$ an indicator that the episode ended with an agent-issued
\texttt{DONE}. We compute
\begin{equation}
  \mathrm{Steps}=\frac{1}{|\mathcal{E}|}\sum_{e\in\mathcal{E}}T_e,
  \qquad
  \mathrm{Steps}^{*}=
  \frac{\sum_{e\in\mathcal{E}}s_eT_e}{\sum_{e\in\mathcal{E}}s_e},
\end{equation}
with $\mathrm{Steps}^{*}$ left undefined when no episode succeeds. False
termination is measured only among self-terminated runs:
\begin{equation}
  \mathrm{FTR}=
  \frac{\sum_{e\in\mathcal{E}} d_e\,\mathbb{I}[s_e=0]}
       {\sum_{e\in\mathcal{E}} d_e}.
\end{equation}
This separation is important because a low raw step count can be misleading:
an agent may appear efficient simply because it declares completion before
performing the required scientific operation. Steps* and FTR therefore provide
complementary views of efficiency and reliability.

\section{Appendix K. Origin Ablation Output Examples}
\label{sec:origin-examples}

\noindent
Origin tasks use two independent evaluation signals. Binary success rate
indicates whether the expected image exists at the designated VM path. For each
generated figure, a judge assigns visual correctness ($C$), aesthetic quality
($A$), and task completeness ($T$) scores in $[1,5]$:
$s=\frac{C+A+T}{3\times 5}$.
If no image is generated, $SR=0$ and $s=0$. The quality score does not affect
SR, but distinguishes low-quality figures from publication-ready outputs.

\noindent
Figure~\ref{fig:origin-examples} shows representative outputs from Claude
Sonnet 4.6. \textbf{script+hint} provides Code Builder guidance and plotting
templates; \textbf{no\_script+hint} removes only the template; and
\textbf{no\_script+no\_hint} removes both supports. The latter often causes
incorrect axes, missing labels, weak formatting, or execution failure.

\noindent
File existence alone is insufficient because a generated figure may still
contain scientific defects, such as reversed axes, missing annotations, or
inconsistent styling. We therefore use $SR$ to measure operational success and
$s$ to diagnose scientific presentation quality.

\noindent
The ablation therefore separates artifact generation from scientific figure
quality. An agent may create the required file without reproducing the intended
plotting conventions, whereas template scripts substantially improve scaling,
annotations, and formatting.

\section{Appendix L. Evaluation Scoring Criteria Examples}
\label{sec:avantage-case}

\begin{tcolorbox}[
colback=white,
colframe=black,
boxrule=0.45pt,
arc=0pt,
left=3pt,
right=3pt,
top=3pt,
bottom=3pt,
title=\textbf{Algorithm 1: Instruction-to-Score Evaluation Framework},
fonttitle=\small,
coltitle=black,
colbacktitle=white]
\small
\begin{algorithmic}[1]
\REQUIRE Instruction $g$, initialization program $c$, budget $T_{\max}$,
criteria $\mathcal{C}=\{(g_i,m_i,w_i,y_i)\}_{i=1}^{n}$
\STATE Execute $c$ to obtain the initial VM state $v_0$.
\FOR{$t=0,\ldots,T_{\max}-1$}
\STATE Build observation $o_t$ from the instruction and current VM state.
\STATE Agent outputs \texttt{COMMAND}, \texttt{WAIT}, \texttt{DONE}, or \texttt{FAIL}.
\IF{\texttt{COMMAND}}
\STATE Execute the action and update $v_{t+1}$.
\ELSIF{\texttt{WAIT}}
\STATE Wait for environment stabilization.
\ELSIF{\texttt{DONE} or \texttt{FAIL}}
\STATE Set the final state to $v_T$ and terminate.
\ENDIF
\ENDFOR
\IF{the budget is exhausted}
\STATE Set $v_T$ to the current VM state.
\ENDIF
\FOR{$(g_i,m_i,w_i,y_i)\in\mathcal{C}$}
\STATE Extract evidence $z_i=g_i(v_T)$ and compute
$r_i=m_i(z_i,y_i)\in[0,1]$.
\ENDFOR
\STATE Return
$\mathrm{Score}=\frac{\sum_i w_i r_i}{\sum_i w_i}$,
$\mathrm{SR}=\mathbb{I}[\forall i,\ r_i=1]$,
and the criterion-level trace.
\end{algorithmic}
\end{tcolorbox}

\noindent This section demonstrates the evaluation framework through an Avantage XPS example and GUI pixel-detection methods.

\subsection{Avantage Task: Step-by-Step Evaluation}

We trace the full instruction-to-output evaluation pipeline for a representative
Avantage task:
\textit{``Reverse the binding-energy axis of the loaded XPS spectrum.''}
The framework is action-sequence agnostic: it evaluates the final scientific
state rather than requiring the agent to follow a fixed menu path. For the
axis-reversal task, the getter is
\texttt{check\_energy\_axis\_reversed}: it crops the axis region, applies OCR,
and checks whether binding-energy labels decrease from left to right. The metric
is \texttt{exact\_match}; the sub-criterion receives 1 if the axis is confirmed
reversed and 0 otherwise. The same framework is used for the harder multi-step
case below, where each GUI milestone is scored independently.

\begin{center}
\small
\setlength{\tabcolsep}{3pt}
\renewcommand{\arraystretch}{1.05}
\begin{tabular}{cL{0.52\linewidth}L{0.26\linewidth}}
\toprule
\textbf{Pt.} & \textbf{Sub-criterion} & \textbf{Evidence} \\
\midrule
1 & Avantage imports \texttt{C1s Scan.VGD}. & title/OCR \\
1 & Window~A and window~B both contain the C1s spectrum. & window state \\
1 & \emph{Add Constant} dialog is opened for window~B. & dialog OCR \\
1 & Constant value equals \texttt{3333}. & parameter getter \\
1 & The transformed spectrum appears in window~B. & pixel/OCR state \\
1 & Windows are arranged vertically for comparison. & layout getter \\
1 & \texttt{stacked\_3333.vgp} is saved at the target path. & file check \\
\bottomrule
\end{tabular}
\end{center}
\subsection{Detailed Avantage Case Study}

We illustrate how the evaluator verifies each scoring sub-criterion using a hard Avantage task (\texttt{8ebfc15b}, 7\,pts), corresponding to the getter types listed in Table~\ref{tab:eval-gui-getters}.

\begin{figure*}[!htbp]
  \centering

  \begin{minipage}[t]{0.31\linewidth}
    \centering
    \includegraphics[
      width=\linewidth,
      height=3.0cm,
      keepaspectratio
    ]{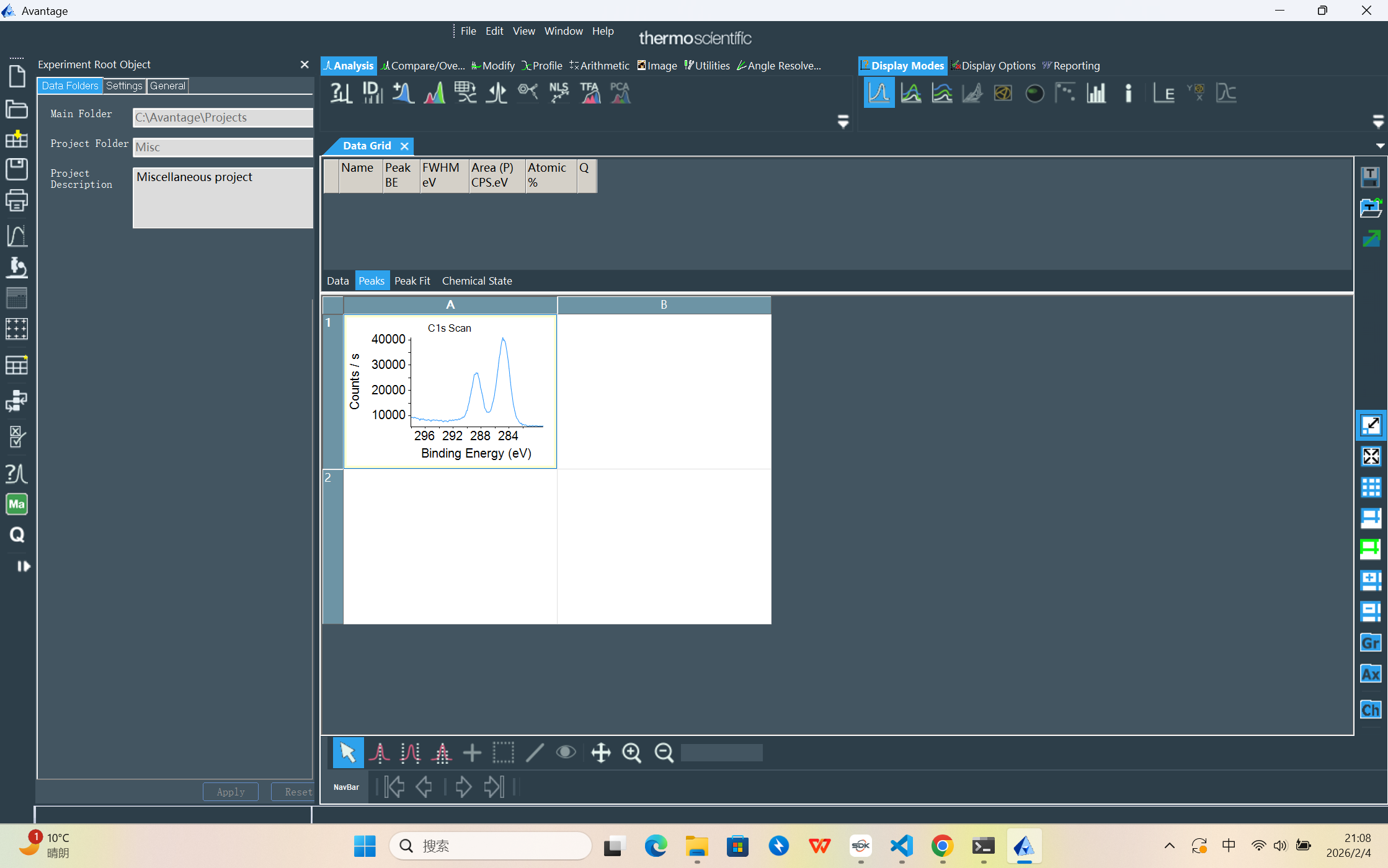}\\[-1pt]
    {\small\textbf{(a) Open file}}\\[-1pt]
    {\scriptsize
    OCR confirms ``C1s Scan'' in window~A, while the saved file verifies the
    output path.}
  \end{minipage}
  \hfill
  \begin{minipage}[t]{0.31\linewidth}
    \centering
    \includegraphics[
      width=\linewidth,
      height=3.0cm,
      keepaspectratio
    ]{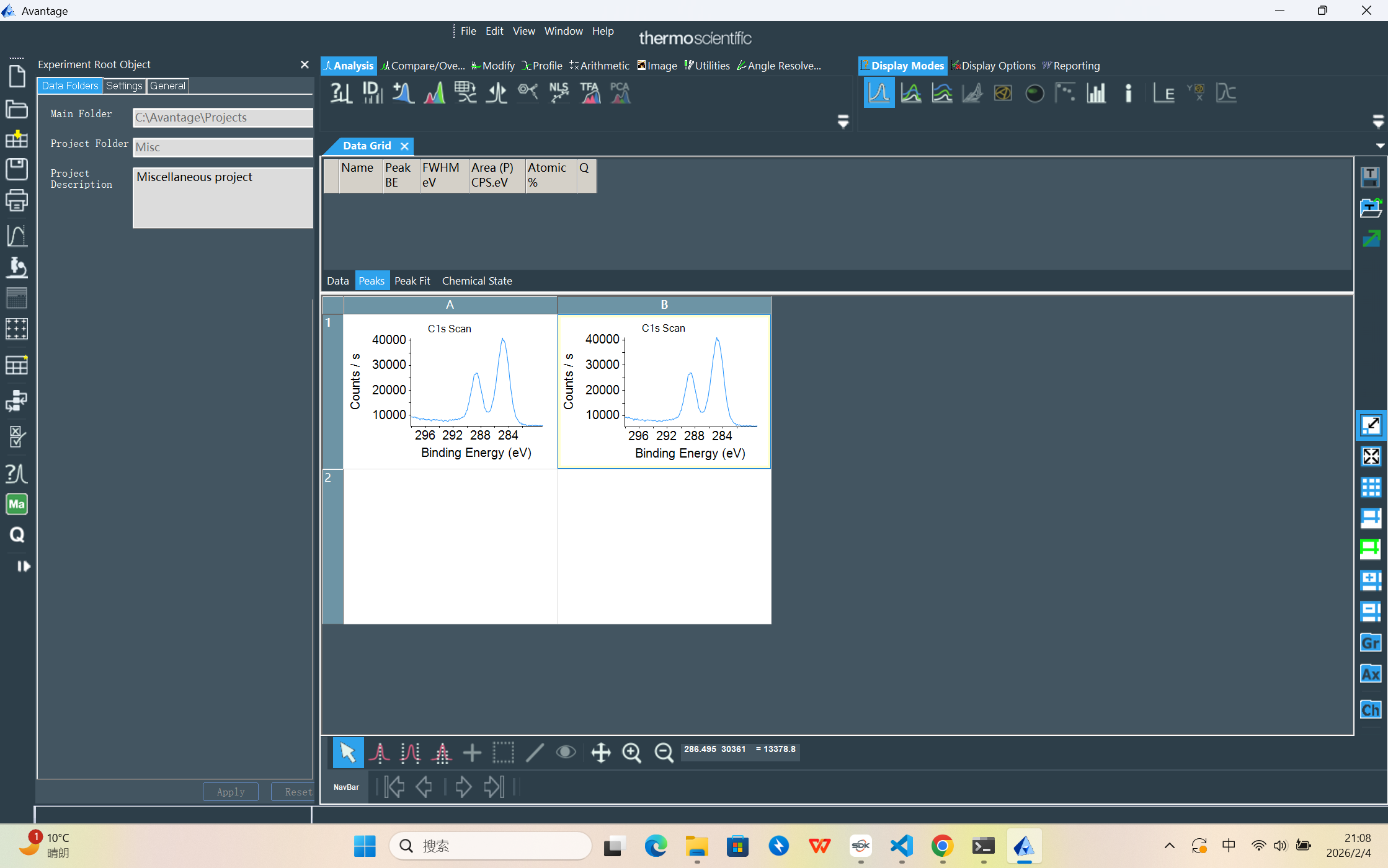}\\[-1pt]
    {\small\textbf{(b) Copy spectrum to window B}}\\[-1pt]
    {\scriptsize
    The evaluator checks whether the same XPS scan is present in both
    window~A and window~B.}
  \end{minipage}
  \hfill
  \begin{minipage}[t]{0.31\linewidth}
    \centering
    \includegraphics[
      width=\linewidth,
      height=3.0cm,
      keepaspectratio
    ]{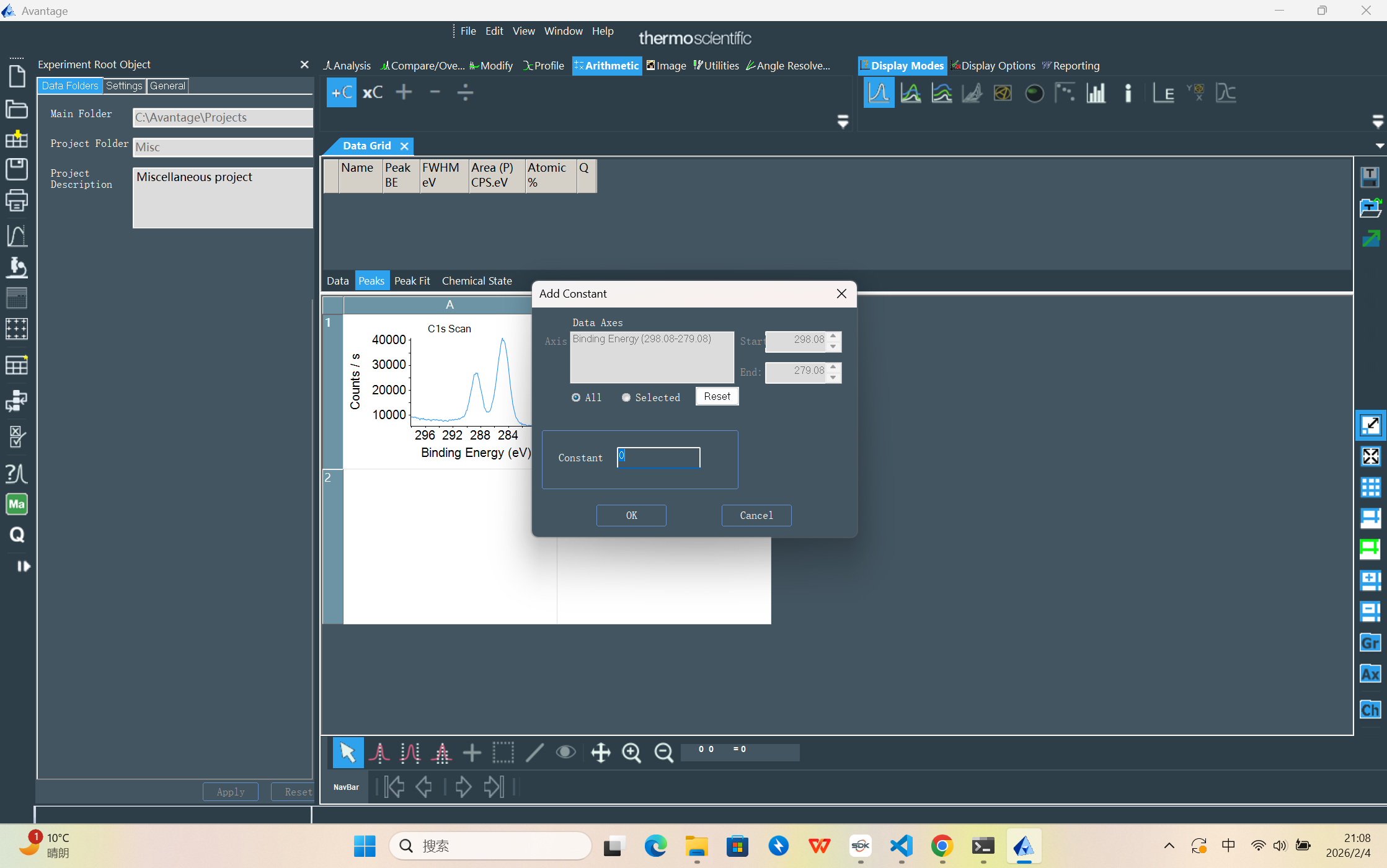}\\[-1pt]
    {\small\textbf{(c) Open Add Constant dialog}}\\[-1pt]
    {\scriptsize
    OCR verifies that the central dialog is titled ``Add Constant''.}
  \end{minipage}

  \vspace{7pt}

  \begin{minipage}[t]{0.31\linewidth}
    \centering
    \includegraphics[
      width=\linewidth,
      height=3.0cm,
      keepaspectratio
    ]{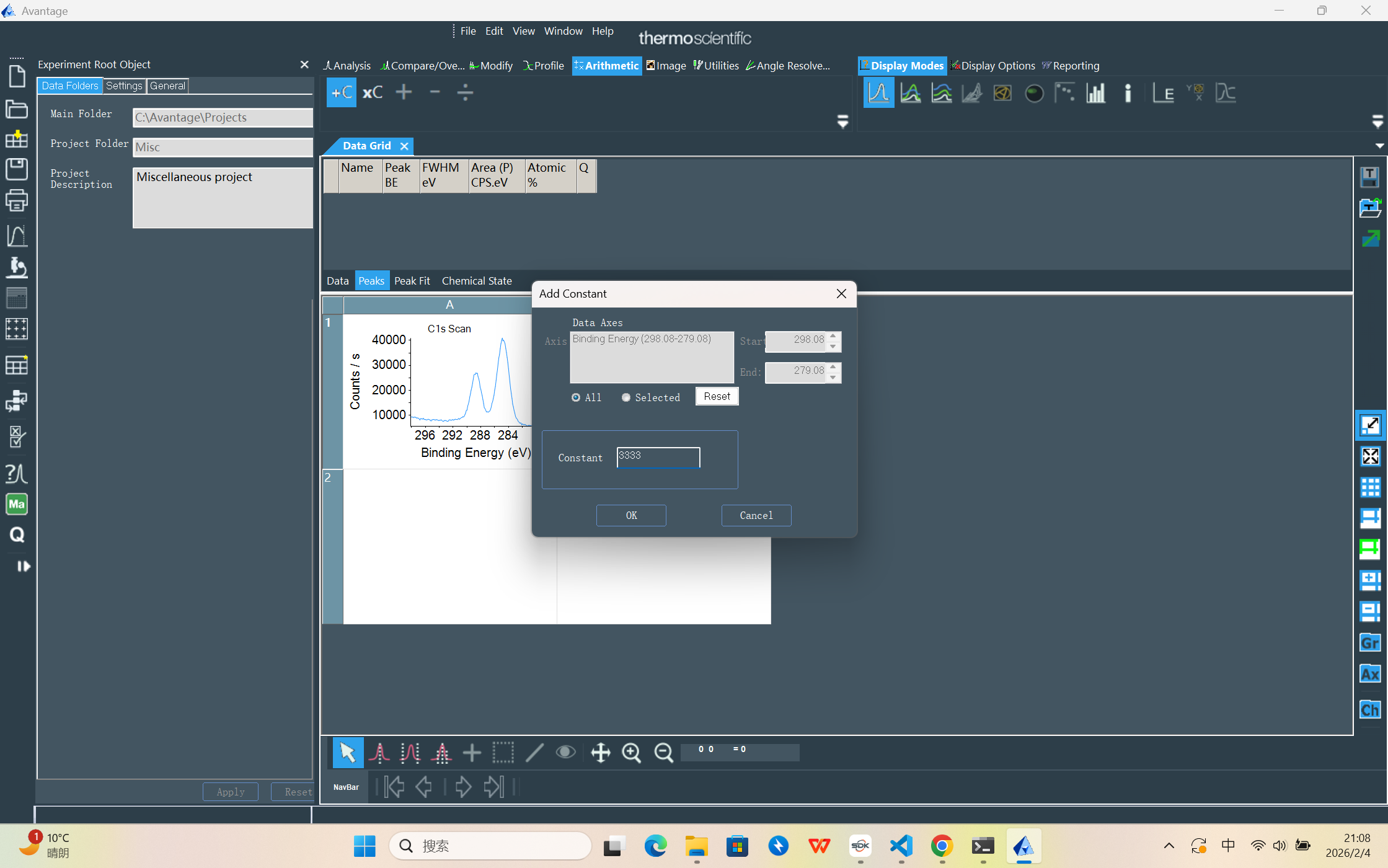}\\[-1pt]
    {\small\textbf{(d) Enter constant value 3333}}\\[-1pt]
    {\scriptsize
    The parameter getter extracts the ``Constant'' field and checks that it
    equals \textbf{3333}.}
  \end{minipage}
  \hspace{0.035\linewidth}
  \begin{minipage}[t]{0.31\linewidth}
    \centering
    \includegraphics[
      width=\linewidth,
      height=3.0cm,
      keepaspectratio
    ]{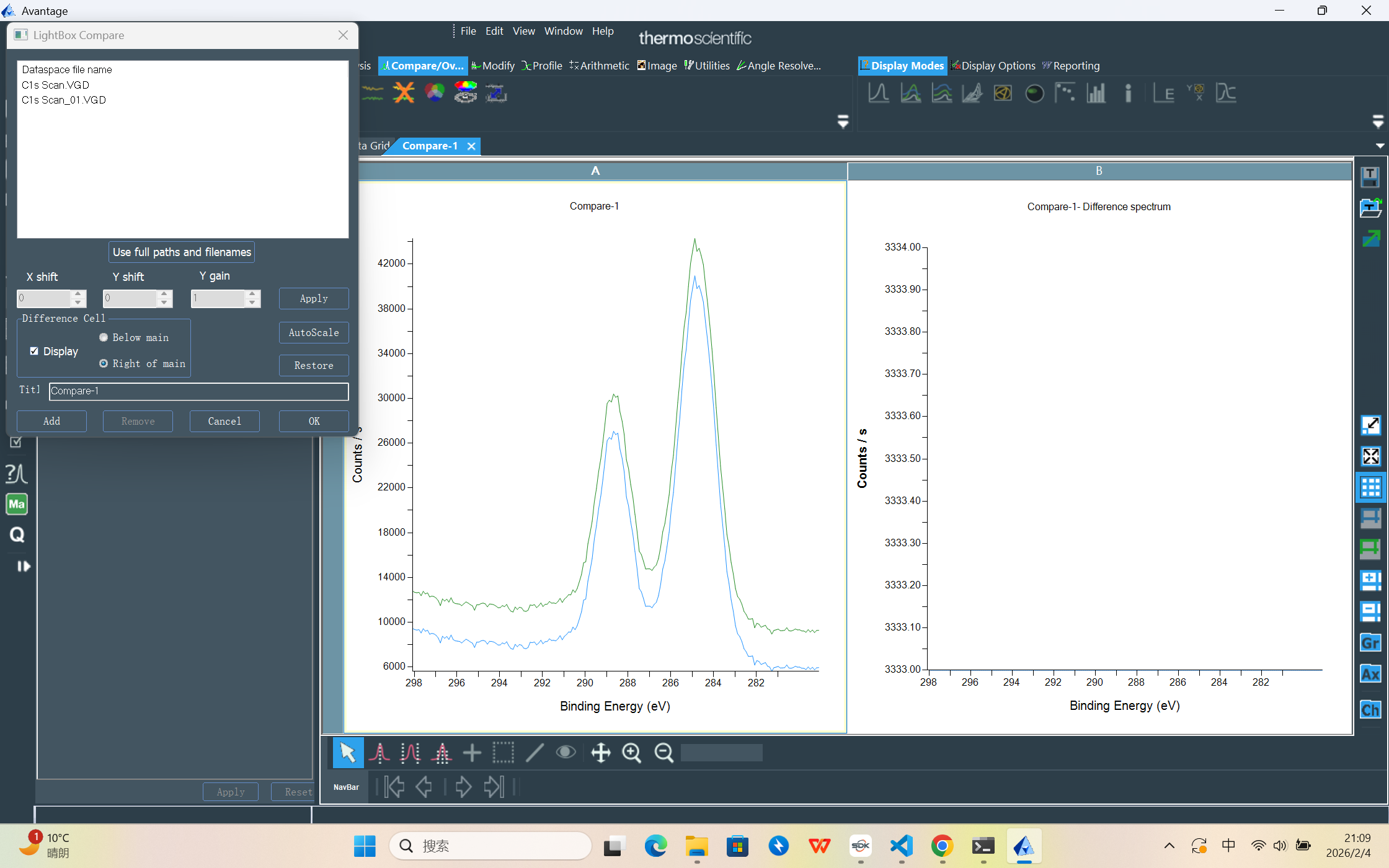}\\[-1pt]
    {\small\textbf{(e) Vertical comparison view}}\\[-1pt]
    {\scriptsize
    The LightBox panel must contain both \textbf{C1s\,Scan.VGD} and
    \textbf{C1s\,Scan\_01}.}
  \end{minipage}

  \caption{Execution trajectory and scoring sub-criteria for the Avantage task
  \texttt{8ebfc15b} (hard, 7\,pts). Each panel shows an intermediate GUI state
  validated by a domain-specific getter, including OCR, parameter extraction,
  file-existence, and application-state checks. The evaluator scores all
  sub-criteria independently at episode end, and the final task score is the
  fraction of passed criteria.}
  \label{fig:avantage-walkthrough}
\end{figure*}
\paragraph{Task instruction.}
Open Avantage and import \texttt{C1s\,Scan.VGD}; copy the spectrum to window~B; apply \emph{Add Constant} with value \textbf{3333} to the spectrum in window~B; arrange window~A and window~B vertically in comparison view; save the active document as \texttt{stacked\_3333.vgp}.

\noindent The seven criteria intentionally mix interface evidence and artifact
evidence. This prevents a shortcut such as saving an unchanged project from
receiving full credit, while still awarding partial credit for completed
milestones. The trajectory in Figure~\ref{fig:avantage-walkthrough} shows the
observable states used by these getters.

\noindent The example also illustrates why the benchmark scores final states
rather than command histories. An agent may use toolbar buttons, menu commands,
keyboard shortcuts, or scripting to reach the same Avantage state. These action
paths differ operationally, but the scientific requirement is identical: the
imported spectrum must be duplicated, transformed, arranged for comparison, and
saved. The getter-metric decomposition makes this equivalence explicit while
still exposing which milestone failed when the final result is incomplete.

\noindent Each of the seven sub-criteria contributes one point, and SR is
achieved only when all seven are satisfied. This makes near-miss
failures visible. For example, an episode that imports the file, opens the
dialog, enters the correct constant, and saves the project but fails to arrange
the two windows vertically receives partial credit rather than being collapsed
into a single failure. Conversely, an episode that only saves a file without
transforming the spectrum cannot obtain high credit because the intermediate
scientific states remain unsatisfied.

\noindent The same trace is used for debugging evaluator errors. If a model
receives a low score despite visually plausible progress, the per-criterion
record identifies whether the failure comes from OCR, parameter extraction,
layout recognition, or file persistence. During benchmark construction, this
trace allowed domain experts to audit ambiguous cases and adjust only the
affected getter threshold rather than rewriting the task definition. This is the
main reason the framework reports both aggregate scores and detailed
sub-criterion outcomes.

\begin{figure*}[!htbp]
  \centering

  \begin{minipage}[t]{0.31\linewidth}
    \centering
    \includegraphics[
      width=\linewidth,
      height=3.0cm,
      keepaspectratio
    ]{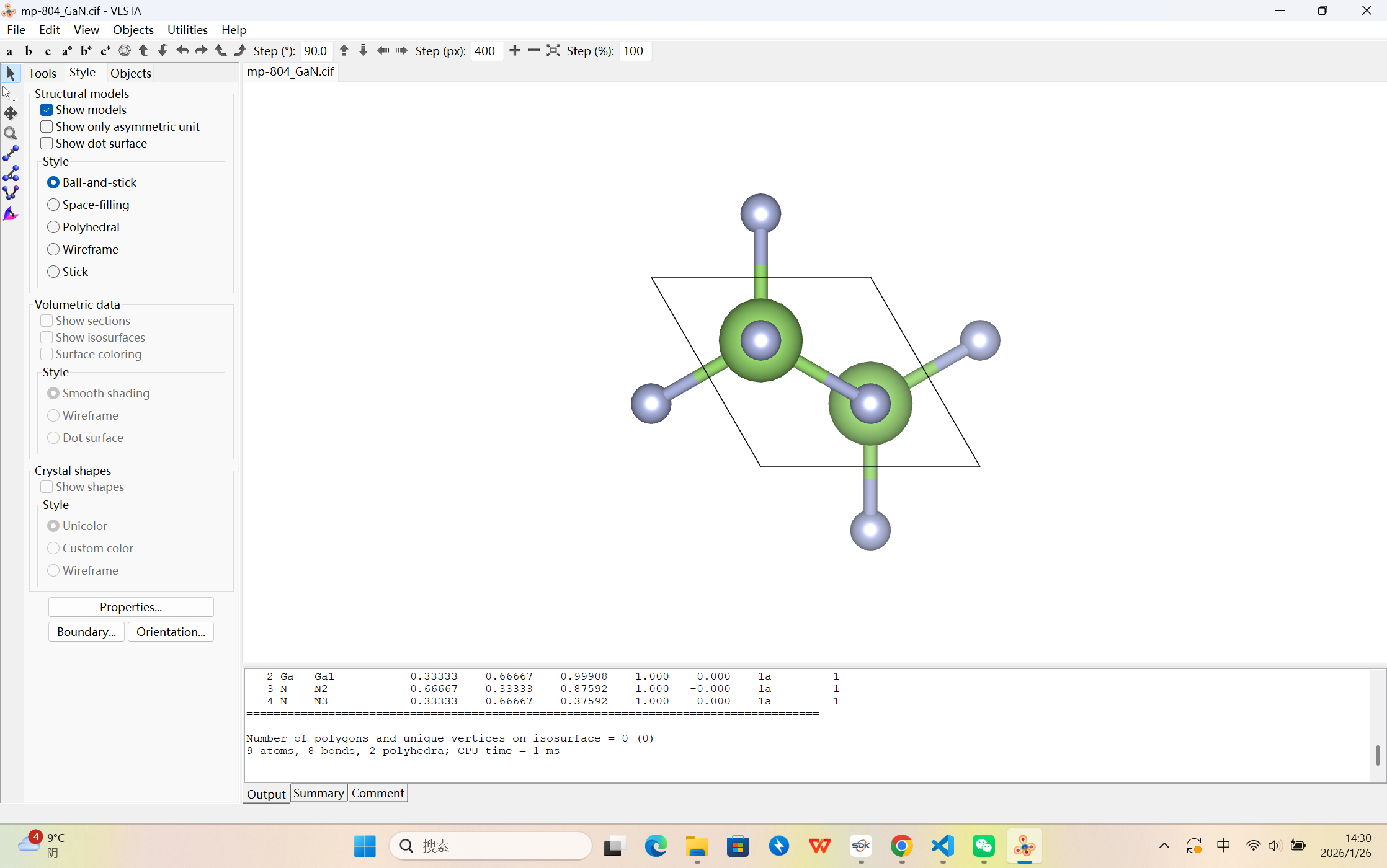}\\[-1pt]
    {\small\textbf{(a) Ball-and-stick style}}\\[-1pt]
    {\scriptsize
    The blue-filled radio button next to ``Ball-and-stick'' identifies the
    active rendering style.}
  \end{minipage}
  \hfill
  \begin{minipage}[t]{0.31\linewidth}
    \centering
    \includegraphics[
      width=\linewidth,
      height=3.0cm,
      keepaspectratio
    ]{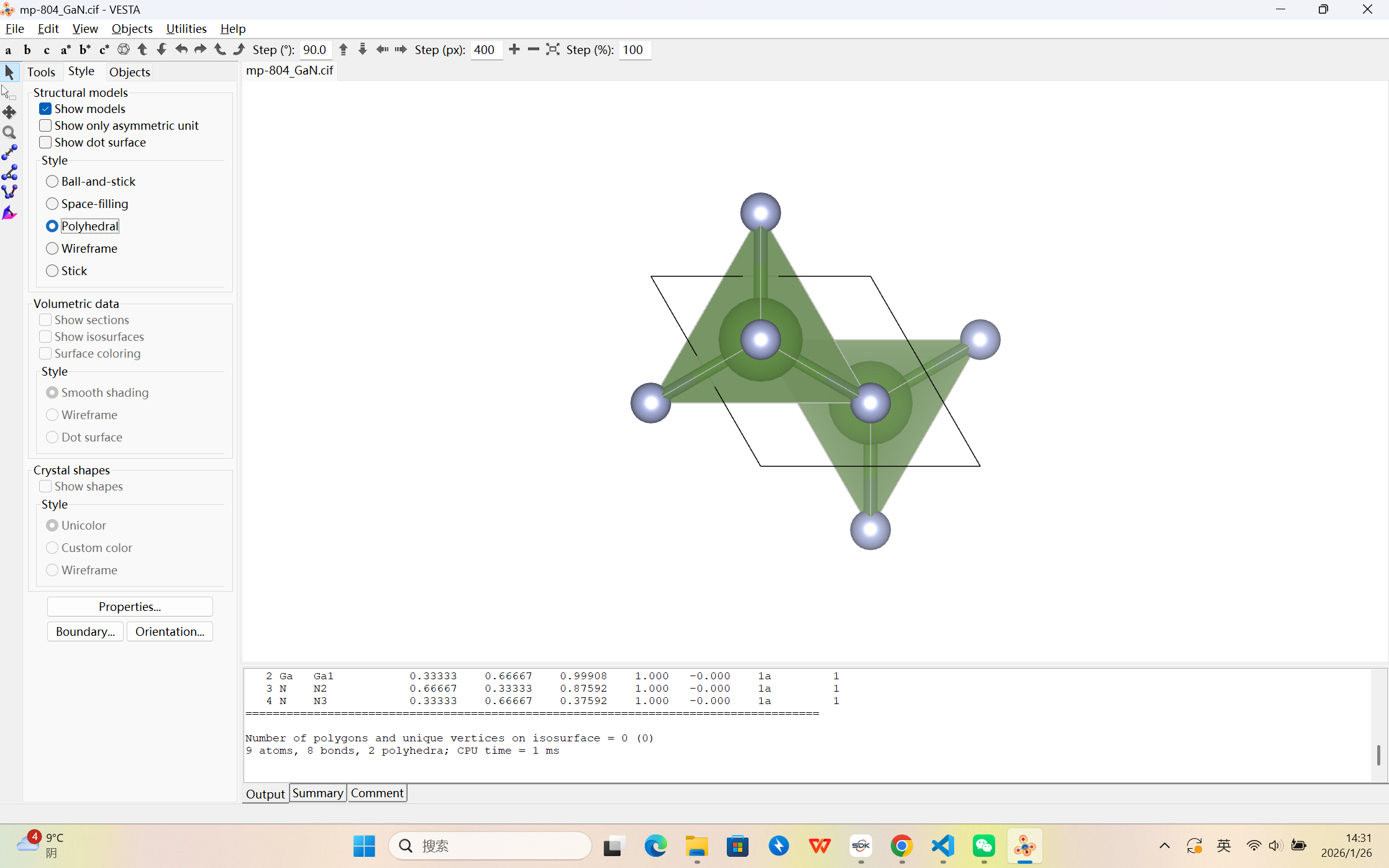}\\[-1pt]
    {\small\textbf{(b) Polyhedral style}}\\[-1pt]
    {\scriptsize
    The selected blue pixel cluster moves to ``Polyhedral'', confirming the
    style transition.}
  \end{minipage}
  \hfill
  \begin{minipage}[t]{0.31\linewidth}
    \centering
    \includegraphics[
      width=\linewidth,
      height=3.0cm,
      keepaspectratio
    ]{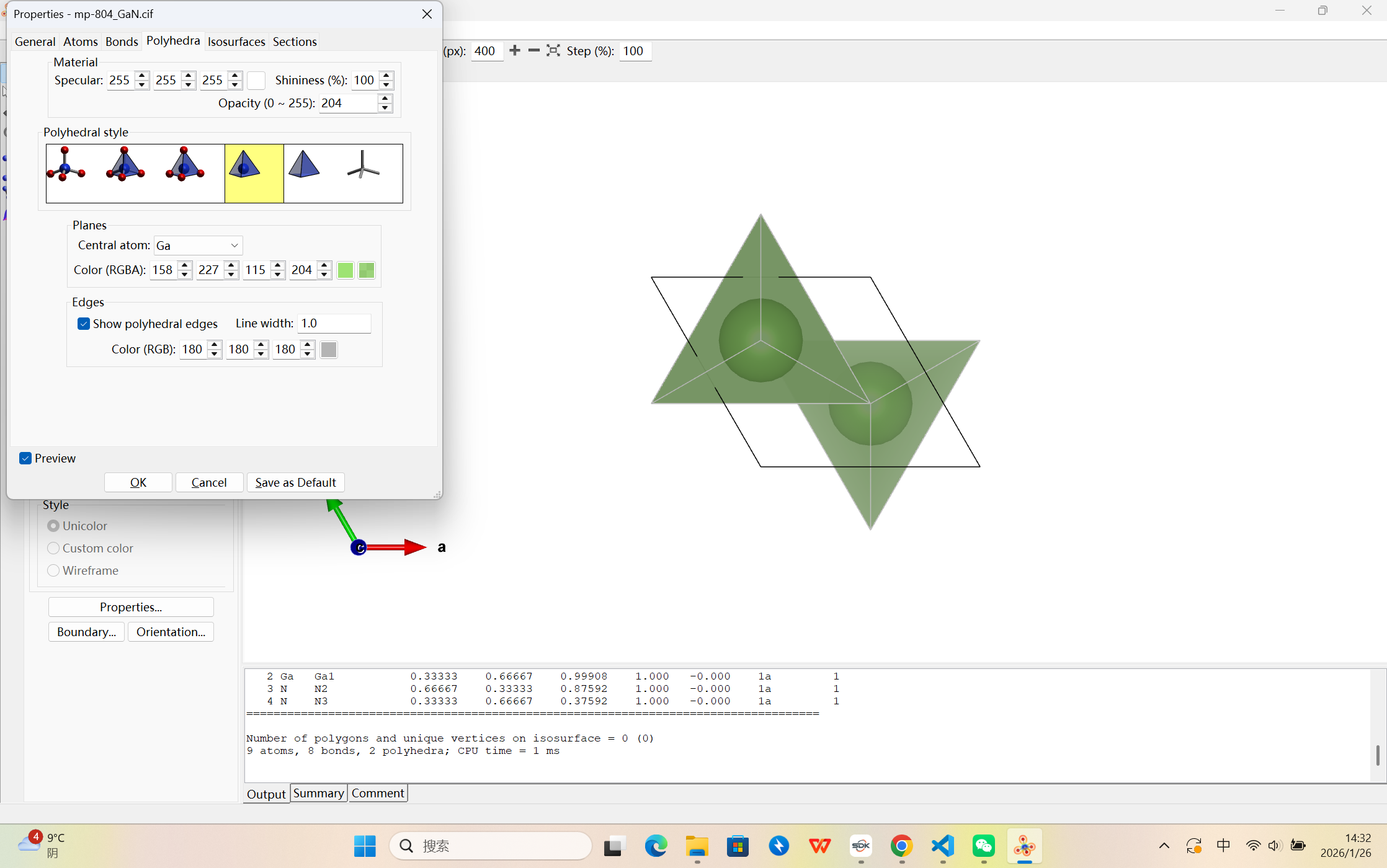}\\[-1pt]
    {\small\textbf{(c) Highlighted polyhedral icon}}\\[-1pt]
    {\scriptsize
    HSV thresholding locates the yellow-highlighted icon and verifies the
    selected polyhedral style index.}
  \end{minipage}

  \vspace{7pt}

  \begin{minipage}[t]{0.31\linewidth}
    \centering
    \includegraphics[
      width=\linewidth,
      height=3.0cm,
      keepaspectratio
    ]{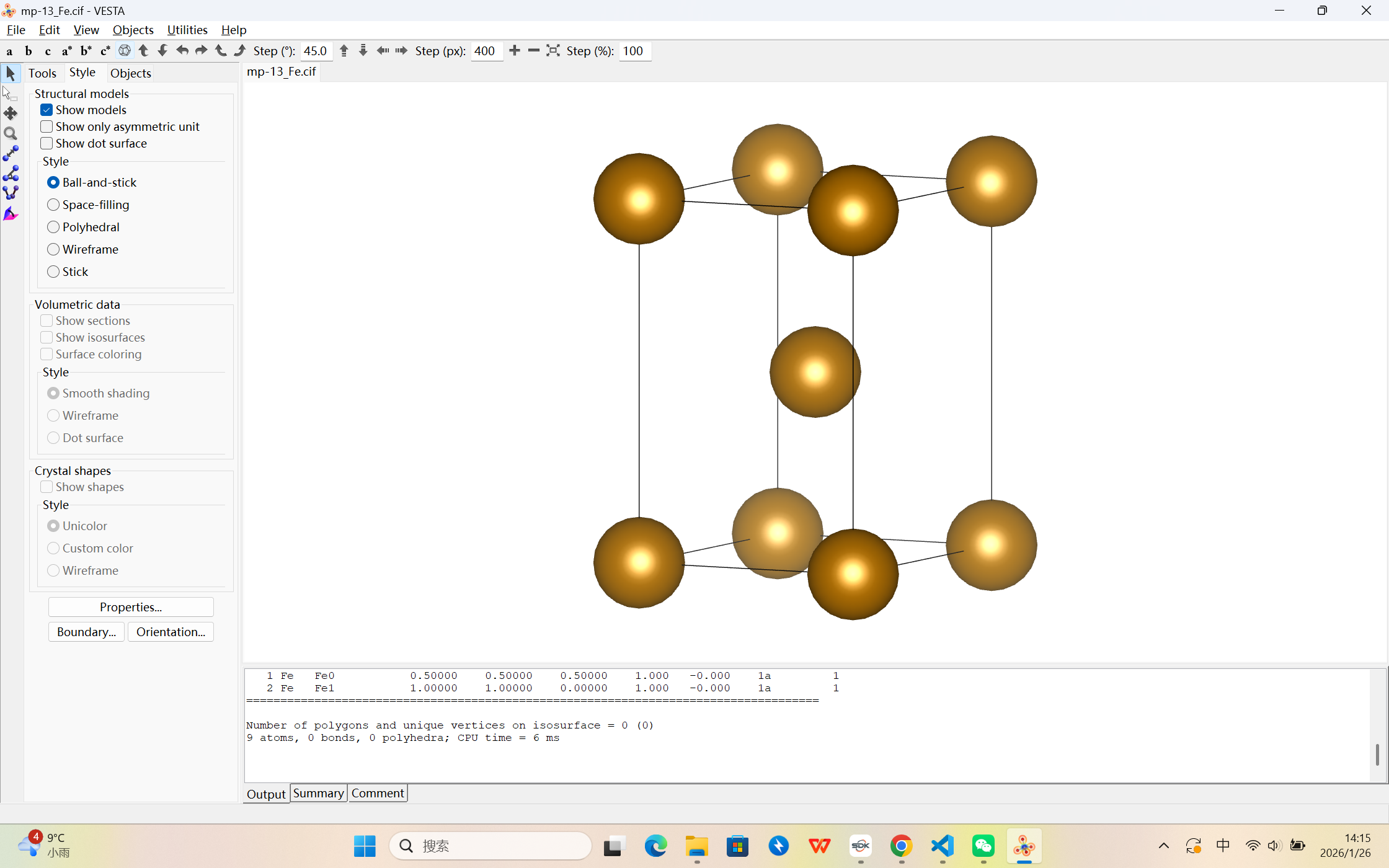}\\[-1pt]
    {\small\textbf{(d) Standard orientation by SSIM}}\\[-1pt]
    {\scriptsize
    The central crystal region is compared with a reference image;
    SSIM $\geq 0.9$ indicates the target orientation.}
  \end{minipage}
  \hfill
  \begin{minipage}[t]{0.31\linewidth}
    \centering
    \includegraphics[
      width=\linewidth,
      height=3.0cm,
      keepaspectratio
    ]{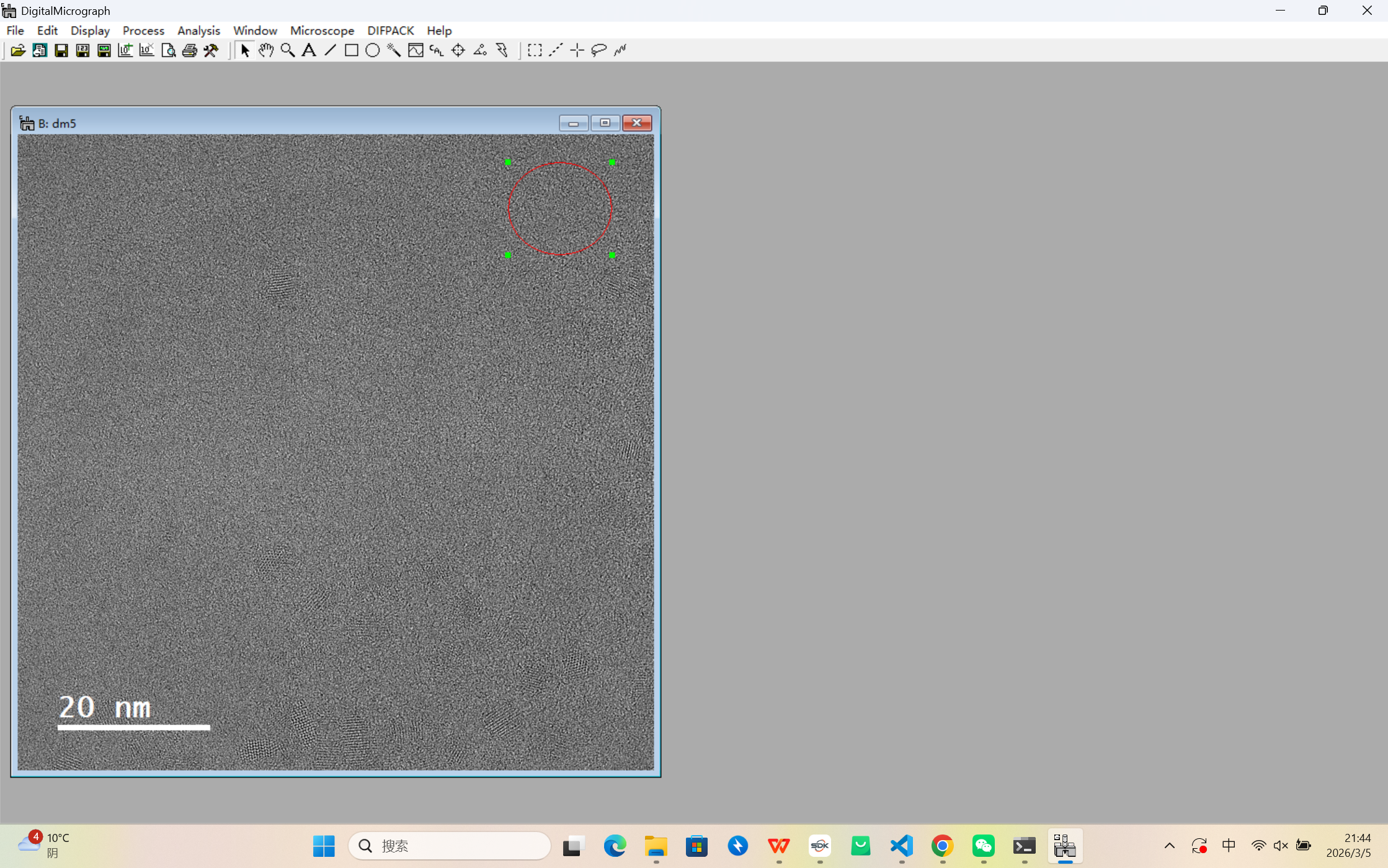}\\[-1pt]
    {\small\textbf{(e) Ellipse detection}}\\[-1pt]
    {\scriptsize
    At least four green control handles detected in HSV space confirm that an
    ellipse has been drawn.}
  \end{minipage}
  \hfill
  \begin{minipage}[t]{0.31\linewidth}
    \centering
    \includegraphics[
      width=\linewidth,
      height=3.0cm,
      keepaspectratio
    ]{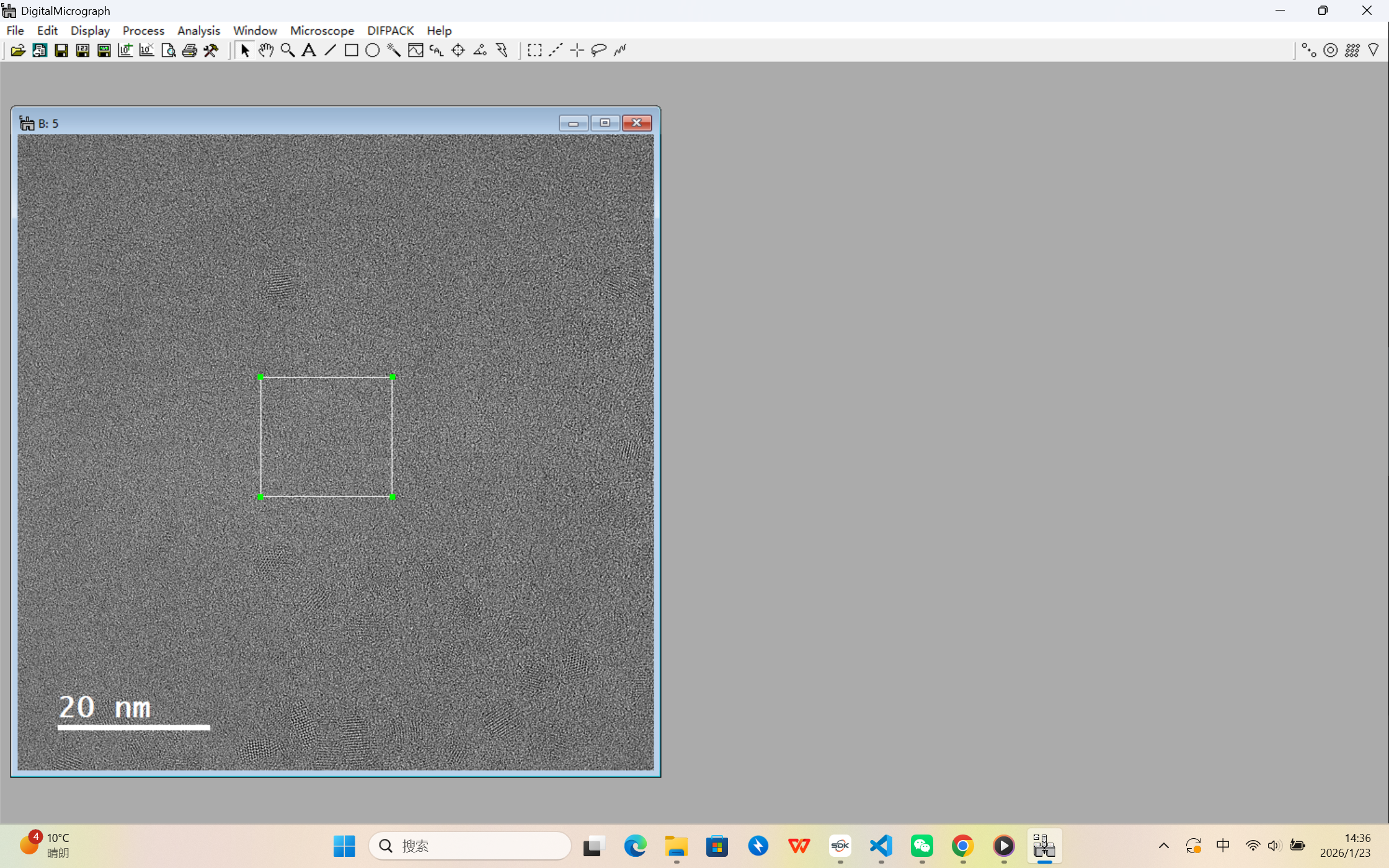}\\[-1pt]
    {\small\textbf{(f) Rectangle detection}}\\[-1pt]
    {\scriptsize
    Four green corner handles verify that the rectangle tool was used and the
    shape is present.}
  \end{minipage}

  \caption{Representative visual-state detection methods used by the GUI
  evaluators. Panels (a)--(b) identify the active VESTA rendering style from
  the vertical position of blue radio-button pixels. Panel (c) detects the
  yellow-highlighted polyhedral icon using HSV thresholding. Panel (d) uses
  structural similarity against a reference image to validate the target
  crystal orientation. Panels (e)--(f) count green control handles in
  DigitalMicrograph to confirm ellipse and rectangle creation. Each detection
  result is paired with \texttt{exact\_match} to produce a binary
  sub-criterion score.}
  \label{fig:visual-detection}
\end{figure*}
\section{Appendix M. Visual Pixel Detection Methods}
\label{sec:visual-detection}

\noindent Beyond OCR text recognition, \textsc{MatToolBench} evaluators use three
visual pixel detection methods, corresponding to
\texttt{check\_style\_change}, 
\texttt{check\_polyhedral\_style},\\
\texttt{check\_visual\_similarity}, and \texttt{check\_border\_color}
in Table~\ref{tab:eval-gui-getters}.
These methods target GUI states where text labels are absent or insufficient: for example,
the active rendering style in VESTA is encoded as the position of a coloured radio button
rather than a text label; the selected polyhedral style is indicated by a coloured highlight
on an icon; and the completion of a drawing operation in DigitalMicrograph is indicated by
the appearance of green corner handles.
All three methods operate directly on raw pixel data captured from the VM screenshot,
without relying on the application's accessibility tree or any structured UI metadata,
and return a binary signal that is passed to \texttt{exact\_match} to produce the
final sub-criterion score.
Figure~\ref{fig:visual-detection} provides concrete examples of all three detection strategies,
together with the expected visual cues and the decision logic applied in each case.

Table~\ref{tab:eval-gui-getters} lists the complete set of getter types used
across all five GUI-based domains.
Each getter encapsulates a self-contained detection strategy — OCR, pixel
colour analysis, or structural similarity — and is paired with
\texttt{exact\_match} to produce a binary pass/fail score.
The diversity of getter types reflects the heterogeneity of GUI states in
real materials-science software, where a single workflow may require
confirming a dialog title, reading a numeric field, and verifying a
rendering style change simultaneously.

\section{Appendix N. Agent System Prompts}
\label{sec:agent-prompts}
\noindent Each agent type receives a tailored system prompt that defines its interaction
modality, available actions, and domain-specific workflow instructions.
The three base prompt families are shown below in condensed form.

\vspace{0.3cm}

\begin{tcolorbox}[guibox, title={GUI Agent: System Prompt (condensed)}]
\small\sffamily
\textbf{You are MatGUI Helper}, an AI agent that controls materials-science GUI software
on a user's computer to complete analytical tasks.
You interact with desktop applications such as Jade (XRD), Avantage (XPS),
VESTA (crystal structure), DigitalMicrograph, and Materials Studio.

\medskip
\textbf{Guidelines:} plan efficiently; execute ONE action per step;
verify progress using the previous screenshot; use coordinates and hotkeys carefully.

\medskip
\textbf{Outputs per step:} screen analysis $\to$ multi-step plan $\to$
next-step rationale $\to$ \texttt{decision} block (COMMAND / DONE / FAIL / WAIT)
$\to$ \texttt{python} code block $\to$ \texttt{memory} update.

\medskip
\textbf{Per-application startup hints (hint mode only):}
\begin{itemize}[leftmargin=*,nosep,topsep=2pt]
  \item \textbf{JADE (ablation-critical):} The ``Read Pattern Files'' database dialog is
    PRE-OPENED at startup. Close it \emph{first} — it accepts only reference patterns,
    not sample data files. Attempting to load task files here causes an unrecoverable deadlock.
  \item \textbf{Avantage:} A file Open dialog is PRE-TRIGGERED (Ctrl+O).
    Type the file path and press Enter directly.
    Workflow is self-evident from the interface; not included in the ablation.
  \item \textbf{VESTA:} A file Open dialog is PRE-TRIGGERED.
    Type the path and press Enter.
    Workflow is self-evident from the interface; not included in the ablation.
  \item \textbf{DM (ablation-critical):} A file Open dialog is PRE-TRIGGERED.
    At startup a floating panel (``Histogram'', ``Images'', ``Status'', etc.) occupies
    the left side of the workspace — close it \emph{first}.
    Without this hint, the panel can obscure the filename shown in the title bar,
    directly disrupting the evaluator's file-import check and causing incorrect scoring.
  \item \textbf{Materials Studio:} Maximize the window \emph{before} any other action.
  \item \textbf{Origin:} The target \texttt{.opju} file is already open and
    Code Builder is already open (Alt+4 pre-pressed).
\end{itemize}
\end{tcolorbox}

\vspace{0.3cm}

\begin{tcolorbox}[codebox, title={Code Agent: System Prompt (condensed)}]
\small\sffamily
\textbf{You are MatCode Helper}, an AI agent that writes Python scripts to query
materials-science databases and APIs.

\medskip
\textbf{Output:} a single, complete, self-contained \texttt{python} code block that
(1) queries the specified database, (2) processes results, and (3) writes output to
the exact file path given in the task.

\medskip
\textbf{Rules:} Python 3.10; mp-api 0.39.5 / emmet-core 0.78.7; no GUI windows;
UTF-8 output; handle errors gracefully; do not include venv activation.

\medskip
\textbf{One-shot API examples (hint mode only):}
\begin{itemize}[leftmargin=*,nosep,topsep=2pt]
  \item \textbf{MP:} use the Materials Project summary-search API with
    task-specific elements and returned fields.
  \item \textbf{OQMD:} use \texttt{requests.get(...)} on the formation-energy endpoint
    with retry logic.
  \item \textbf{OPTIMADE:} \texttt{requests.get(}\\
    \texttt{"}\url{https://optimade.materialsproject.org/v1/structures}\texttt{",}\\
    \texttt{params=\{"filter": "...", "page\_limit": 100\})} with auto-pagination
  \item \textbf{Pymatgen:} \texttt{Structure.from\_file(path)},
    \texttt{SpacegroupAnalyzer}, \texttt{Poscar}
\end{itemize}
\end{tcolorbox}

\vspace{0.3cm}

\begin{table*}[!htbp]
\centering
\caption{Overview of the ten software tools and APIs covered by \textsc{MatToolBench}.}
\label{tab:tool-descriptions}
\small
\resizebox{0.95\textwidth}{!}{%
\begin{tabular}{L{3.0cm}L{2.8cm}L{10.2cm}}
\toprule
\textbf{Tool} & \textbf{Domain} & \textbf{Description} \\
\midrule
JADE & characterization & MDI JADE is a commercial XRD analysis suite for phase identification, Rietveld refinement, and diffractogram comparison using the PDF card database. \\
Avantage & characterization & Thermo Fisher Avantage is the standard software for XPS spectrum processing, including peak fitting, quantification, and elemental binding-energy analysis. \\
VESTA & Visualisation & VESTA (Visualisation for Electronic and STructural Analysis) renders crystal structures in 3-D; tasks cover bond/polyhedra style changes, CIF import, and orientation control. \\
DigitalMicrograph & characterization & Gatan DigitalMicrograph (DM) is the de-facto TEM image analysis platform; tasks cover FFT, line profiles, annotation, and drawing-tool operations. \\
Materials Studio & Simulation & Dassault BIOVIA Materials Studio is a molecular modelling environment; tasks require building supercells, setting force-field parameters, and running geometry optimisations. \\
OriginPro & Analysis & OriginLab OriginPro is a scientific graphing and statistics package; tasks span curve fitting, multi-panel figure layout, axis formatting, and batch export via Code Builder (Python). \\
Materials Project & Database & The Materials Project (MP) REST API provides computed properties (bandgap, formation energy, elastic moduli) for $>$150 000 inorganic compounds. \\
OQMD & Database & The Open Quantum Materials Database (OQMD) offers DFT-computed thermodynamic properties for $>$1 million crystal structures via a public REST API. \\
OPTIMADE & Database & The OPTIMADE standard defines a unified JSON:API query language across multiple materials databases; tasks require constructing cross-database queries with filter expressions. \\
Pymatgen & Simulation & Pymatgen (Python Materials Genomics) is a Python library for materials analysis; tasks cover structure creation, POSCAR/vasprun.xml parsing, INCAR editing, and energy extraction. \\
\bottomrule
\end{tabular}%
}
\end{table*}
\begin{tcolorbox}[originbox, title={Origin Agent: System Prompt (condensed)}]
\small\sffamily
\textbf{You are an AI agent} that controls OriginLab to complete plotting and
data-analysis tasks by writing Python (originpro) scripts in Code Builder.

\medskip
\textbf{Standard workflow (hint mode):}
\begin{enumerate}[leftmargin=*,nosep,topsep=2pt]
  \item Data file is already open in Origin's worksheet — do NOT reopen it.
  \item Code Builder is already open (Alt+4 pre-pressed) — go straight to step~3.
  \item Press Ctrl+N to create a new Python file; name it to match the plot type
    (\texttt{raman.py}, \texttt{xrd.py}, \texttt{xps.py}, \texttt{ftir.py}, …).
  \item Paste the full script via clipboard:
    \texttt{copy\_text("...script...")} $\to$ \texttt{paste()}.
    \textbf{Never use} \texttt{keyboard.write()} for scripts — ASCII-only, corrupts Unicode.
  \item Press F5 to run. Fix errors by clearing (Ctrl+A, Delete) and re-pasting.
  \item Verify the output file exists at the specified path, then mark DONE.
\end{enumerate}

\medskip
\textbf{Baseline mode (no\_hint):} No workflow instructions are provided.
The agent must independently discover whether to use Code Builder or GUI-only
interaction, and must write the entire plotting script without a template.
Code Builder is \emph{not} pre-opened, removing the implicit workflow signal.

\medskip
\textbf{Available packages:} \texttt{originpro}, \texttt{numpy}, \texttt{pandas},
\texttt{matplotlib}, \texttt{scipy}, \texttt{re}, \texttt{os}, \texttt{math}.
\end{tcolorbox}

\section{Appendix O. Software and Tool Descriptions}
\label{sec:tool-descriptions}

\noindent \textsc{MatToolBench} spans ten domain-specific software packages drawn from experimental
characterization, computational modelling, and database retrieval workflows.
Table~\ref{tab:tool-descriptions} provides a brief description of each tool, its domain, and the
primary task categories it covers in the benchmark.

\noindent The ten tools span four interaction modalities: (1)~GUI-only software operated entirely through
mouse-and-keyboard actions; (2)~GUI software with an embedded scripting interface (OriginPro Code Builder);
(3)~REST API databases accessed via Python; and (4)~file-based structure analysis pipelines (Pymatgen).
This mix prevents the benchmark from reducing to a single interaction paradigm.

\section{Appendix P. Failure Mode Analysis}
\label{sec:failure-modes}

\noindent To gain qualitative insight beyond aggregate scores, we manually inspected
agent trajectories from the Doubao-seed-1-8 experiment across all task types and identified
four recurring failure patterns, illustrated with real trajectory screenshots below.

\subsection{GUI Grounding Errors}
\label{sec:failure-grounding}

\begin{figure}[htbp]
  \centering
  \begin{minipage}[b]{0.48\linewidth}
    \includegraphics[width=\linewidth]{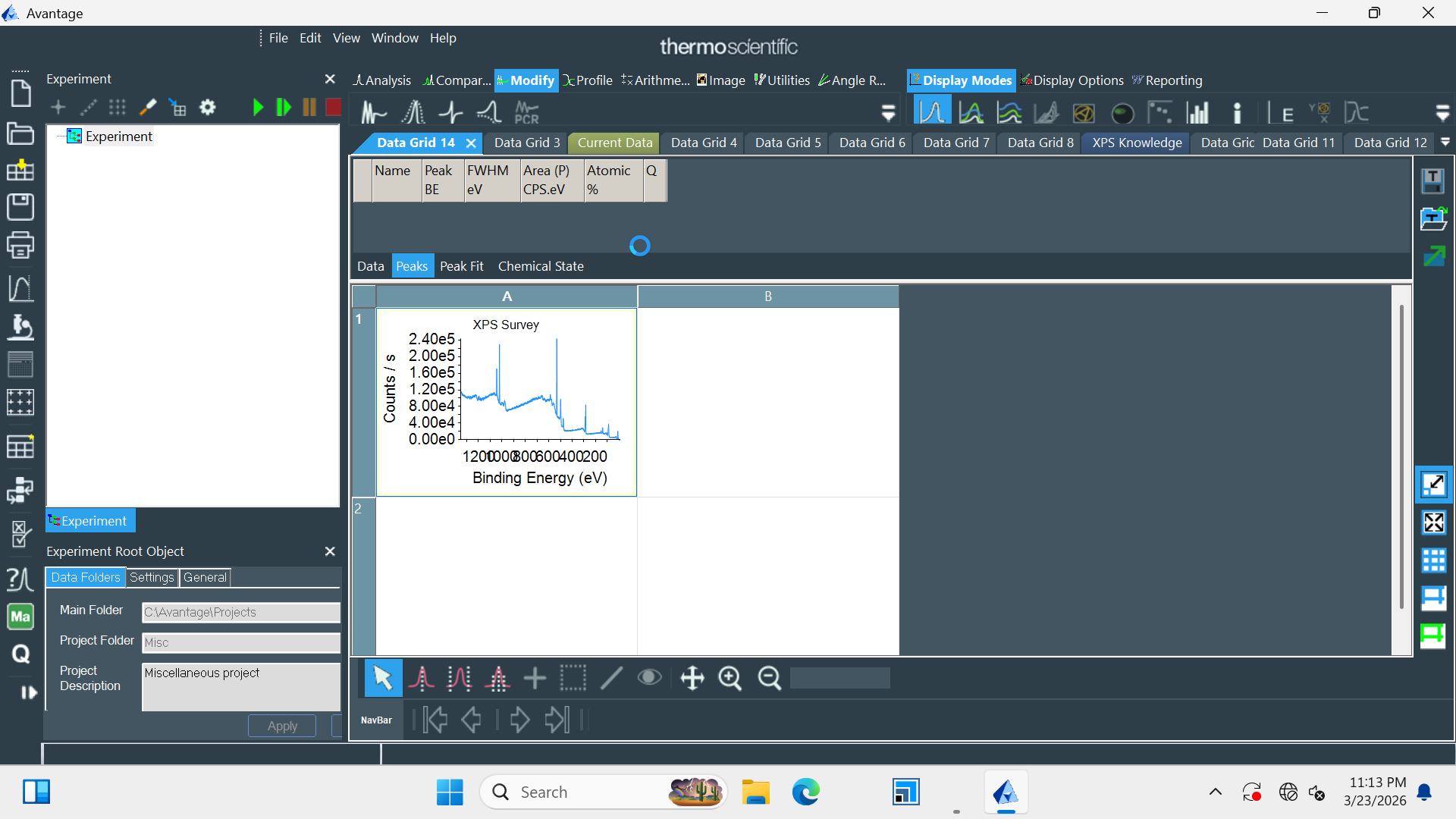}
    {\scriptsize (a) Step~20: one spectrum loaded in \textit{Data Grid 14}.}
  \end{minipage}\hfill
  \begin{minipage}[b]{0.48\linewidth}
    \includegraphics[width=\linewidth]{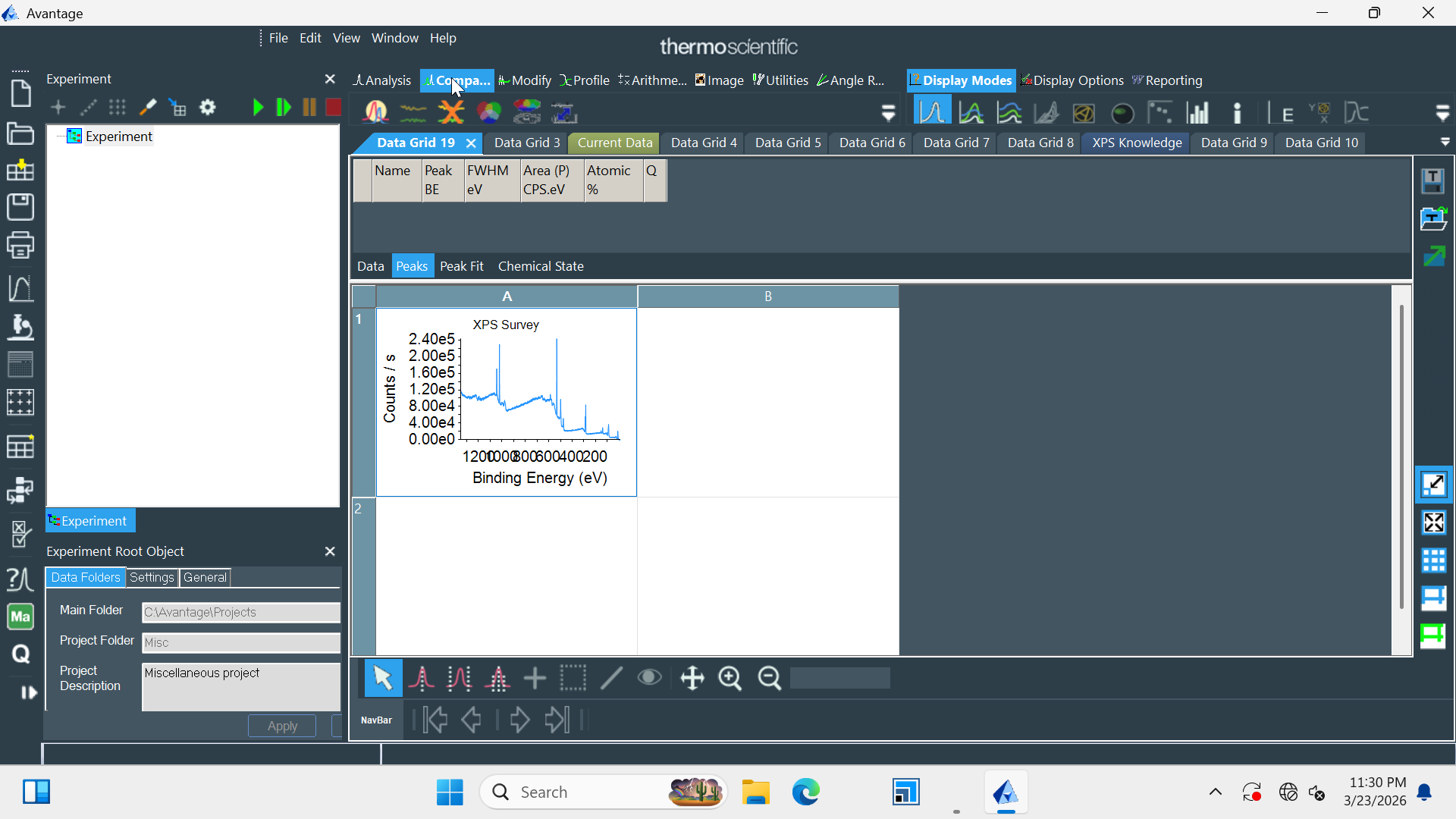}
    {\scriptsize (b) Step~40: repeated misclicks open 19 data grids.}
  \end{minipage}
  \caption{GUI grounding error in an Avantage comparison-view task (Doubao-seed-1-8).
    Each misclick on a visually similar toolbar icon opens a new independent data grid
    instead of selecting the target layout option.}
  \label{fig:failure-grounding}
\end{figure}

The most frequent failure mode for GUI agents is incorrect element localisation: the agent
generates a valid high-level action (e.g., ``click the Display Modes tab'') but the coordinate
prediction lands on the wrong widget or opens an unintended panel.
This is especially common in Avantage, where multiple toolbars contain visually similar icons.
Figure~\ref{fig:failure-grounding} shows a representative trajectory in which the agent attempts
to arrange two spectra in comparison view.
At step~20, the agent has imported one file and is inspecting the spectrum under \textit{Data Grid 14};
by step~40, repeated misclicks on toolbar icons have opened 19 independent data grids
(\textit{Data Grid 19} is active) without ever selecting both spectra or invoking the stacking layout,
ultimately exhausting the step budget.
In approximately 30\% of inspected GUI trajectories at least one grounding error occurred;
in half of these cases the agent self-corrected, but in the remaining cases the trajectory diverged unrecoverably.

\begin{figure}[htbp]
  \centering
  \begin{minipage}[b]{0.48\linewidth}
    \includegraphics[width=\linewidth]{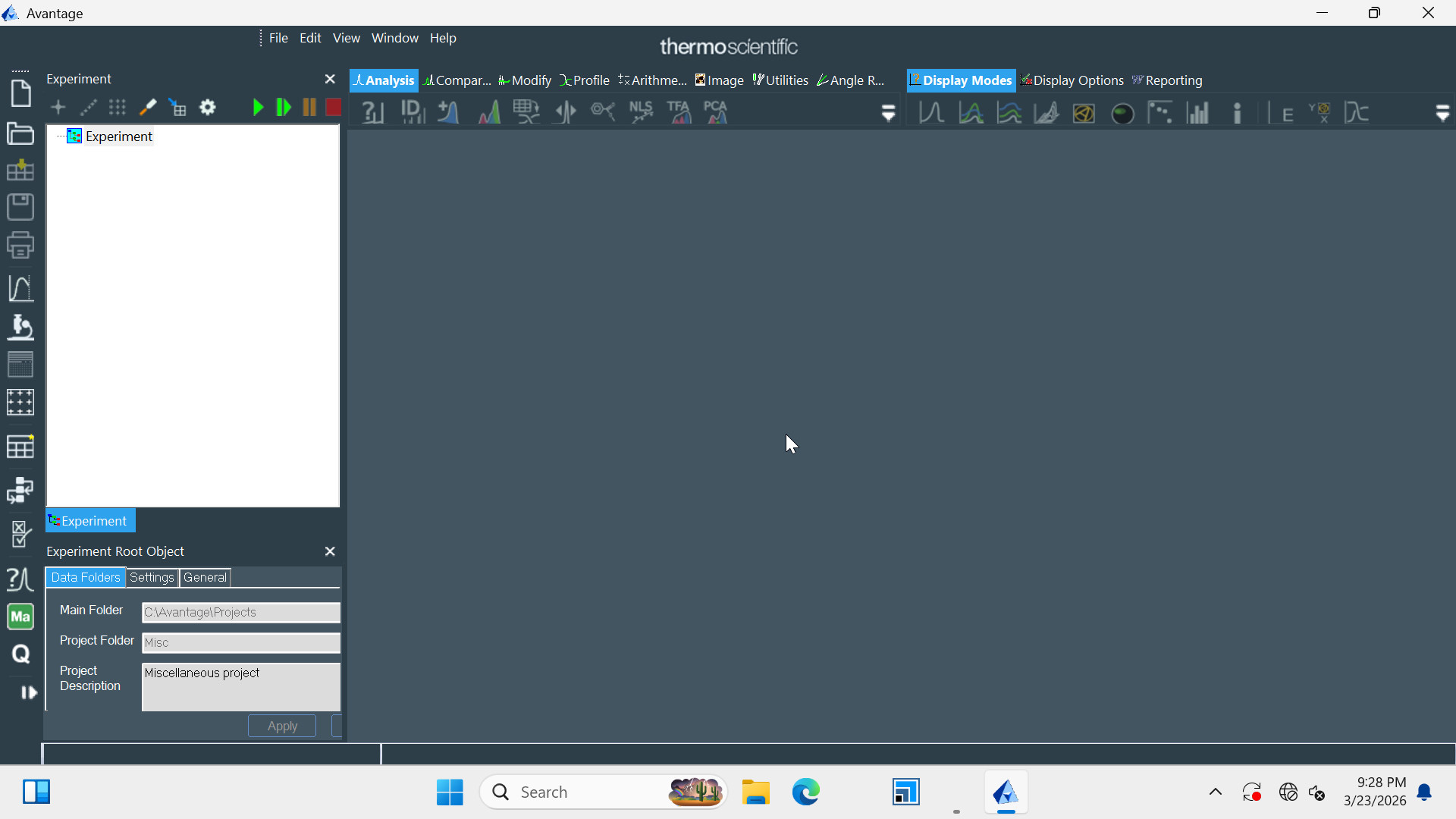}
    {\scriptsize (a) Step~0: empty workspace; no data imported.}
  \end{minipage}\hfill
  \begin{minipage}[b]{0.48\linewidth}
    \includegraphics[width=\linewidth]{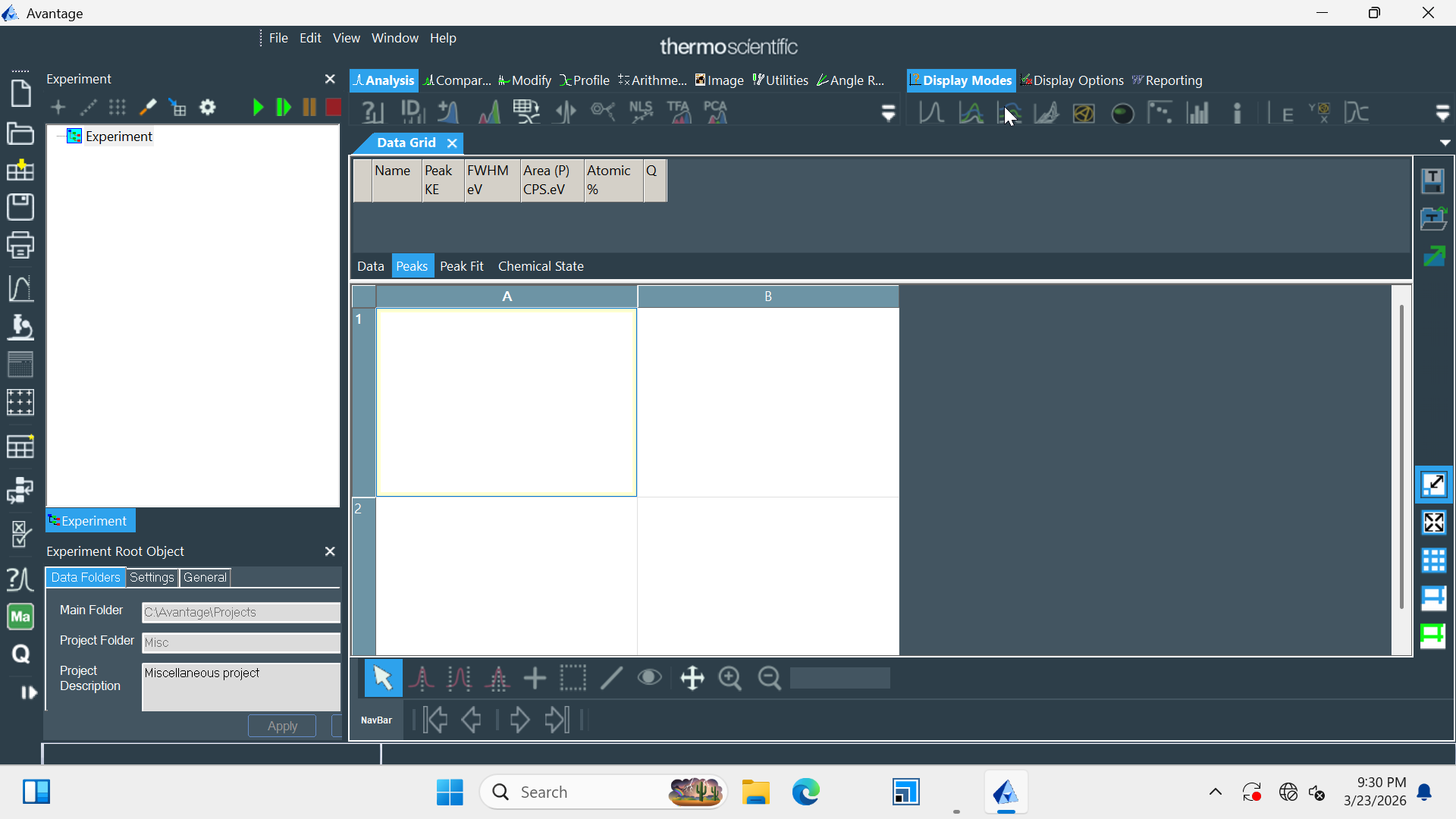}
    {\scriptsize (b) Step~6: \texttt{DONE} issued while canvas remains empty.}
  \end{minipage}
  \caption{Premature termination in an Avantage stacked-graph task (Doubao-seed-1-8).
    The agent mistakes the pre-highlighted ``Display Modes'' tab for evidence of
    task completion after only 6 of $\leq$50 steps.}
  \label{fig:failure-premature}
\end{figure}

\subsection{Premature Termination}
\label{sec:failure-premature}

A significant fraction of failures result from the agent issuing a \texttt{DONE} action
too early — declaring task completion before the required steps (e.g., file import,
stacked-graph layout, file export) have been performed.
Figure~\ref{fig:failure-premature} shows an Avantage task requiring the agent to import
\texttt{O1s Scan.VGD} and enable stacked-graph display mode.
The initial state (step~0) shows an empty workspace; after only 6 steps the agent
observes that the ``Display Modes'' toolbar tab appears highlighted (it is highlighted by default)
and issues \texttt{DONE}, leaving the canvas empty with no data ever imported.
This pattern is more pronounced for longer tasks ($\geq$8 steps) and correlates with
agents losing track of progress in the absence of explicit state memory.
Workflow hints reduce this failure mode as shown in Section~\ref{sec:ablation}.

\subsection{Deadlock / Repetitive Loops}
\label{sec:failure-deadlock}

\begin{figure}[htbp]
  \centering
  \begin{minipage}[b]{0.48\linewidth}
    \includegraphics[width=\linewidth]{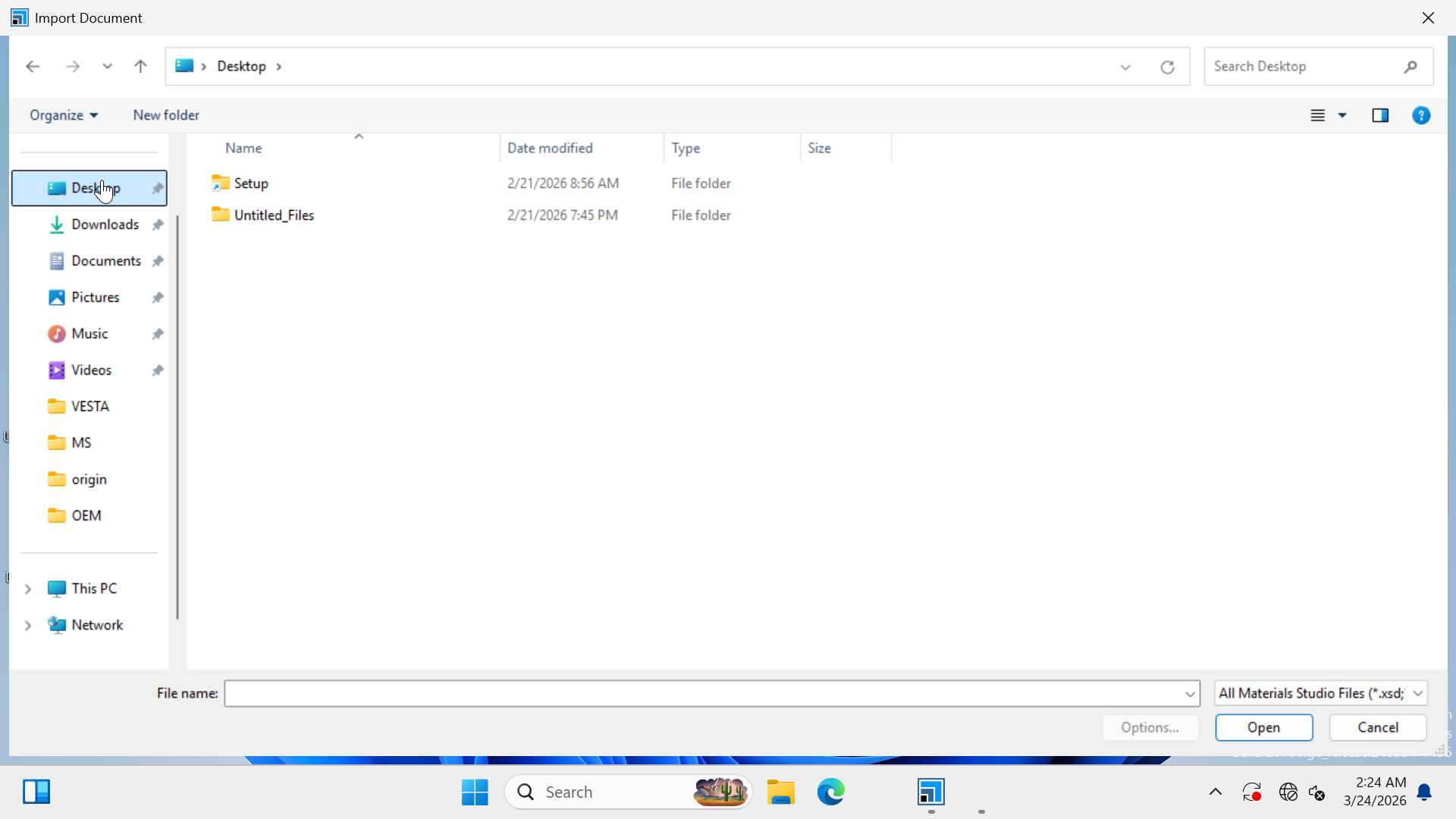}
    {\scriptsize (a) Step~45: import dialog stuck at Desktop.}
  \end{minipage}\hfill
  \begin{minipage}[b]{0.48\linewidth}
    \includegraphics[width=\linewidth]{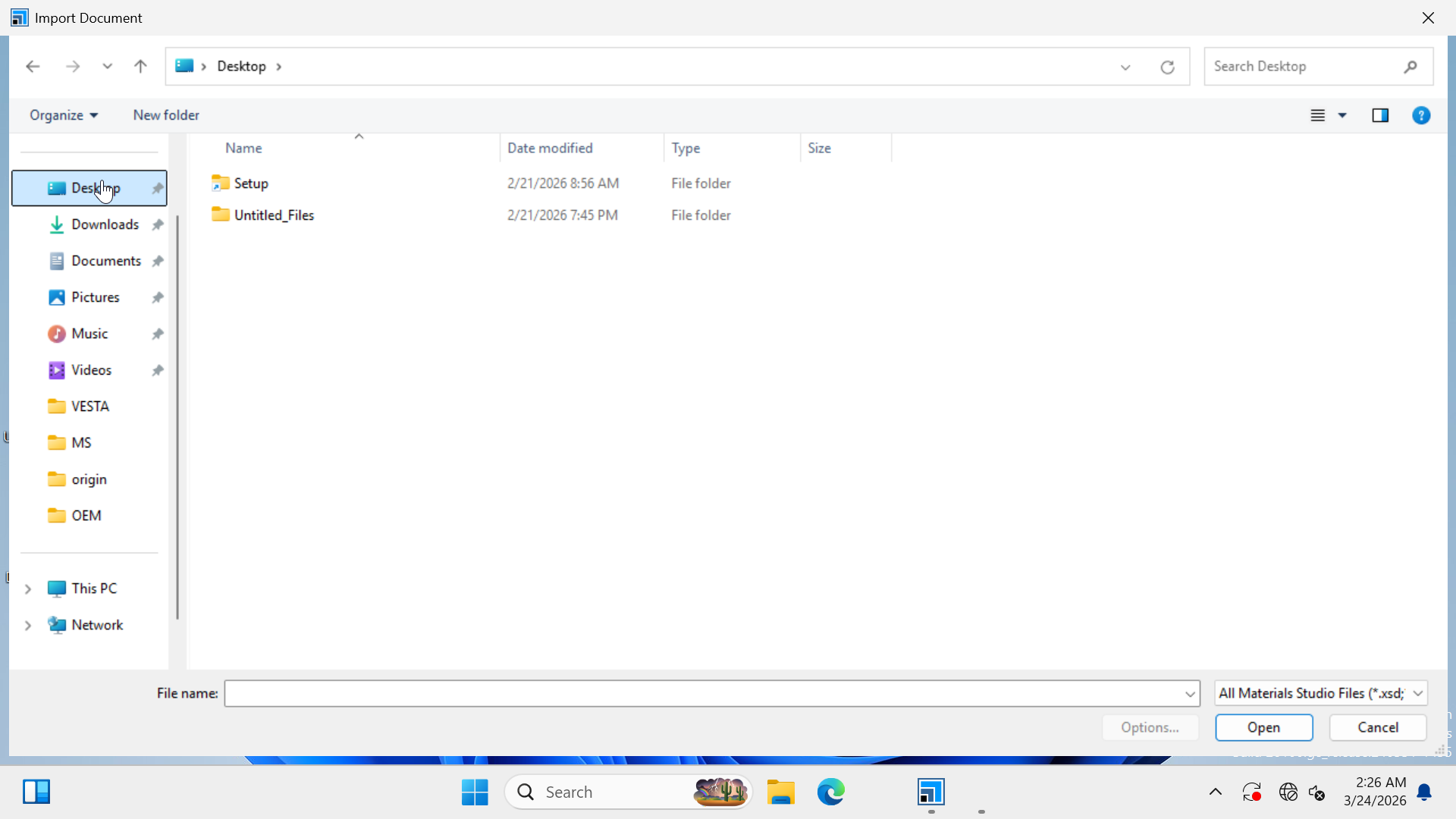}
    {\scriptsize (b) Step~49: identical UI state four steps later.}
  \end{minipage}
  \caption{Deadlock in a Materials Studio file-import task (Doubao-seed-1-8).
    The agent alternates clicks between two adjacent left-panel entries every step,
    never entering the target subdirectory; the step budget expires with the dialog still open.}
  \label{fig:failure-deadlock}
\end{figure}

For tasks involving modal dialogues or complex nested menus, agents occasionally
enter a repetitive loop: they perform the same unsuccessful action sequence
repeatedly without escaping to an alternative approach.
Figure~\ref{fig:failure-deadlock} shows a Materials Studio task in which the agent
must navigate to a specific \texttt{.xsd} file via an ``Import Document'' dialogue.
The agent becomes stuck in the Desktop folder, alternating clicks between
the left-panel shortcut and an adjacent folder entry every step. Steps~45 and~49
show an identical UI state: the dialogue remains open at the Desktop root, the
file name field is empty, and no progress has been made.

\subsection{Domain Knowledge Gaps}
\label{sec:failure-domain}

\begin{figure}[htbp]
  \centering
  \begin{minipage}[b]{0.48\linewidth}
    \includegraphics[width=\linewidth]{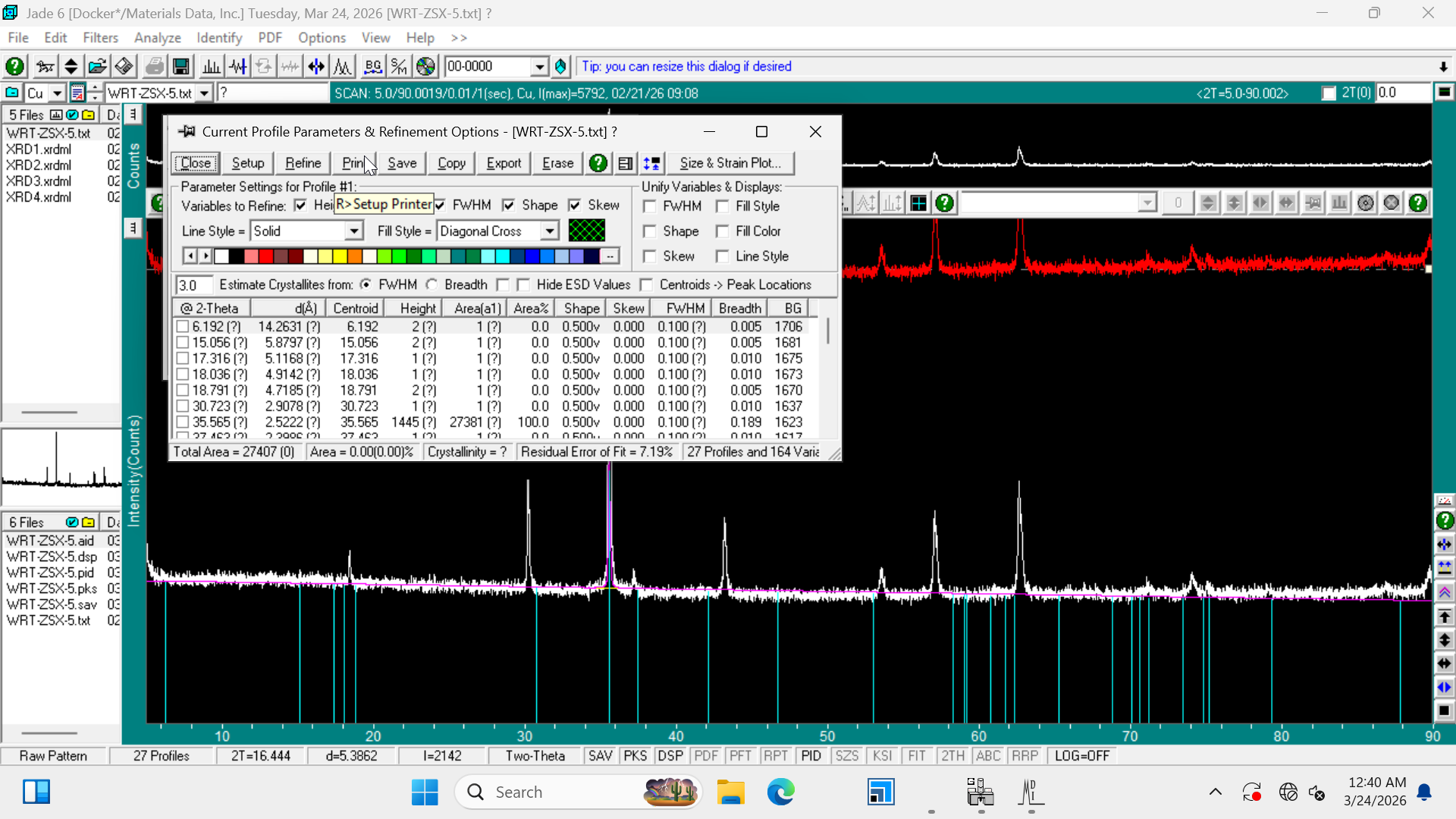}
    {\scriptsize (a) Step~45: fitting converged; parameters populated.}
  \end{minipage}\hfill
  \begin{minipage}[b]{0.48\linewidth}
    \includegraphics[width=\linewidth]{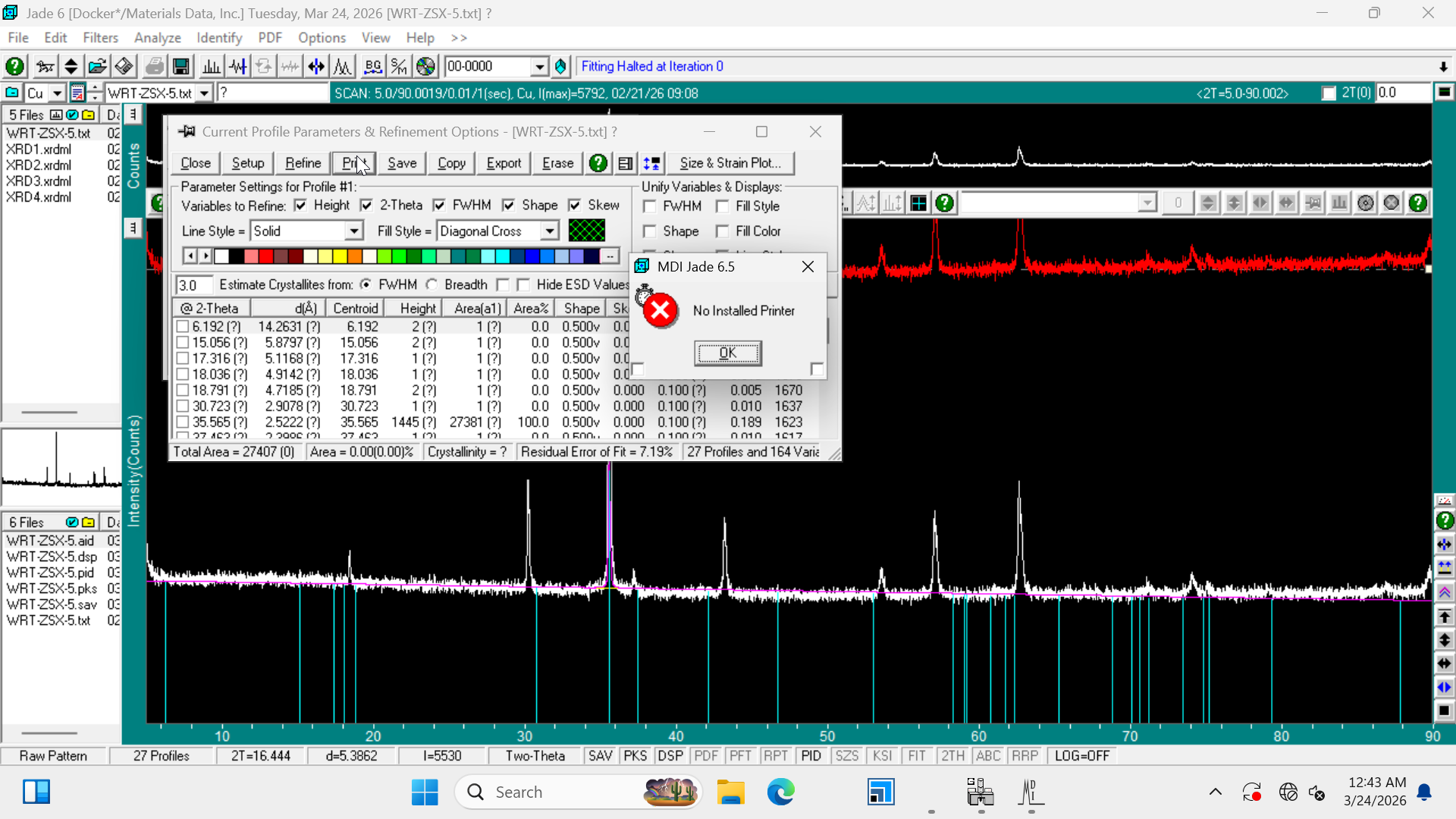}
    {\scriptsize (b) Step~49: no printer is available for report export.}
  \end{minipage}
  \caption{Domain knowledge gap in a JADE profile-fitting task (Doubao-seed-1-8).
    The agent correctly completes the fitting but fails at the final export step
    because it lacks knowledge that the VM has no virtual PDF printer installed.}
  \label{fig:failure-domain}
\end{figure}

GUI and code agents sometimes fail because of missing materials-science or
tool-specific knowledge that cannot be inferred from the interface alone.
Figure~\ref{fig:failure-domain} shows a JADE task requiring the agent to perform
profile fitting and then print the report to a PDF file.
At step~45 the fitting has converged and the parameter table is populated correctly;
at step~49 the agent invokes the \textit{Print} menu expecting to export to PDF,
but receives a ``\texttt{JADE 5.5 — No Installed Printers}'' error dialog because
the evaluation VM has no virtual PDF printer configured.

\noindent Similar gaps appear in code tasks: scripts that are syntactically correct
but use wrong units (e.g., lattice parameter in nm instead of \AA) or outdated API
endpoints run without raising exceptions, silently producing incorrect output and
scoring zero.
\section{Appendix Q. Contrast with General GUI Benchmarks}
\label{sec:general-gui-contrast}

\noindent \textsc{MatToolBench} complements broad GUI-control benchmarks by
isolating specialized materials-software workflows. We use three existing
benchmark families as diagnostic context rather than as a controlled leaderboard:
OSWorld/OSWorld-Verified for broad desktop-control performance,
WindowsAgentArena for general Windows tasks under comparable 50-step settings,
and MMBench-GUI for hierarchical GUI automation and cross-application tasks
across platforms.
Across these references, general GUI agents can obtain substantially higher
success rates on broad desktop or Windows tasks than on \textsc{MatToolBench},
while MMBench-GUI shows that long-horizon automation and cross-application
collaboration remain difficult even outside scientific software. These comparisons
therefore support a transfer-gap diagnostic rather than a controlled head-to-head
claim, since the benchmarks differ in task distribution, agent wrappers,
observations, action spaces, and evaluators.

\begin{center}
\captionof{table}{Diagnostic comparison with general GUI benchmarks. Values are drawn from public reports and are not a controlled leaderboard comparison.}
\label{tab:general-gui-contrast}
\scriptsize
\setlength{\tabcolsep}{3pt}
\resizebox{\linewidth}{!}{%
\begin{tabular}{lccc}
\toprule
\textbf{Benchmark context} & \textbf{Public setting / result} & \textbf{\textsc{MatToolBench} reference} & \textbf{Diagnostic role} \\
\midrule
OSWorld-Verified & Matched/closest families report 41.6--72.1\% SR at 100 steps & 7.0--25.0\% GUI SR & Broad desktop contrast \\
WindowsAgentArena & GPT-5 and Qwen3-VL-32B families report 63.5\% and 45.3\% SR at 50 steps & 21.0\% and 7.0\% GUI SR & Same step-budget Windows contrast \\
MMBench-GUI & Best 50-step L3/L4 average SR: 26.6\% / 8.8\% & Best \textsc{MatToolBench} GUI SR: 25.0\% & Automation difficulty context \\
\bottomrule
\end{tabular}
}
\end{center}

\noindent The qualitative source of the gap is also different. OSWorld and
WindowsAgentArena emphasize everyday applications, system operations, web/file
workflows, and general Windows productivity tasks. MMBench-GUI further shows
that even broad 50-step single-application automation remains below 30\% average
SR, and cross-application collaboration remains below 10\% for the best reported
method. In contrast, \textsc{MatToolBench} requires preserving scientific state
across specialized interfaces, domain file formats, instrument-like visual
feedback, and software-specific operational conventions. The mixed tasks extend
this idea only as a small diagnostic set for artifact handoff; the main
quantitative conclusions are based on the larger GUI, Origin, and code splits.

\section{Appendix R. Limitations and Broader Impacts}
\label{sec:limitations}

\paragraph{Limitations.}
\textsc{MatToolBench} focuses on a curated set of ten tools that are representative of
experimental and computational materials science, but the selection inevitably omits
important platforms such as LabView instrument-control workflows, experimental-planning
notebooks, and proprietary simulation codes.
Task difficulty is intentionally constrained to single-session interactions of at most 50 steps;
longer multi-session or multi-agent pipelines — increasingly relevant in autonomous laboratory
settings — are outside the current scope.
The automated graders achieve $>$97\% agreement with human annotators on average
(Table~\ref{tab:per_task_detail}), but edge cases remain, particularly for tasks involving
visually-rendered crystal structures where SSIM thresholds may be sensitive to GPU-dependent
anti-aliasing differences.
Finally, the benchmark evaluates English-only agent prompts; performance may differ for agents
instructed in other languages.

\paragraph{Broader Impacts.}
\textsc{MatToolBench} is intended to accelerate the development of AI assistants for materials
researchers.
Reliable autonomous agents could reduce the time experts spend on routine software operation
and lower the barrier of entry for early-career researchers unfamiliar with legacy instruments.
At the same time, over-reliance on automated agents for instrument operation carries risks:
incorrect parameter settings in characterization software (e.g., wrong peak assignments in XPS)
could propagate errors into published datasets.
We therefore recommend that any agent-assisted workflow include a human-in-the-loop review step
before results are used for publication or engineering decisions.
The benchmark data and code are released under a permissive open-source licence to encourage
transparent and reproducible research.
\end{document}